\documentclass[main]{main}

\AtBeginDocument{%
  \renewcommand\normalsize{\fontsize{8}{9}\selectfont}
  \renewcommand\small{\fontsize{8}{9}\selectfont}
  \renewcommand\footnotesize{\fontsize{6}{8}\selectfont}
}

\usepackage{comment}
\usepackage{graphicx}%
\usepackage{multirow}%
\usepackage{amsmath,amssymb,amsfonts}%
\usepackage{amsthm}%
\usepackage{mathrsfs}%
\usepackage[title]{appendix}%
\usepackage[table]{xcolor}
\usepackage{csquotes}
\usepackage[margin=1in]{geometry}
\usepackage{accents}
\usepackage{textcomp}%
\usepackage{manyfoot}%
\usepackage{booktabs}%
\usepackage{algorithm}%
\usepackage{algorithmicx}%
\usepackage{algpseudocode}%
\usepackage{listings}%
\usepackage{caption}
\usepackage{subcaption}
\usepackage{orcidlink}
\usepackage{hyperref}
\MakeOuterQuote{"}
\usepackage{etoolbox}
\usepackage{enumitem}
\usepackage{makecell}
\usepackage{url}

\setlist[enumerate]{
  topsep=0pt,    
  partopsep=0pt, 
  parsep=0.1em,    
  itemsep=0.25em,
  font=\normalfont\normalsize}

\setlist[itemize]{
  topsep=0pt,    
  partopsep=0pt, 
  parsep=0.1em,    
  itemsep=0.25em,
  font=\normalfont\normalsize}
  
\theoremstyle{thmstyleone}%
\theoremstyle{thmstyletwo}%

\theoremstyle{thmstylethree}%

\renewenvironment{thebibliography}[1]{
  \begin{oldthebibliography}{#1}
    \setlength{\itemsep}{0em}
    \setlength{\parskip}{0em}
}
{
  \end{oldthebibliography}
}

\begin{document}
%

\title{Surrogate Modeling and Explainable Artificial Intelligence for Complex Systems: A Workflow for Automated Simulation Exploration}%

\author*[1,2]{\fnm{Paul} \sur{Saves}\orcidlink{0000-0001-5889-2302} }\email{paul.saves@ird.fr}
\author[3]{\fnm{Pramudita Satria} \sur{Palar}\orcidlink{0000-0002-7066-0763} }\email{pramsp@itb.ac.id}
\author[4]{\fnm{Muhammad Daffa} \sur{Robani}\orcidlink{0009-0003-0567-504X} }\email{robani.daffa@gmail.com}
\author[2]{\fnm{Nicolas} \sur{Verstaevel}\orcidlink{0000-0002-7879-6681}  }\email{nicolas.verstaevel@irit.fr}
\author[2]{\fnm{Moncef} \sur{Garouani}\orcidlink{0000-0003-2528-441X} }\email{moncef.garouani@irit.fr}
\author[2,5]{\fnm{Jordan} \sur{Levy}\orcidlink{0009-0002-4337-2608} }\email{jordan.levy@irit.fr}
\author[2]{\fnm{Julien} \sur{Aligon}\orcidlink{0000-0002-1954-8733} }\email{julien.aligon@irit.fr}
\author[2]{\fnm{Benoit} \sur{Gaudou}\orcidlink{0000-0002-9005-3004 }}\email{benoit.gaudou@ut-capitole.fr}
\author[6]{\fnm{Koji} \sur{Shimoyama}\orcidlink{0000-0001-8896-7707}}\email{ shimoyama@mech.kyushu-u.ac.jp}
\author[7]{\fnm{Joseph} \sur{Morlier}\orcidlink{0000-0002-1511-2086}}\email{joseph.morlier@isae-supaero.fr}

\affil*[1]{
\orgdiv{IRD, Unité de Modélisation Mathématique et Informatique des Systèmes Complexes,  UMI 209},  
\orgname{Sorbonne Université}, 
\orgaddress{\street{4, Place Jussieu},  \city{Paris}, \postcode{75005},\state{ Île-de-France},\country{ France.\vspace{0.12cm}}}}

\affil[2]{\orgdiv{Université Toulouse Capitole, IRIT},  \city{Toulouse}, \postcode{31000},\state{ Occitanie},\country{ France\vspace{0.12cm}}}

\affil[3]{\orgdiv{Faculty of Mechanical and Aerospace Engineering}, \orgname{Institut Teknologi Bandung}, \orgaddress{\street{Jl. Ganesha 10}, \city{Bandung}, \postcode{15414}, \state{West Java}, \country{Indonesia\vspace{0.12cm}}}}

\affil[4]{\orgdiv{AwanTunai}, \orgname{Urban Suites}, \orgaddress{\street{3 Hullet Road}, \city{Singapore}, \postcode{229158}, \state{Singapore}, \country{Singapore\vspace{0.12cm}}}}

\affil[5]{\orgdiv{TwinswHeel}, \orgname{Soben}, \orgaddress{\city{Cahors}, \postcode{46000}, \state{Occitanie}, \country{France\vspace{0.12cm}}}}

\affil[6]{\orgdiv{Mechanical Engineering}, \orgname{Kyushu University}, \orgaddress{\street{744 Motooka}, \city{Fukuoka}, \postcode{819-0395}, \country{Japan\vspace{0.12cm}}}}

\affil[7]{\orgdiv{ISAE-SUPAERO}, \city{Toulouse}, \postcode{31400}, \state{Occitanie}, \country{France}}

\abstract{
Complex systems are increasingly explored through simulation-driven engineering workflows that combine physics-based and empirical models with optimization and analytics. Despite their power, these workflows face two central obstacles: a high computational cost, since accurate exploration requires many expensive simulator runs; and limited transparency and reliability when decisions rely on opaque blackbox components.
We propose a workflow that addresses both challenges by training lightweight emulators on compact designs of experiments that (i) provide fast, low-latency approximations of expensive simulators, (ii) enable rigorous uncertainty quantification, and (iii) are adapted for global and local eXplainable Artificial Intelligence (XAI) analyses. 

\color{black}{
This paper formalizes this integration by providing a methodological bridge that handles continuous, categorical, and integer variables across diverse simulation domains, from engineering design to socio-environmental agent-based models. 
\color{black}{To validate surrogate explanations and analyze how they reflect simulator behavior, we introduce a comparative methodology incorporating active validation filters}, such as explanation consistency and aleatoric uncertainty quantification. \color{black}{Furthermore, this study provides a practical approach for coupling designs of experiments with emulator training}, offering practical guidelines for selecting surrogate architectures that balance computational efficiency with global and local interpretability.}

\color{black}{We illustrate our framework's applicability across two contrasting paradigms: a deterministic multidisciplinary design simulation of a hybrid-electric aircraft and a stochastic agent-based simulation of urban segregation.
Results show that the coupling of surrogate with XAI enables large-scale exploration in seconds, \color{black}{helps identify nonlinear interactions, and highlights potential design or policy levers}. By evaluating explanation consistency across multiple surrogate architectures, \color{black}{this workflow contributes to scalable and explainability-driven machine learning for complex-system exploration}.
}

}

\maketitle
\section{Introduction}

Modern engineering challenges in complex system modeling, ranging from electric aircraft design to urban infrastructure planning, are increasingly being tackled using simulation-driven approaches~\cite{fouda2022automated, michel2018}. 

A complex system is defined as one composed of many interconnected and interdependent elements whose nonlinear interactions give rise to emergent behaviors that cannot be directly inferred from the properties of the individual parts~\cite{kane2025ecosysml}. Such systems typically display emergent behaviors, self-organization, adaptation, and feedback loops, which make them challenging to capture holistically.

In the context of Agent-Based Modeling (ABM), simulations are often used to model ill-defined and context-sensitive dynamics where controlled experimentation is infeasible~\cite{angione2022}. 
Although Simulations of Agent-Based Models (SABM) reproduce bottom-up emergence effectively, their flexibility and opacity raise credibility and interpretability challenges, presenting barriers for decision support and policy alignment~\cite{blanco-volle2024}.
The challenges of credibility and interpretability in SABM are not unique to agent-based approaches; many stochastic blackbox simulators face similar issues~\cite{arnejo2025communicating}. To manage these, formal modeling, verification, and trade-space analysis are employed, not only to evaluate designs but also to shape possibilities in response to evolving objectives and constraints~\cite{chan2025goal}.  

For both modeling and understanding, we adopt an iterative “co-design” approach. 
\color{black}{While co-design can apply to other processes and contexts~\cite{zamenopoulos2018co}, here it denotes a twofold workflow. First, it aligns the diverse stakeholders and experts involved in complex system design that negotiate trade-offs across disciplines and establish a shared modeling ground~\cite{chan2025goal} and, second, it coordinates participants around shared experimental materials such as models, scenarios, and data, so that agent roles, assumptions, and artefacts are jointly developed, interrogated, and analyzed to support an informed decision-making process~\cite{drogoul2002multi}.}

Most of the processes considered require balancing of fidelity, computational cost, interpretability, and operational realism. Also, the considered stochastic simulators used to represent complex systems behave as opaque blackboxes due to their inherent complexity and nonlinear dynamics. Therefore, they can be difficult to understand, even when composed of simple and transparent rules. Consequently, there is a growing interest in embedding explainability, interpretability, and user-centric design~\cite{taillandier2019participatory}. As they expand in size and complexity, the demand for dedicated automated tools increases accordingly. 
Moreover, since these simulations are typically computationally expensive and can only be queried a limited number of times, careful selection of inputs through Design of Experiments (DoE) or active learning is essential to maximize their utility under tight computational budgets. Simulation toolkits go beyond purely analytical purposes, serving as interactive, participatory platforms where diverse stakeholders, including engineers, decision makers, and domain experts, collaborate on the sustainable design or analysis of complex systems~\cite{chan2023stroke,drogoul2002multi}. 
\color{black}{ However, while the individual components of this process (DoE, surrogate modeling, and explainable artificial intelligence) are well-established, their standard sequential application~\cite{fuchkina2018design, li2008exploration} suffers from critical blind spots. Specifically, current surrogate-assisted workflows face three major limitations. First, standard pipelines select surrogates based purely on predictive accuracy. However, multiple models can achieve identical accuracy while relying on completely divergent internal causal logic, meaning stakeholders might base critical decisions on algorithmic artifacts rather than true physics, a phenomenon known as the Rashōmon effect. Second, existing workflows often fail on stochastic simulators (like ABMs) because they cannot disentangle intrinsic simulator noise (aleatoric uncertainty) from surrogate approximation error (epistemic uncertainty), risking the interpretation of pure noise as a meaningful XAI signal. Third, conventional pipelines do not handle mixed variables and, when they do, generally force a one-hot or integer encoding of categorical variables. This artificially breaks feature coalitions, thereby violating the cooperative game-theory axioms of Shapley values~\cite{au2022grouped} and corrupting the resulting explanations. In contrast, our approach preserves this structural integrity by building upon our foundational SMT-EX tools~\cite{saves2024, robani2025, palar2024multi}. Motivated by the pressing need to resolve those blind spots, we propose an improved workflow where uncertainty quantification and XAI are not merely used as post-hoc analysis tools, but as active validation filters. By reducing the purely computational complexity of the simulations, this surrogate-assisted approach frees up resources to better handle the contextual complexity of conflicting objectives and domain expertise. By mathematically linking explanation consistency to predictive fidelity, our framework ensures that surrogate models are not just numerically accurate, but physically sound~\cite{garouani2025xstacking}.}

Actually, surrogate models like Gaussian processes, tree ensembles, or neural networks approximate expensive simulations using few high-fidelity runs. Applying XAI techniques to these surrogates reveals how input parameters influence global and individual outcomes~\cite{molnar2019,unlock_blackbox}, enabling AI that is both frugal and explainable.
This workflow is generic and can be applied to many applications, \color{black}{thereby supporting the exploration of simulations}, whether they represent an engineering design task or a complex system behavior analysis.  
We illustrate our framework with two contrasting case studies. The first one, from engineering and design, is based on a deterministic simulator whose equilibrium existence is guaranteed for admissible designs; the objective is the optimization of a given quantity of interest from the model. 
The second one consists of a stochastic multi-agent model and is based on a stochastic simulator whose equilibria depend on input values; the objective is the sensitivity analysis and mechanistic explanation of the system. Despite their different mathematical character, both problems are analyzed with the same methodology, \color{black}{illustrating the potential applicability of this work across different domains}. Indeed, complex systems models and simulations often share common underlying characteristics, making their exploration methodologically similar even when conceptually distinct~\cite{SaittaZucker2013}. 

Our contributions are structured along four main axes:
\begin{enumerate}
  \item[\textbf{(1)}] \textbf{Unified and explainable simulation workflow.} We introduce a methodological framework that integrates XAI techniques directly into surrogate-assisted simulation workflows. \color{black}{The simulation-centric framework aims to bridge simulators with diverse objectives through an iterative exploration loop, fostering a more transparent, interpretable, and efficient analysis} of stochastic, high-dimensional models through automated XAI.
 \item[\textbf{(2)}] \textbf{Local and global XAI complementarity.} We clarify how global and local explanations play complementary roles in surrogate-assisted simulation: global analyses (feature effects, sensitivity, uncertainty) reveal system-level relationships, while local attributions (instance-level importance scores) explain individual predictions and highlight actionable drivers. We introduce simple metrics to compare explanations across surrogate models and the original simulator, detect divergences, and guide active sampling and model refinement.

    \item[\textbf{(3)}] \textbf{Comparison and recommendation of surrogate-based explainability tools.}
    Building on surrogate models and XAI techniques that handle both continuous and categorical features, we propose a formal comparative methodology to assess surrogate adequacy. The methodology comprises global and local effect analyses and evaluates how consistently different surrogate models generate explanations.
    
  \item[\textbf{(4)}] \textbf{Evaluation and insights across domains.} We demonstrate the approach on two representative case studies: (\emph{i}) a multidisciplinary engineering design optimization problem, \color{black}{where surrogate models help identify system-level trade-offs and sensitivities}, and (\emph{ii}) an agent-based analysis and exploration scenario, \color{black}{where surrogate-based XAI helps explore behavioral drivers and emergent macro-patterns}. From these studies, we identify complementarities and discrepancies between local and global explanations and across surrogate models, and we provide guidelines and perspectives for future research and applications.
  
\end{enumerate}

The remainder of this paper is organized as follows. Section~\ref{sec:soa} reviews the state of the art on explainable AI for simulations, covering surrogate modeling approaches, explainability techniques, and the concepts underpinning their integration. Section~\ref{sec:methodo} presents the proposed unified methodology that combines simulation-based modeling, surrogate learning, and interpretability analysis within an iterative exploration workflow. Section~\ref{sec:dragon} applies this framework to a multidisciplinary aircraft design study, illustrating how interpretable surrogate models can accelerate exploration and uncover engineering insights in the $\texttt{DRAGON}$ hybrid-electric aircraft. Section~\ref{sec:abs} applied the workflow to a socio-technical domain through an agent-based urban segregation model, highlighting how surrogate-assisted explainability supports policy assessment and uncertainty analysis. 
Finally, Section~\ref{sec:conclu} concludes the paper and outlines future perspectives for embedding interpretability and transparent exploration principles into next-generation simulation workflows.  

\section{Related work on explainable AI for 
simulations}
\label{sec:soa}

Engineering simulators, such as computational fluid dynamics or multidisciplinary design analysis, approximate coupled governing equations to predict complex physical behavior, enabling detailed performance and reliability assessment in highly nonlinear settings~\cite{shihua2025bayesian,palar2024multi,audet2017introduction,dos2008model}. Complementarily, SABM adopts a bottom-up perspective: heterogeneous agents following local decision rules interact over spatial or networked environments to produce emergent, multi-scale dynamics applied to epidemic forecasting, urban planning, infrastructure management, flood mitigation, and more~\cite{macal2010tutorial,kaur2022multi,cueille2025assessing}.
Overall, modern simulators, whether physics-based or agent-based, face common challenges: high evaluation cost, mixed continuous and categorical inputs, strong nonlinearity, and limited transparency of causal mechanisms~\cite{drogoul2002multi,robani2025}.
These challenges have driven rapid progress in coupling surrogate modeling with XAI methods that can scale explanation and uncertainty quantification to expensive simulators~\cite{ML_industry,arnejo2025automatic}. 

In this section, we survey the literature on surrogate-based XAI for complex system simulations, with particular emphasis on methodological advances in surrogate modeling and explainability techniques.

\subsection{Surrogate models}

Surrogate models, also known as metamodels or emulators, provide fast approximations of expensive blackbox simulations. They are trained on a carefully designed set of high-fidelity runs, often generated with a DoE strategy such as Latin hypercube or Sobol’ sampling, and evaluated through the expensive opaque simulator whose mechanisms we seek to uncover~\cite{llacay2025,barry2024optimal}.
Building an emulator involves optimizing its parameters to best reproduce the simulator outputs, for example, by maximizing the likelihood of the observed data~\cite{petit2023parameter}. Once trained, the surrogate model can be evaluated quickly, making it suitable to replace the expensive simulator in optimization loops, sensitivity analyses, or uncertainty quantification studies~\cite{deleeuw2022,dubourg2011reliability}.
Common surrogate models include Gaussian process regression (Kriging), radial basis function networks, multivariate adaptive regression splines, neural networks, and tree ensembles; each offers different trade-offs in accuracy, interpretability, and computational cost~\cite{angione2022}. In principle, almost any machine learning model can serve as an emulator. However, surrogate models are typically designed with specific goals in mind, such as uncertainty quantification or interpretability. Also, the surrogate models should account for limited training data and be tailored to approximate expensive simulators efficiently.
Advanced strategies, such as multi-fidelity modeling, gradient-enhanced Kriging, and adaptive sampling via active learning, further improve surrogate precision while minimizing the number of expensive simulator calls~\cite{ravutla2025effects}. 
Some models come with intrinsic uncertainty quantification like Gaussian processes (posterior variance), random forest ensembles (prediction intervals), or transformer-based predictors (\textit{e.g.} TabPFN). Some models come with derivative-based methods (\textit{e.g.} polynomial chaos expansions or kernel-based surrogate models) that support coupled systems emulation through adjoint-based methods~\cite{saves2024,martins2022aerodynamic}.   
In the context of agent-based modeling, previous works~\cite{Lamperti2018} used machine learning surrogates to accelerate ABM parameter sweeps while maintaining interpretable mappings between inputs and outputs, enabling practical calibration workflows~\cite{angione2022}. 
By substantially reducing simulation costs, often by several orders of magnitude, surrogate modeling enables comprehensive exploration of high-dimensional design spaces, facilitates global sensitivity analyses, and supports robust, real-time decision-making. Across engineering and agent-based modeling domains, these emulators transform previously intractable scenarios into computationally efficient and reliable workflows.

\paragraph{Notations} 
Let the feature indices be $\{1,\ldots,n\}$ and let an input point $\textbf{x}$ be such that $\textbf{x}= [x_1, x_2, \ldots, x_n]^\top\in{\Omega}=\prod_{i=1}^n\Omega_i$ where each feature $i$ may be either continuous, integer ($x_i\in\{a_1,\dots,a_k\}$), or categorical ($x_i\in\{c_1,\dots,c_l\}$)~\cite{saves2023mixed}. An integer variable is ordered (\textit{e.g.} $ a_4<a_5< a_6$) while the values that a categorical variable can take, termed "levels", are generally considered without intrinsic order.
The quantity of interest $y$ is related to an input $\textbf{x}$ via a blackbox simulator $y = f(\textbf{x})  $, which we approximate by $\hat f(\textbf{x}) \approx y $ using the dataset of $m$ points denoted $\mathcal{D} = \{\mathcal{X}, \textbf{y}\}$. Here, $\mathcal{X}  = \{{x}_j^{(i)}\}_{j=1 \ldots n}^{i=1 \ldots m} $ is a chosen DoE (\textit{e.g.} a small Latin hypercube or Sobol' sampling),  $\textbf{y} = \{f(\textbf{x}^{(i)})\}_{i=1}^m$ are the corresponding simulator outputs~\cite{barry2024optimal} and 
$\hat{f}:\boldsymbol{\Omega}\to\mathbb{R}$ is the fast-to-evaluate emulator of $f$. 
Building an emulator consists in training $\hat f(\textbf{x}\mid\mathcal{D},\boldsymbol{\Theta})$  by learning optimal surrogate model hyperparameters  $\boldsymbol{\Theta}$, for example by maximizing the likelihood of the data $\mathcal{D}$ given hyperparameter $ \boldsymbol{\Theta}$~\cite{petit2023parameter}. Once trained, $\hat f(\textbf{x}) $ is a fast-to-evaluate surrogate that can replace costly simulations within optimization loops or uncertainty quantification studies~\cite{deleeuw2022,dubourg2011reliability}.

\paragraph{Model validation} To provide a unified assessment of surrogate model quality, we combine complementary metrics that are conceptually and practically linked: pointwise accuracy for numerical fidelity, classification reliability for feasibility or constraint predictions, uncertainty calibration for safe decision-making, and ranking consistency for interpretability.

The pointwise error is measured by the Root Mean Square Error (RMSE) on another dataset called the validation set $\mathcal{D_{\text{val}}}$ consisting of $m_{val}$ blackbox evaluations,
\begin{equation}\label{eq:rmse}
  \mathrm{RMSE} = \sqrt{\frac{1}{m_{val}}\sum_{i=1}^{m_{val}} \left(y_i - \hat f(x^{(i)})\right)^2},
\end{equation}
which quantifies the average magnitude of residuals and thereby serves as the baseline indicator of numerical precision~\cite{armstrong1992error}. 

For problems that involve binary outcomes (\textit{e.g.} feasible/infeasible, constraint satisfied / violated), we complement RMSE with a balanced classification metric, the Matthews Correlation Coefficient (MCC),
\begin{equation}\label{eq:mcc}
\mathrm{MCC}
= \frac{TP\cdot TN - FP\cdot FN}
       {\sqrt{(TP+FP)(TP+FN)(TN+FP)(TN+FN)}},
\end{equation}
where $TP,TN,FP, \ \text{and }FN$ denote the number of true/false positive/negative values. Because MCC synthesises the confusion matrix into a single correlation-like score that remains meaningful under class imbalance~\cite{chicco2021benefits}. 

Beyond point estimates and categorical accuracy, reliable uncertainty estimates are essential for Bayesian assessment or risk-aware design; we therefore evaluate variance calibration with the Predictive Variance Adequacy (PVA),
\begin{equation}\label{eq:pva}
\mathrm{PVA}
= \left|
\log\!\left(
\frac{1}{m_{\text{val}}}\sum_{i=1}^{m_{\text{val}}}  \frac{\left(y_i- \hat f(x^{(i)}) \right)^2}{\hat\sigma_i^2}
\right)
\right|,
\end{equation}
where \(\hat\sigma_i^2\) is the model’s predictive variance for sample \(i\); values near zero indicate well-calibrated uncertainty, while large deviations reflect either overconfidence or underconfidence and motivate recalibration or richer probabilistic models~\cite{bachoc2013cross}.

\subsection{Explainable AI}
XAI comprises methods and practices aimed at rendering machine-learning and simulation models transparent and their outputs intelligible to human users; its central aim is to clarify how inputs, model structure, and uncertainty lead to particular predictions or decisions. 
Important distinctions in XAI include intrinsic versus post-hoc approaches: intrinsic models are interpretable by design, whereas post-hoc methods generate explanations after training to illuminate blackbox behavior. Another key distinction is global versus local explanations: global methods characterize system-level behavior and dominant input-output relationships, while local methods provide instance-specific rationales useful for debugging, risk assessment, and actionable decision support~\cite{madsen2024interpretability}.
A representative technique for global explainability is Global Sensitivity Analysis (GSA), for example, Sobol’ indices, which quantify how input uncertainties contribute to output variability, thereby guiding feature prioritization and robustness assessment~\cite{sobol2001global,saltelli2008global}.
For local explainability, XAI generally conveys influence details with instance-based methods such as SHAP (SHapley Additive exPlanations), LIME (Local Interpretable Model-agnostic Explanations), or counterfactual explanations~\cite{knab2025lime,song2016shapley}. These model-agnostic tools assign credit to input features at the local prediction level~\cite{molnar2019}, producing human-readable explanations or actionable change suggestions.

XAI techniques, when integrated with surrogate modeling, provide an effective framework to address the interpretability challenges of complex systems. Surrogate models enable the efficient evaluations and systematic exploration required for explainability analyses~\cite{angione2022,audet2017introduction}. Yet, without suitable interpretability tools, surrogates may themselves act as blackboxes. The key objective remains to explain the behavior of the underlying complex system rather than the surrogate model itself, especially since many XAI methods focus on the surrogate without distinguishing between explanations based on raw data and those arising from the surrogate’s internal structure, thereby introducing another level of bias and uncertainty~\cite{xai_iai, bussemaker2024surrogate}.
On the visual aspects, surrogate models allow for computing Partial Dependence Plots (PDP) and Individual Conditional Expectation (ICE) quickly to reveal marginal and instance-level effects of variables on model predictions, even in the presence of complex interactions~\cite{goldstein2015peeking,friedman2001greedy}.  For instance, in~\cite{gavriilidis2023surrogate}, the authors propose a surrogate model framework that autonomously approximates a robotic agent’s behavior and uses feature attribution together with a natural-language generation pipeline to produce human-readable explanations, validated on simulated and real industrial data.

\paragraph{Notations} 
Let the feature indices $N=\{1,\ldots,n\}$ be partitioned into two disjoint subsets: \(A\) (variables of interest) and \(C\) (complement). Also, let \(\mathbf{x}=(\mathbf{x}_A,\mathbf{x}_C)\) with \(\mathbf{x}_A\in\Omega_A\) and \(\mathbf{x}_C\in\Omega_C\) and \(P_{X_C}\) denote the marginal law of \(X_C\). The partial dependence function of \(\hat f\) on \(\mathbf{x}_A\) is
\begin{equation}
\hat f_A(\mathbf{x}_A)\;=\;\mathbb{E}_{X_C\sim P_{X_C}}\!\big[\hat f(\mathbf{x}_A,X_C)\big]
\;=\;\int_{\Omega_C}\hat f(\mathbf{x}_A,x_C)\,dP_{X_C}(x_C).
\end{equation}
Evaluating $\hat{f}_A$ over $\Omega_A$ yields the PDP, which visualizes the average effect of $\mathbf{x}_A$ on $\hat{f}$. 
A simple global sensitivity metric for the \(i\)-th feature is the variance of its one-dimensional PDP: 
\begin{equation}
\sigma^2_{\mathrm{pd},i}
= 
\mathbb{V}\!\left[\hat{f}_i(x_i)\right].
\end{equation}
For continuous or ordinal variables, this variance quantifies the magnitude of influence of $x_i$ on the model output. 
However, for categorical variables, the notion of variance is not directly meaningful because categories lack a natural ordering. 
In such cases, a practical heuristic alternative sensitivity measure is the (normalized) range of the partial dependence values:
\begin{equation}
\label{eq:GSA_PDP_cat}
\sigma_{\mathrm{pd},i}
\;=\;
\frac{\max_{x_i\in \mathcal{X}}\hat f_i(x_i)-\min_{x_i\in \mathcal{X}}\hat f_i(x_i)}{4}.
\end{equation}

When interactions or non-stationarities are present, PDPs may obscure significant local variations. 
ICE plots address this issue by conditioning on each observed background instance $\mathbf{x}_C^{(j)}$ and displaying
\begin{equation}
\hat f_A^{(j)}(\mathbf{x}_A)
\;=\;
\hat f\big(\mathbf{x}_A,\mathbf{x}_C^{(j)}\big),\qquad j=1,\dots,m.
\end{equation}
The collection \(\{\hat f_A^{(j)}\}_{j=1}^m\) reveals the spread of individual responses around the PDP mean. Note that PDP and ICE rely on (marginal) input independence; strong input correlations can render these visualizations misleading, and call for conditional partial dependence, accumulated local effects     plots, or other dependence-aware alternatives~\cite{apley2020visualizing}. 
Still, most surrogate models employ space-filling DoEs and are based on the input-independence assumption~\cite{pasupathy2015modeling}, making PDP+ICE visualizations and sensitivity tools fitted for this context~\cite{lambert2025quantization}. Notwithstanding, a better practice is to test input independence beforehand on the dataset~\cite{arsac2025fast}.

The Shapley value, originally introduced in cooperative game theory~\cite{shapley1953stochastic}, provides an axiomatic rule to fairly attribute a total payoff to individual players. Let $N$ denote the set of players (features) and \(v:2^{N}\to\mathbb{R}\) be a value function that assigns a payoff to each coalition \(S\subseteq N\). The Shapley value of player \(i\in N\) is
\begin{equation}\label{eq:shapley}
\Phi_i(v)
\;=\;
\sum_{S\subseteq N\setminus\{i\}}
\frac{|S|!\,(n-|S|-1)!}{n!}\,
\bigl(v(S\cup\{i\})-v(S)\bigr),
\end{equation}
which equals the average marginal contribution of \(i\) over all possible coalitions and satisfies the classical Shapley axioms (efficiency, symmetry, dummy, additivity).

The SHAP framework~\cite{lundberg2017unified} adapts this principle to model explanations by treating features as players and the model prediction as the payoff. For a predictive model \(\hat f\) and a specific instance \(\mathbf{x}\), define the instance-specific value function
\begin{equation}
v_{\mathbf{x}}(S) \;=\; \mathbb{E}\!\bigl[\hat f(\mathbf{X}) \mid \mathbf{X}_S=\mathbf{x}_S\bigr],
\qquad S\subseteq N,
\end{equation}
where the expectation is taken with respect to a chosen background distribution for \(\mathbf{X}\) (the choice of which must be reported in practice). The SHAP value (feature attribution) of the feature \(i\), for instance \(\mathbf{x}\) is obtained by applying \eqref{eq:shapley} to \(v_{\mathbf{x}}\):
\begin{equation}\label{eq:shap}
\phi_i(\mathbf{x})
\;=\;
\sum_{S\subseteq N\setminus\{i\}}
\frac{|S|!\,(n-|S|-1)!}{n!}\,
\Bigl(v_{\mathbf{x}}(S\cup\{i\})-v_{\mathbf{x}}(S)\Bigr).
\end{equation}
By construction, the SHAP values provide the additive decomposition
\begin{equation}
\hat f(\mathbf{x}) \;=\; v_{\mathbf{x}}(\varnothing) \;+\; \sum_{i=1}^n \phi_i(\mathbf{x}),
\end{equation}
where \(v_{\mathbf{x}}(\varnothing)\) is the prediction of the baseline (reference). To capture pairwise interactions, SHAP introduces interaction values \(\phi_{i,j}(\mathbf{x})\) for \(i\neq j\). One convenient symmetric definition is
\begin{equation}\label{eq:shap-interaction}
\phi_{i,j}(\mathbf{x})
\;=\;
\frac{1}{2}\sum_{S\subseteq N\setminus\{i,j\}}
\frac{|S|!\,(n-|S|-2)!}{(n-1)!}\,
\Bigl(v_{\mathbf{x}}(S\cup\{i,j\}) - v_{\mathbf{x}}(S\cup\{i\}) - v_{\mathbf{x}}(S\cup\{j\}) + v_{\mathbf{x}}(S)\Bigr),
\end{equation}
which yields \(\phi_{i,j}(\mathbf{x})=\phi_{j,i}(\mathbf{x})\) and decomposes each feature attribution as
\begin{equation}
\phi_i(\mathbf{x}) \;=\; \sum_{j=1}^n \phi_{i,j}(\mathbf{x}).
\end{equation}
The main (non-interaction) effect of feature \(i\) is then commonly defined by
\begin{equation}
\phi_{i,i}(\mathbf{x})
\;=\;
\phi_i(\mathbf{x}) - \sum_{j\neq i}\phi_{i,j}(\mathbf{x}),
\end{equation}
so that the full decomposition across main effects and pairwise interactions is coherent and sums to the model prediction difference \(\hat f(\mathbf{x})-v_{\mathbf{x}}(\varnothing)\).

Practical computation of the expectations \(v_{\mathbf{x}}(S)\) requires a choice of background distribution and typically relies on approximations (sampling, model-specific fast algorithms, or conditional expectation estimators) due to the combinatorial number of subsets. These implementation choices affect both the numerical values and the interpretation of SHAP attributions and should therefore be reported alongside any SHAP-based analysis~\cite{lundberg2017unified,idrissi2021developments,demange2023shapley}.

The Harsanyi set of allocations~\cite{vasilev2002harsanyi} extends the concept of Shapley values to encompass various forms, including the weighted and proportional versions. This broader framework enables the development of diverse importance measures, each reflecting distinct normative principles. Consequently, it provides a versatile foundation for crafting importance metrics tailored to a wide array of objectives.

A commonly used global importance score for feature \(i\) is the expected absolute SHAP value:
\begin{equation}   
S_i \;=\; \mathbb{E}\bigl[\;|\phi_i|\;\bigr]
\;\approx\;
\frac1m \sum_{j=1}^m \bigl|\phi_i^{(j)}\bigr|,
\end{equation}
which aggregates instance-level contributions into a summary ranking of feature importance across the dataset \cite{molnar2019,lee2023shap}.  Still, to derive GSA metrics with Shapley values, some other techniques such as Shapley effects~\cite{owen2014sobol} or proportional marginal effect~\cite{herin2024proportional} are more appropriate, as they show more desirable properties.

\paragraph{Model validation}

Because surrogate models are often used as interpretable metamodels of a more complex simulator, we check whether the surrogate preserves the feature-importance structure using a ranking-agreement metric based on SHAP values: the Normalized Discounted Cumulative Gain (NDCG). Let \(\mathrm{rel}_{i,j}\) be the absolute SHAP magnitude of sample \(i\) for the \(j\)-th ranked feature ; then, for two models $\mathcal{M}_1$ and $\mathcal{M}_2$:
\begin{equation}\label{eq:dcg}
\mathrm{DCG}_i=\sum_{j=1}^n \frac{\mathrm{rel}_{i,j}}{\log_2(j+1)},\qquad
\mathrm{NDCG_{\mathcal{M}_1,\mathcal{M}_2}}=\frac{1}{m}\sum_{i=1}^m \frac{\mathrm{DCG(\mathcal{M}_1)}_i}{\mathrm{DCG(\mathcal{M}_2)}_i},
\end{equation}
where values close to 1 indicate that the surrogate preserves the reference model’s explanation ranking~\cite{burges2005learning}.
\color{black}
This pointwise NDCG metric is asymmetric as its value depends on which model is designated as the "true" reference~\cite{levy2026analyzing}. Therefore, we rely on a symmetrized explanation consistency score to establish an undirected pairwise similarity between models: $\rho_{NDCG} = (\mathrm{NDCG}_{\mathcal{M}_1,\mathcal{M}_2} + \mathrm{NDCG}_{\mathcal{M}_2,\mathcal{M}_1}) / 2$.

The necessity of such an explanation consistency metric arises directly from the \textit{Rashōmon effect} on model multiplicity in machine learning. Actually,   multiple surrogate models with entirely different mathematical foundations (\textit{e.g.}, tree ensembles vs. kernel methods) can achieve identically high predictive accuracy (low RMSE / high MCC) while relying on completely different internal feature attributions. Therefore, standard accuracy metrics alone are insufficient to guarantee that a surrogate's explanation reflects the true underlying simulator's mechanics rather than its own algorithmic artifacts~\cite{semenova2022existence}.

Ideally, explanation fidelity would be validated by computing SHAP values directly on the high-fidelity simulator. However, since SHAP requires thousands of evaluations per instance, doing so on an expensive simulator is computationally intractable, which is precisely the motivation for using surrogates in the first place~\cite{palar2026interpretable}.

\color{black}

\subsection{\color{black}{Uncertainty in simulation-based workflows}}
\color{black}{Simulators and their surrogate models are necessarily imperfect representations of reality. A simulator introduces assumptions, simplifications, uncertain parameters, and numerical approximations, while a surrogate adds another layer of approximation by learning the simulator's input--output relationship from a finite dataset~\cite{ferson2004summary}. Consequently, uncertainty should be identified, characterized,
and, where appropriate, propagated throughout the entire modeling workflow.}

\color{black}{An important distinction between controlled design factors, whose values can be selected by the designer, and uncontrolled noise factors, which represent environmental, manufacturing, or operating variations that cannot be directly controlled during system operation, has been coined in~\cite{taguchi1986introduction}. In this regard, controllable design factors are selected to achieve the desired performance while minimizing sensitivity to uncontrollable noise factors~\cite{chen1996procedure}. Within the context of simulation-based Design Under Uncertainty (DUU), uncertainties arising from unknown model parameters, discrepancies between the simulator and reality, finite computer experiments, and surrogate approximations must also be considered, as they can affect both performance predictions and the resulting design decisions~\cite{kennedy2001bayesian,apley2006understanding}.
Accordingly, validation and uncertainty quantification provide a rigorous foundation for assessing how faithfully a simulator represents the system of interest, how uncertainties and approximation errors propagate through the
modeling workflow, and how reliably its predictions can support subsequent analyses and decisions~\cite{roy2011comprehensive}. Another useful complementary classification distinguishes aleatory uncertainty, which stems from the inherent randomness or variability of the system, from epistemic uncertainty, which arises from incomplete knowledge, limited data, or imperfect modeling assumptions~\cite{hullermeier2021aleatoric, der2009aleatory}.
}

\color{black}{The various types of uncertainty may coexist and interact throughout a simulation-based workflow. Because comprehensively quantifying every source is generally impractical, the relevant uncertainties should be selected and modeled according to their expected influence on the quantities of interest and the intended use of the predictions.} 
\color{black}{In modern engineering practice, propagating these aleatory and epistemic uncertainties through expensive simulators often requires reliability-based design analysis and optimization or multi-fidelity approaches. In these contexts, active learning and surrogate models (such as Gaussian processes) have become essential for maintaining computational tractability while managing model discrepancy~\cite{espoeys2024overview}.}

\subsection{Positioning of our work}

\paragraph{Surrogate Modeling and Automated Exploration}
Modern complex systems, whether physical or socio-technical, increasingly rely on high-fidelity simulations to capture emergent, multi-scale phenomena. However, as these models grow in complexity, the computational cost of parameter exploration and uncertainty analysis becomes prohibitive. To overcome this, surrogate modeling has become a foundational methodological tool. In engineering, frameworks for Multidisciplinary Design Analysis and Optimization (MDAO) mostly rely on Gaussian processes and polynomial chaos expansions to bypass expensive deterministic solvers~\cite{forrester2008engineering, martins2022aerodynamic}. Similarly, in the socio-technical domain, machine learning emulators are increasingly deployed to approximate the stochastic and dynamic behaviors of ABM for policy testing~\cite{Lamperti2018, angione2022}.
{While alternative paradigms, such as symbolic regression, attempt to derive intrinsically interpretable equations directly from data~\cite{brunton2016discovering, cranmer2023interpretable}, they frequently fail to scale to the high-dimensional, highly nonlinear spaces characteristic of modern MDAO and ABM.} Recently, Automated Machine Learning (AutoML) pipelines have sought to streamline this surrogate training process~\cite{he2021automl}; however, this automation often exacerbates the "blackbox" problem, prioritizing predictive accuracy over mechanistic transparency.

\paragraph{From Global Sensitivity to Explainable AI}
To bridge the gap between predictive accuracy and mechanistic understanding, the scientific community is shifting its analytical focus. Historically, Global Sensitivity Analysis (GSA), such as variance-based Sobol' indices, has been the gold standard for identifying influential inputs~\cite{saltelli2008global}. \color{black}{However, standard Sobol' indices rely on the assumption of independent inputs and can become mathematically ill-posed or prohibitively expensive to compute when dealing with correlated features or mixed continuous-categorical variable spaces~\cite{pianet2026sensitivity}.} However, GSA provides limited insight into local, instance-level dynamics or directional effects. Consequently, XAI is rapidly being adopted~\cite{roscher2020explainable}. Techniques such as SHAP~\cite{lundberg2017unified} and PDP are being integrated into engineering pipelines to ensure data-driven surrogates respect underlying physical constraints~\cite{henckaerts2022stakes}, offering deeper causal insights than traditional statistical correlations~\cite{iooss2022}. 

\color{black}{Nevertheless, XAI should not be regarded as a replacement for conventional GSA, but as a complementary set of tools addressing different interpretive questions~\cite{palar2026interpretable}. GSA quantifies how much each input and its interactions
contribute to the overall output variability, whereas XAI can reveal the shape and direction of input--output relationships and explain the factors driving individual predictions. For example, PDPs describe average directional trends,
while local SHAP values indicate how each input contributes to a particular model response (noting that SHAP values can be globally averaged to obtain a GSA metric). Hence, GSA identifies which variables are globally influential,  whereas XAI helps clarify how, in which direction, and under which configurations their effects arise~\cite{saves2026model}. Their combined use therefore provides a more comprehensive interpretation of surrogate-based simulation models. In this paper, we demonstrate this complementarity through two case studies by combining global importance analysis with local and directional explanations to reveal both the dominant
input variables and how their effects vary across individual system configurations. }
\color{black}{Moreover, conventional GSA methodologies often prove inadequate for models characterized by discontinuous and non-uniform input distributions, especially when dealing with input parameter dependencies~\cite{pianet2026sensitivity}.
}

\paragraph{The "Fidelity Gap" and Out-of-Distribution Risks in Current Explainable AI}
Despite the enthusiasm for XAI, a critical vulnerability remains largely unaddressed in the current literature: the "fidelity gap." Leading researchers have warned against the blind application of post-hoc XAI to blackbox models, noting that methods like SHAP and LIME can be highly sensitive to perturbations and may produce unreliable attributions~\cite{slack2020fooling, rudin2019stop}. {Furthermore, mainstream XAI implementations often rely on feature permutation or marginalization. In highly coupled engineering systems, this forces the surrogate model to evaluate unrealistic or out-of-distribution data combinations, leading to physically impossible feature attributions~\cite{hooker2021unrestricted, kumar2020problems}.} In the context of simulation science, this creates a profound epistemic risk: post-hoc methods frequently explain the mathematical artifacts or extrapolation errors of the surrogate model rather than the true underlying physics or agent behaviors.

In summary, surrogate models speed up the extraction of data-driven insights, support quantitative analysis of complex systems under uncertainty, and make it feasible to evaluate computationally expensive metrics that characterize model behavior. Paired with XAI, surrogate models therefore enhance the interpretability and explanatory reach of simulation studies. Accordingly, our workflow replaces the true simulator \(f\) with its surrogate \(\hat{f}\) whenever we compute XAI measures or uncertainty estimates. While the canonical definitions of the relevant quantities are given with respect to \(f\), direct evaluation on \(f\) is typically intractable; instead, we compute their surrogate approximations via \(\hat{f}\). This substitution introduces an unavoidable \emph{epistemic uncertainty} that captures the unknown discrepancy between \(\hat{f}\) and \(f\) and which we account for in our analyses~\cite{menz2021variance}.

Also, surrogate-based XAI operates at two complementary levels.
At the local level, instance-wise analyses probe specific configurations to check reliability (\textit{e.g.}, exceedance probabilities), identify tipping points, and assess decision-relevant risk. These analyses typically require targeted resampling in regions of interest.
At the global level, aggregated analyses quantify dominant drivers, nonlinear effects, and trade-offs across the input domain using variance- or dependency-based sensitivity indices~\cite{palar2023enhancing}; global analyses generally rely on space-filling DoEs for broad coverage.

Both levels demand rigorous validation: predictive fidelity, calibration of uncertainty estimates, and stability of explanations should be verified, preferably across multiple surrogate families or via selective high-fidelity reevaluation. We use the four measures introduced earlier jointly: RMSE to compare regressors, MCC to flag classification weaknesses, PVA to validate uncertainty-driven acquisition, and NDCG to check that surrogate-derived explanations remain faithful to the high-fidelity model.

\color{black}
\paragraph{The Explainable Surrogate Framework}
In terms of state-of-the-art contributions to simulation-driven decision-making, the intersection of XAI and surrogate modeling represents a critical frontier. As comprehensively surveyed in~\cite{palar2026interpretable}, traditional surrogate workflows predominantly prioritize computational efficiency, treating the resulting emulators as opaque, black-box artifacts. However, to support trustworthy, high-stakes decision-making, it is imperative to map specific XAI techniques directly onto the various stages of the simulation lifecycle. Moving beyond treating XAI as a passive, post-hoc visualization task, recent literature synthesizes an "explainability toolbox" consisting of four operational layers: (i) global parameter screening to bound the feasible space, (ii) visualization of non-linear interactions to map physical trends, (iii) instance-specific local diagnostics to detect failure boundaries, and (iv) logic- and argumentation-based reasoning to bridge the cultural gap with traditional engineering verification. Including these tools in the training loop makes explainability a key part of the design from the start. This allows users to not only run faster simulations, but also get trustworthy answers from complex systems~\cite{palar2026interpretable}.

Still, recent developments have sought to democratize the use of surrogate models for domain researchers. Notably, tools like AutoEmulate~\cite{jersakova2026autoemulate} offer fully automated, end-to-end PyTorch pipelines that abstract away the complexities of model selection, hyperparameter tuning, and active learning behind a high-level API. While such frameworks excel at providing accessible emulators for non-experts, they fundamentally act as automated orchestrators rather than specialized analytical engines. In contrast, our proposed workflow, built heavily upon the foundations of SMT or scikit-learn~\cite {saves2024,pedregosa2011}, prioritizes theoretical rigor and deep diagnostic granularity over pure automation. 
While automated frameworks treat the emulator as the final product, our approach treats the surrogate as an intermediary object designed to extract insights from. By providing native support for complex hierarchical mixed variables input spaces without one-hot encoding, and explicitly disentangling aleatoric from epistemic uncertainty to mitigate the Rashōmon effect, our framework offers the advanced verification instruments that domain experts require~\cite{shimizu2025agror}. Ultimately, these two paradigms are highly complementary: automated tools streamline the initial construction of the emulator, whereas our framework provides the rigorous, specialized XAI mechanisms necessary to validate and interpret these models in high-stakes, real-world decision-making context.
\color{black}

The primary objective of this work is to leverage these explainability metrics to make complex simulation models more interpretable, trustworthy, and actionable within design exploration processes, while using surrogate models to reduce the computational cost of large-scale exploration. A detailed methodology for doing so is described in the following section.
\section{Framework}
\label{sec:methodo}

\color{black}{
\color{black}{This workflow embeds uncertainty quantification and XAI not merely as post-hoc visualization tools, but as active mathematical validation filters. First, it introduces cross-model meta-analyses to use explanation consistency as a secondary filter, mitigating the Rashōmon effect by ensuring the selected surrogate aligns with true underlying physics. Second, it integrates block-subsampling techniques to compute robust 95\% confidence intervals for XAI attributions, preventing stakeholders from misinterpreting aleatoric simulator noise as causal physical signals.}
To operationalize the objectives outlined above, we propose a structured methodological pipeline. Crucially, this workflow serves to bridge the epistemological and communicative gaps between domain experts (\textit{Thematicians}), software developers (\textit{Computer Scientists}), and modeling experts (\textit{Modelers})~\cite{drogoul2002multi}. It builds upon the principle that simulation and inference are complementary iterative processes rather than distinct stages of inquiry~\cite{varenne2009simulation, varenne2013modeliser}. The framework is executed cyclically, allowing stakeholders to develop a progressively more informed perspective on the system at hand through a dialectical process~\cite{fauriat2022discourse}.}

\begin{figure}[H]
\centering
\includegraphics[width=0.95\linewidth]{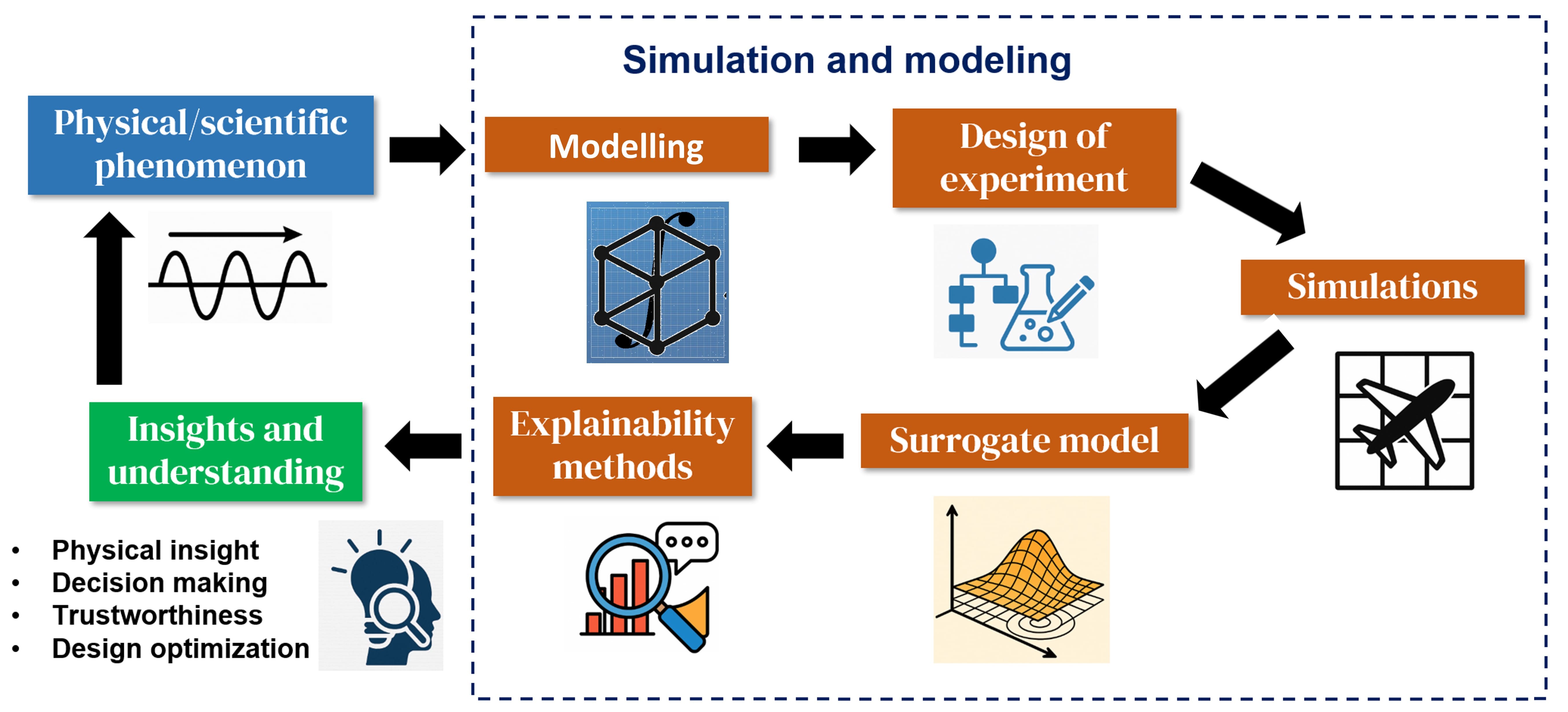}
\caption{Workflow for interpretable surrogate modeling in complex system exploration, illustrating the integration of domain assumptions, computational limitations, and adaptive learning pathways.}
\label{fig:graph_abs}
\end{figure}

Figure~\ref{fig:graph_abs} summarizes the five key stages of the proposed workflow, colored in orange, highlighting the different processes at hand for each phase:
\begin{enumerate}
\item \textbf{Conceptual and physical modeling:} Driven by the \textbf{Modeler}, this stage formalizes mechanistic assumptions, physical boundaries, and domain knowledge. The goal is to ensure coherence with the underlying real-world phenomenon while defining the "plausible" space for exploration. It needs to be validated and can be progressively enriched to model more complex physics and increase the fidelity of the model with respect to the scientific object one seeks to understand. 
\item \textbf{Design of experiments:} Guided by the \textbf{Thematician}, this step selects an initial, structured sampling plan (\textit{e.g.}, Latin hypercube, Sobol' sequence). It actively integrates expert-guided sampling to focus computational resources on regions of high epistemic importance rather than mathematically possible but practically absurd extremes. The produced dataset can be sequentially enriched and validated to converge to a given objective within a given budget via adaptive sampling and active learning. This stage encapsulates the choice of exploring either locally or globally, by defining trust regions and target objectives.
\item \textbf{High-fidelity simulation:} Developed and optimized by the \textbf{Computer Scientist}, this stage translates the conceptual model into an operational simulator. It explicitly introduces \textit{implementation limitations and approximations} (driven by computational cost, memory, and algorithmic constraints). The simulator is then evaluated at the previously selected input points, recording outputs and measurement noise. The simulation comes with its uncertainty and hypothesis. Simulators generally require calibration or optimization of internal parameters, and need to be validated to be as reliable as possible.
\item \textbf{Surrogate modeling:} Executed by the \textbf{Data Scientist}, this step trains interpretable, fast-to-evaluate surrogate models (\textit{e.g.}, Gaussian Processes, ensemble trees, or sparse neural networks) that encode the prior information and quantify the posterior uncertainty generated by the simulator. This step is often the most crucial, as the structure of the chosen surrogate model may strongly bias the subsequent analysis, yet its validation may inform practitioners about the underlying data structure. 
\item \textbf{Explainability and sensitivity analysis:} Closing the exploration loop, global (\textit{e.g.}, Sobol’, HSIC, PDP) and local (\textit{e.g.}, SHAP, LIME) techniques are applied. The extracted main effects, interactions, and stability metrics are fed back to the \textbf{Thematician} and \textbf{Data Scientist} to validate them against their domain assumptions, enabling active dimensionality reduction or adaptive sampling in the next iteration if necessary to gather even more knowledge thanks to a more informed perspective.
\end{enumerate}

Priors should, whenever practicable, be elicited from expert domain knowledge (\textit{e.g.}, physical bounds, expert judgement) to serve as guardrails for the computational system. Interpretable, computationally inexpensive surrogates are then introduced to transform costly, high-fidelity simulators into interactive co-design and decision-making tools during system exploration~\cite{fauriat2022discourse}. 

Our methodological stance is explicitly epistemic: models and simulations are treated as approximate, \textit{a priori} constructs representing a necessary compromise between the Thematician's theoretical vision and the Computer Scientist's implementation realities~\cite{varenne2009simulation}. Explicit identification and quantification of prior assumptions, modeling errors, and approximation gaps are therefore central to the workflow. Moreover, modeling choices encode prior beliefs about mechanisms, plausible parameter ranges, and relevant scales, and introduce structural approximations that require rigorous assessment.

We distinguish four sources of uncertainty and error that the methodology explicitly addresses across the different expert domains~\cite{kennedy2001bayesian}: 
First, \textbf{hidden variables uncertainty}, arising from intrinsic stochasticity within the system or simulator, for instance, the seed parameter in a pseudo-random socio-technical simulation. 
Second, \textbf{data-related uncertainty}, due to limited, noisy, or incomplete observations and unobserved factors. 
Third, \textbf{model discrepancy}, corresponding to the mismatch between the simulator’s structural assumptions and the target phenomenon. This often arises during the translation phase, where a Computer Scientist might replace a complex model with a simplified object to maintain tractability. 
Finally, \textbf{surrogate approximation error} refers to the residual gap between the Data Scientist's surrogate predictions and the high-fidelity simulator outputs. For instance, a surrogate model can perform poorly in terms of absolute accuracy but remain useful and interpretable for mapping coarse stability boundaries.

Framing these uncertainties within a Bayesian perspective enables coherent expression of \textit{a priori} expert knowledge and systematic belief updating as new simulation data are acquired~\cite{ramsey1926truth}. Even deterministic simulations reflect choices of formalism, scale, and representation in their structures, leading to underlying decisions that affect how surrogate-based explanations are later constructed, presented, and understood by stakeholders~\cite{varenne2013modeliser}. Consequently, it is essential to verify that surrogate approximation error is acceptably small for the intended exploration task and that the chosen DoE provides sufficient coverage for the planned validation~\cite{wilhelm2024hacking}.

This framework supports two main research objectives:  
(i) How can surrogate models provide a local, instance-wise understanding of complex models to validate thematic assumptions?  
(ii) What global insights can they reveal about model dynamics, structural behavior, and implementation artifacts?  

In summary, the proposed methodology transforms costly, high-fidelity simulations into transparent supports for reasoning and decision-support during system exploration, keeping domain experts informed~\cite{bussemaker2024surrogate, fauriat2022discourse}. 
In the next sections, we demonstrate through experiments how these tools enable systematic exploration of simulation behaviors and user-oriented analysis. Section~\ref{sec:dragon} presents the application of our workflow to goal-oriented design of a complex system, and Section~\ref{sec:abs} presents the application of our framework to policy assessment and robust analysis of a complex system.

\section{Engineering Insights from Multidisciplinary Design Exploration}  
\label{sec:dragon}

\begin{figure}[H]
\centering
\hspace{-0.25cm}
\includegraphics[width=0.95\linewidth]
{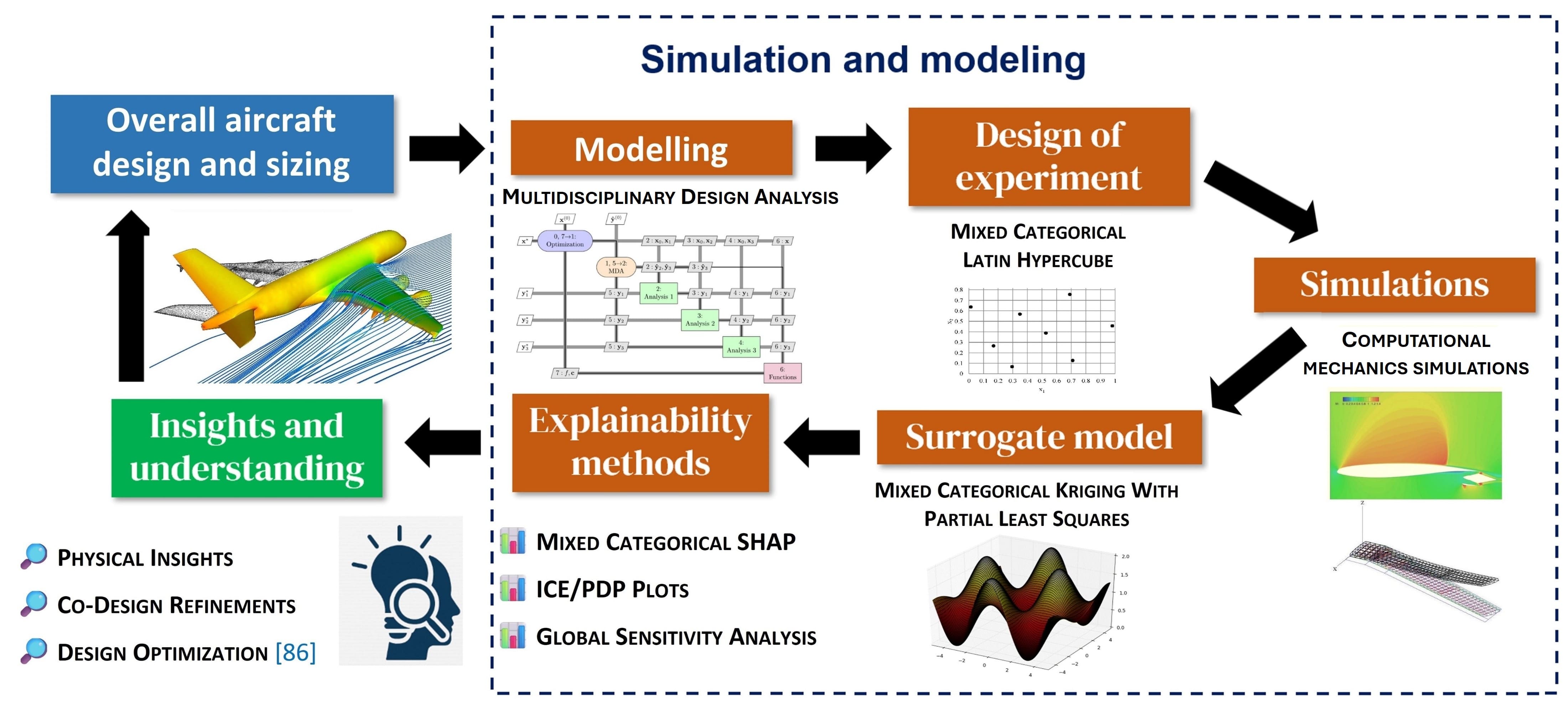}
\caption{Workflow for interpretable surrogate models for a design-oriented simulation.}
\label{fig:graph_abs_dragon}
\end{figure}

Multidisciplinary design analysis and optimization integrates models from different engineering domains to study trade-offs and interactions in complex systems across potentially conflicting engineering disciplines. 
Figure~\ref{fig:graph_abs_dragon} integrates surrogate models into a surrogate-assisted exploration process. This is an instance of the workflow introduced in Fig.~\ref{fig:graph_abs}, specifically instantiated for this engineering problem. It begins with the definition of the multidisciplinary design analysis integrated model and the generation of a static DoE across the input space of the aircraft configuration, encompassing key geometric and operational parameters, including mixed categorical variables.

Each configuration is evaluated using a high-fidelity simulator to obtain target outputs, including fuel mass. A surrogate model is then trained on this dataset to emulate the costly simulator with substantially reduced computational cost. XAI techniques are applied to the surrogate to quantify the influence of individual design variables on the objectives and constraints outputs. These insights inform stakeholders during the design exploration process, guiding refinements and highlighting trade-offs between competing objectives and constraints, such as energy performance and passenger capacity. This section demonstrates how our simulation workflow can be applied to the sequential design of a complex system, integrating explainability and extending our previous work~\cite{satria2024design} by handling mixed-variable input spaces, whereas most existing approaches are restricted to continuous variables.

First, Section~\ref{subsec:mdao} introduces the physical engineering design of interest and describes its main geometric and operational design variables together with the equilibrium target. After that, the workflow unfolds as described in Fig.~\ref{fig:graph_abs_dragon}. Section~\ref{subsec:dragon_mdao} then describes the modeling approach for coupling multidisciplinary submodels in a multidisciplinary design frame; Section~\ref{subec:dragon_doe} introduces the DoE for the mixed continuous/categorical space and the aggregated output metrics used to train surrogates; Section~\ref{subec:dragon_simulator} details the simulator implementation and numerical settings; Section~\ref{subec:dragon_surrogates} presents the surrogate model, and Section~\ref{subsec:resdragon} reports the global and local XAI analyses, highlighting key interactions and trade-offs.

\subsection{Overall aircraft design and sizing of a Hybrid-Electric aircraft}
\label{subsec:mdao}

\begin{figure}[H]
\begin{centering}
\includegraphics[height=4.8cm]{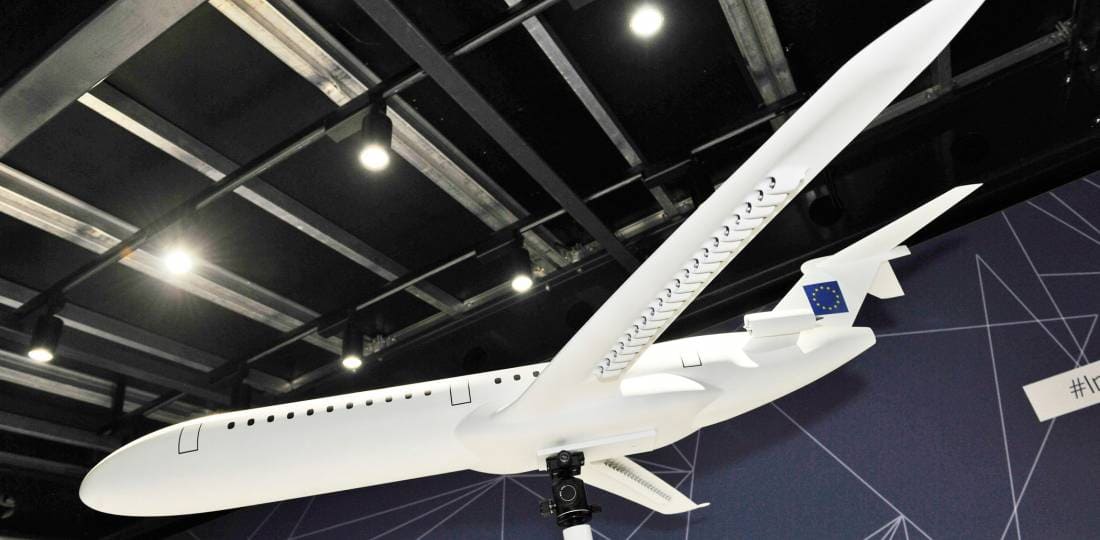}
   \caption{\texttt{DRAGON} aircraft mock-up~\cite{ridel2021dragon}.}
        \label{SMO_Dragon2020}  
\end{centering}
\end{figure}

For the first demonstration, we applied various explainability methods to the "\texttt{DRAGON}" configuration (Distributed fans Research Aircraft with electric Generators by ONERA) to investigate how the aircraft's input variables influence fuel mass from the stakeholder point of view. 
This aircraft concept study is funded by the European Commission, aiming to design the next generation of long-range aircraft that reduces climate impact as measured via radiative forcing~\cite{joulia2024first}. 
The \texttt{DRAGON} concept represents a distributed hybrid electric propulsion architecture, which combines multiple electrically driven fans with conventional gas turbines~\cite{ridel2021dragon} whose mock-up is given in Fig.~\ref{SMO_Dragon2020}. Assessing such configurations demands high-fidelity coupling of electrical engineering, aerodynamic performance, thermal-structural integrity, fluid-structure interaction, flight mechanics, and system-level failure modes~\cite{planas2025multi} and is generally modeled with an extended design structure matrix~\cite{lambe2012extensions}. 

This distributed electric propulsion concept, proposed by ONERA in 2019, was developed to support the European CleanSky2 initiative, which targets a 30\% reduction in CO\textsubscript{2} emissions by 2025. The \texttt{DRAGON} aircraft was designed to carry 150 passengers over a range of 2750 nautical miles at a cruise speed of Mach 0.78. While the configuration has previously been studied in the context of optimization for minimizing fuel mass~\cite{saves2024high}, there has been no in-depth investigation into the relationships between the design variables and fuel mass, an aspect that this study aims to address using explainability analysis~\cite{satria2024design}.

Because DRAGON is part of multidisciplinary and international research, its design process already involves extensive coupling among aerodynamics, propulsion, and structural disciplines to integrate multiple expert assessments and technology assumptions. It is a so-called \textit{coupled system} whose objective is to find a multidisciplinary trade-off corresponding to a functioning equilibrated aircraft that respects a certain static margin while being able to fulfill its mission at a minimal speed and over a minimal range~\cite{sobieszczanski1990sensitivity}.

This problem features 10 continuous and 2 categorical input variables, in which the quantity of interest is fuel mass. 
The 10 continuous variables correspond to the geometrical and flight parameters of the aircraft as shown in Tab.~\ref{tab:DRAGON_variable}. 
The first and second categorical variables correspond to the electric architecture with 17 levels and the turboshaft layout level with 2 levels, respectively.
The architecture variable and the turboshaft layout are defined in Tab.~\ref{SMO_tab:dragon_archi1} and Tab.~\ref{SMO_tab:dragon_archi2}, respectively.

\begin{table*}[h!]
\centering
\begin{tabular}{|l|c|c|l|}
\hline
\textbf{Function/Variable} & \textbf{Nature}  & \textbf{Range} \\
\hline
Fan operating pressure ratio & cont &  [1.05, 1.3] \\
Wing aspect ratio & cont &  [8, 12] \\
Angle for swept wing & cont &  [15, 40]$\ \left(^\circ\right)$ \\
Wing taper ratio & cont & [0.2, 0.5] \\
Horizontal Tail (HT) aspect ratio & cont &  [3, 6] \\
Angle for swept HT & cont &  [20, 40]$ \ \left(^\circ\right)$ \\
HT taper ratio & cont &  [0.3, 0.5] \\
Takeoff Field Length (TOFL) for sizing & cont &  [1800, 2500] (m) \\
Top of climb vertical speed for sizing & cont &  [300, 800] (ft/min) \\
Start of climb slope angle & cont &  [0.075, 0.15] (rad) \\
Architecture (levels) & cat  & \{1, 2, ..., 16, 17\} \\
Turboshaft layout (levels) & cat  & \{1, 2\} \\
\hline
\end{tabular}
\caption{Design variables for fuel mass analysis.}
\label{tab:DRAGON_variable}
\end{table*}

\begin{table}[ht]
\centering
 \caption{Definition of the architecture variable and its 17 associated levels.}
\begin{tabular}{ccc}
\hline
  \textbf{Architecture number} & number of motors & number of cores and generators\\
  \hline
  \textbf{1} & 8 &2 \\
  \textbf{2} & 12 &  2\\
  \textbf{3} & 16 &  2\\
  \textbf{4} &20 &2 \\
  \textbf{5} & 24 &  2\\
  \textbf{6} & 28 &  2\\
  \textbf{7} &32 & 2\\
  \textbf{8} & 36  & 2\\
  \textbf{9} & 40 &  2\\
  \textbf{10} & 8   & 4\\
  \textbf{11} & 16  & 4\\
  \textbf{12} & 24  & 4\\
  \textbf{13} & 32  & 4\\
  \textbf{14} & 40  & 4\\
  \textbf{15} & 12  & 6\\
  \textbf{16} & 24  & 6\\
  \textbf{17} & 36  & 6\\
\hline
\end{tabular}
\label{SMO_tab:dragon_archi1}
\end{table}

\begin{table}[!h]
\centering
 \caption{Definition of the turboshaft layout variable and its 2 associated levels.}
\begin{tabular}{cccccc}
\hline
  \textbf{Layout} & position & y ratio & tail & VT aspect ratio & VT taper ratio\\
  \hline 
  \textbf{1} & under wing &0.25 & without T-tail& 1.8 & 0.3 \\
  \textbf{2} & behind & 0.34 & with T-tail& 1.2 & 0.85\\
\hline
\end{tabular}
\label{SMO_tab:dragon_archi2}
\end{table}     

\subsection{Multidisciplinary design analysis}
\label{subsec:dragon_mdao}

Designing complex engineered systems, such as an aircraft, requires the coordinated work of several tightly coupled disciplines (aerodynamics, structures, propulsion, controls, and so on). A decision made in one subsystem, for example, a change in wing geometry, often triggers responses in others (drag, loads, engine demand), producing nonlinear interactions across multiple feedback loops. To tame this complexity, \emph{eXtended Design Structure Matrix} (XDSM) diagrams provide a compact, yet expressive, graphical language for showing modules, data exchanges, solver loops, and execution order; these diagrams make it easy to spot which components must be solved iteratively and which can be treated more loosely~\cite{lambe2012extensions}. Put simply, an XDSM is a schematic of the problem structure that helps teams see where coupling and iteration are required.
These schematic descriptions can be converted into executable \emph{Multidisciplinary Design Analysis and Optimization} workflows in frameworks such as OpenMDAO. In that setting, each discipline is wrapped as a component, drivers coordinate solvers and optimizers, and system-level derivatives or surrogate models are propagated to support efficient gradient-based exploration of the design space~\cite{gray2019openmdao}. In plain terms, XDSM gives us the blueprint, and MDAO provides the assembly line to run experiments and optimizations automatically. Used together, they improve traceability and reproducibility, guide the choice of solvers and surrogate strategies, and enable faster, more disciplined trade studies before committing significant resources to costly high-fidelity simulations.

For modeling the electric architecture, it proves more effective to use two integer variables rather than a single categorical one to capture the design options. However, this choice significantly enlarges the space of possible architectures beyond the original 17 configurations. There are, nevertheless, two limitations that restrict this expanded domain.
The first limitation is about how components are wired together because building a certifiable electrical topology is not simple. For example, deciding how to connect eight motors to six generators needs careful engineering. The second limitation concerns the distributed propulsion system and its many propellers. For safety and stability reasons, not every possible electrical connection is acceptable, and some arrangements would create too many potential failure modes.
To keep the optimization manageable and avoid overcomplicating the constraint structure, the model ultimately reverts to a single categorical variable that encodes only the feasible and certified architectures.

\subsection{Design of experiments and aggregated output metrics}
\label{subec:dragon_doe}

We generate a 1000-point DoE using a Latin Hypercube Sampling (LHS) that we adapted for mixed-variable spaces~\cite{saves2024}. While we have also developed adaptations of other sampling methods, such as Sobol' sequences for mixed hierarchical variables~\cite{bussemaker2025system},   we consistently employ LHS in this study. This choice is due to its low-discrepancy properties, which, although slightly less optimal for variance approximation and uncertainty quantification compared to Sobol' sequences, offer a practical balance between efficiency and simplicity~\cite{l2000variance}. 

In addition, we will consider only the fuel mass response as the target output, which is the total mass of fuel consumed by the aircraft during the mission as predicted by the simulator.

The computational time was around 2 to 5 minutes by \textit{datum} for a total of 2 days using an Intel® Xeon® CPU E5-2650 v4 @ 2.20
GHz core and 128 GB of memory with a Broadwell-generation processor front-end
and a compute node with a peak power of 844 GFlops.

\subsection{Implementation of the simulator}
\label{subec:dragon_simulator}
In our workflow, the abstract XDSM diagrams are converted into executable multidisciplinary design analysis simulations that constitute the computational backbone of the DRAGON multidisciplinary model. These simulations couple aerodynamic, structural, propulsion, and electrical submodels inside mission-analysis and sizing loops so that reported quantities correspond to a physically consistent operating point obtained by iterating across disciplines until mutual agreement is reached.

We implement these coupled analyses using the open-source Future Aircraft Sizing Tool with Overall Aircraft Design (FAST-OAD)~\cite{david2021fast}, which we have extended with domain expertise and recent modules for hybrid-electric and distributed propulsion systems. FAST-OAD is built on the OpenMDAO infrastructure and follows a modular component paradigm: each discipline is encapsulated as a component, data transfers and solver couplings are made explicit, and the problem is defined by the set of design variables, constraints, driver(s), and chosen discipline submodels. This modularity makes the tool flexible in practice: it can run isolated single-point evaluations for sampling, execute nested solver sequences to obtain fully coupled convergence across subsystems, or serve as the computationally expensive blackbox evaluator inside surrogate-assisted global optimization loops. Because individual discipline modules (geometry, aerodynamics, mass estimation, propulsion and power electronics, mission integration, electrical load-flow, and so on) can be replaced or enriched with user-provided submodels, the framework supports targeted fidelity control and straightforward incorporation of component-specific behavior across the design space.

Because the DRAGON test case exhibits strong nonlinear interactions between its aerodynamic, structural, and electrical subsystems, the standard fixed-point coupling between disciplines used in baseline FAST-OAD was adapted to better handle equilibrium between disciplines. This ensures that key feedback loops, such as those linking propulsion power, mass, and energy balance, converge more reliably under tight coupling conditions. Moreover, we use low-fidelity disciplines by relying on equation-based approximations or very coarse space discretizations. 
Still, a full multidisciplinary evaluation, including powertrain and load-flow analyses, sizing loops, and mission fuel computation, requires approximately 2 to 5 minutes per run. Despite being much faster than high-fidelity disciplinary chains because of the many approximations the simulator relies on, this computational cost makes the multidisciplinary design analysis model effectively a nonlinear blackbox for large-scale optimization.

To address this, the following section trains surrogate models on FAST-OAD outputs to efficiently approximate the equilibrium solutions of the coupled disciplines. These metamodels accelerate global exploration and enable XAI techniques to reveal key sensitivities and design drivers, while remaining grounded in the underlying multidisciplinary physics.
\subsection{\color{black}{Surrogate modeling }}
\label{subec:dragon_surrogates}
\color{black}{
Although the relationships between categorical levels can theoretically be analyzed by extracting correlation matrices directly from Kriging hyperparameters \cite{saves2023mixed}, this intrinsic approach scales poorly to high-dimensional categorical spaces. Specifically, encoding the 17 architectural levels through continuous relaxation introduces a substantial number of hyperparameters, making the Maximum Likelihood Estimation (MLE) highly vulnerable to severe local minima on restricted training sets (e.g., $N=300$). Conversely, while dimensionality reduction techniques like Kriging with Partial Least Squares (KPLS) ensure convergence for high-dimensional mixed-variable models \cite{saves2024high}, projecting inputs into a latent space distorts the original feature topology. This inherently compromises the analytical capacity to directly interpret pairwise categorical correlations. Such fundamental limitations in intrinsic surrogate explainability strongly justify our use of robust, post-hoc XAI methods, such as SHAP and PDP, to consistently evaluate complex architectural choices.
Consequently, we rely on a reduced-order model. We employed KPLS using three latent variables, a squared exponential kernel for continuous variables, and a continuous relaxation kernel for the categorical variables. Trained on the 300-point dataset, this KPLS surrogate achieves a low RMSE of just 1.42\% over the 700-point holdout set, confirming it is highly accurate and appropriate for reliable XAI knowledge extraction. All associated data are available online\footnote{\url{https://github.com/ANR-MIMICO/SMPT_XAI}}.}
\subsection{Explainable methods, analyses, and insights extraction}
\label{subsec:resdragon}

For this problem, we use SHAP and PDP complemented with ICE for explainability analysis. We also performed Global Sensitivity Analysis (GSA) using SHAP and PDP to investigate how input variables influence the output. One thing worth mentioning here is that the use of SHAP and PDP allows the quantification of the impact of categorical variables, which is not possible if we use variance-based decomposition methods.
Based on the extension of KPLS to categorical variables in~\cite{saves2024high} and the extension of PDP, SHAP, and GSA to such variables in~\cite{robani2025}, we can represent the importance of the categorical variables, their coupling effects, and even their influence in the form of a boxplot.

\subsubsection{Global sensitivity analysis}

\begin{figure}[H]
\centering
\includegraphics[width=0.95\linewidth]
{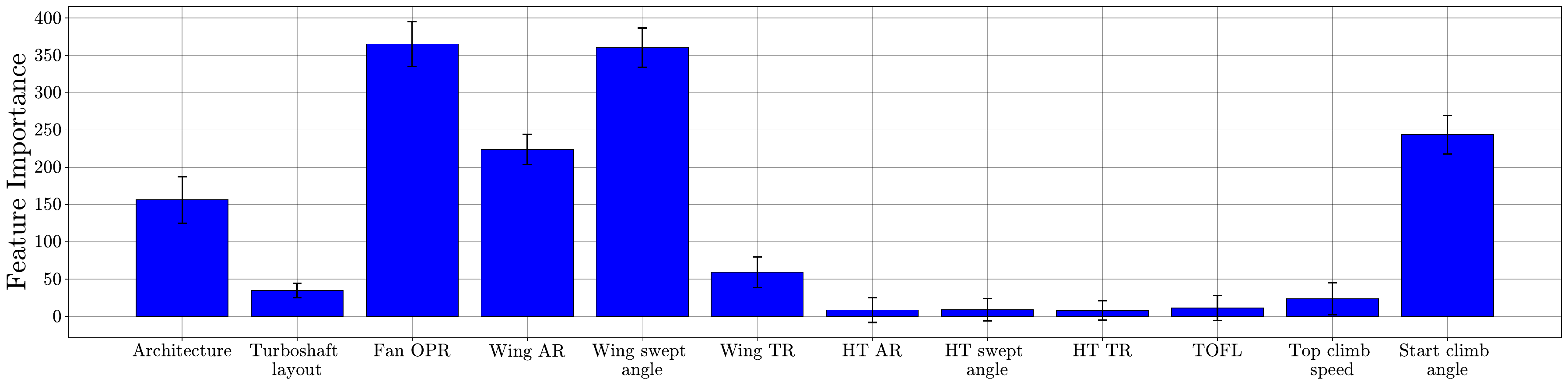}
    \caption{GSA results from partial dependence functions for the \texttt{DRAGON} problem.}
\label{fig:DRAGON_PD_FI}
\end{figure}

\begin{figure}[H]
\centering
\includegraphics[width=0.95\linewidth]
{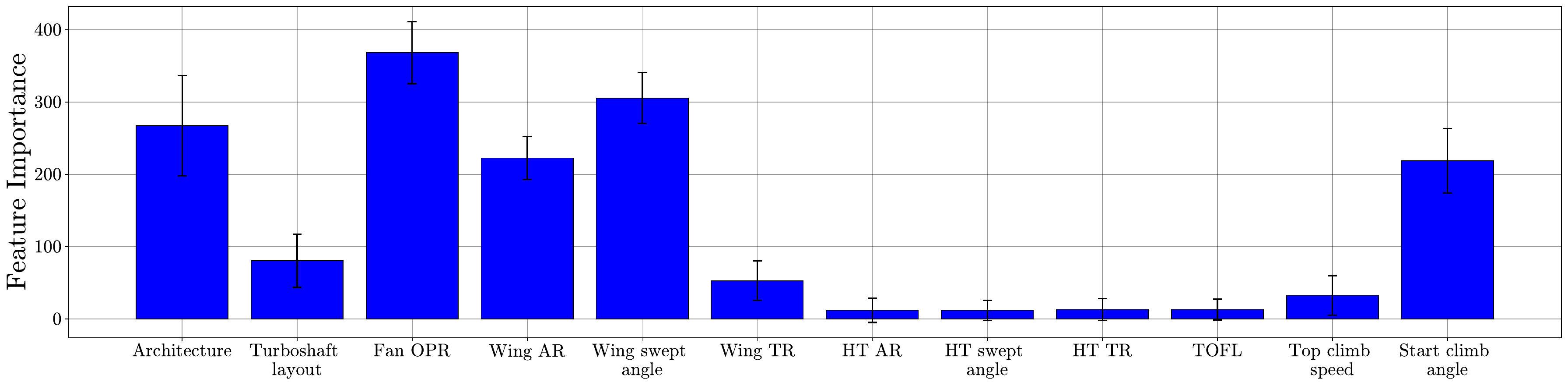}
\caption{GSA results from the averaged SHAP values for the \texttt{DRAGON} problem.}
\label{fig:DRAGON_SHAP_FI}
\end{figure}

Global sensitivity analysis identifies the fan Overall Pressure Ratio (OPR) and the wing sweep angle as the most influential variables; their partial dependence and SHAP results are shown in Fig.~\ref{fig:DRAGON_PD_FI} and Fig.~\ref{fig:DRAGON_SHAP_FI}, respectively. The four variables, Horizontal Tail Aspect Ratio (HT AR), HT sweep angle, HT Taper Ratio (HT TR), and the sizing Take-Off Field Length (TOFL), contribute to less than 1\% of the model’s explained importance on the output and can therefore be considered effectively non-influential. Notably, the choice of electric architecture ranks comparably in importance to the initial climb slope angle and the wing aspect ratio. The turboshaft layout and wing taper ratio exhibit moderate effects, though their influence is smaller than that of the architecture choice. Because the underlying model is a fast equation-based proxy, the impact of turboshaft layout is likely underestimated; this moderate effect therefore advocates for follow-up studies with higher-fidelity models~\cite{habermann2023study}.

Clearly, the importance of categorical variables (primarily architecture) is notably higher when using averaged SHAP values compared to the partial dependence values. The difference arises from the way the importance of categorical variables is formulated in PDP (as shown in Eq.~\ref{eq:GSA_PDP_cat}), which relies on heuristics and differs from the treatment of continuous variables. In contrast, SHAP applies the same equation for both categorical and continuous variables, ensuring a consistent approach across feature types. Given this consistency and the theoretical foundations of SHAP, it is generally better to trust the averaged SHAP values over PDP when assessing feature importance (especially when involving categorical variables), as they provide a more reliable and unified measure of an input’s contribution to the model’s predictions~\cite{lee2023shap}.

\color{black}{To ensure the robustness of the global sensitivity analysis, 95\% confidence intervals for both partial dependence and SHAP indices were computed via bootstrapping. Specifically, 100 random subsamples were drawn with replacement from the 300-point training dataset (with duplicate points removed, yielding approximately 63.2\% of the original dataset size per subsample). For each bootstrap iteration, a new surrogate model was trained, and the XAI metrics were re-evaluated. Due to the high computational cost of Exact SHAP (approximately 1.2 hours per run), the efficient Kernel SHAP approximation was employed for the bootstrap simulations, enabling feasible execution while maintaining accurate attribution estimates~\cite{herman2017salib}.}

\subsubsection{Local explainability visualization}

\begin{figure}[H]
\centering
\hspace{-0.25cm}
\includegraphics[width=1\linewidth]
{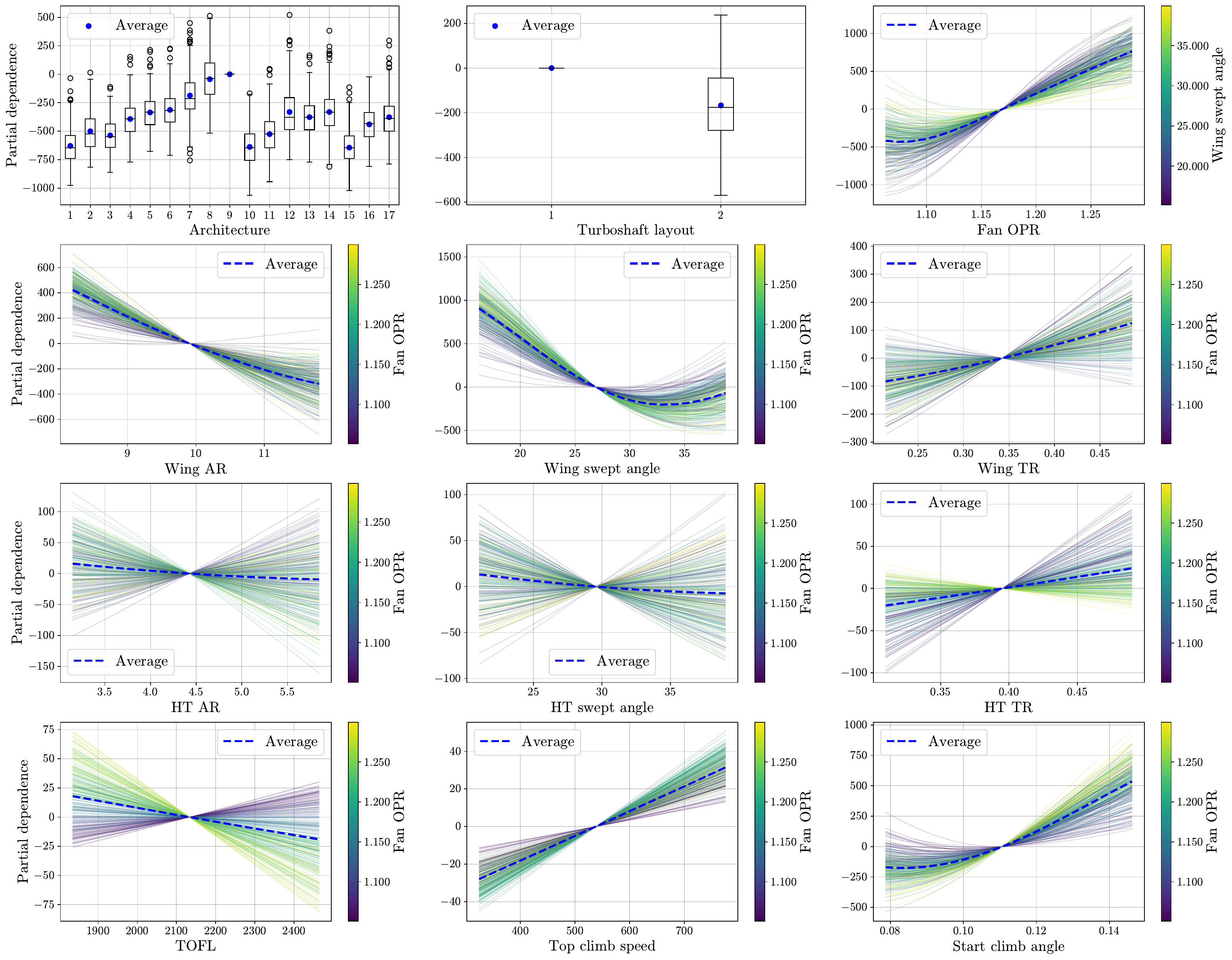}
\caption{PDP and ICE curves for the \texttt{DRAGON} problem.}
\label{fig:DRAGON_PD_ICE_PLOT}
\end{figure}

\begin{figure}[H]
\centering
\hspace{-0.25cm}
\includegraphics[width=1\linewidth]
{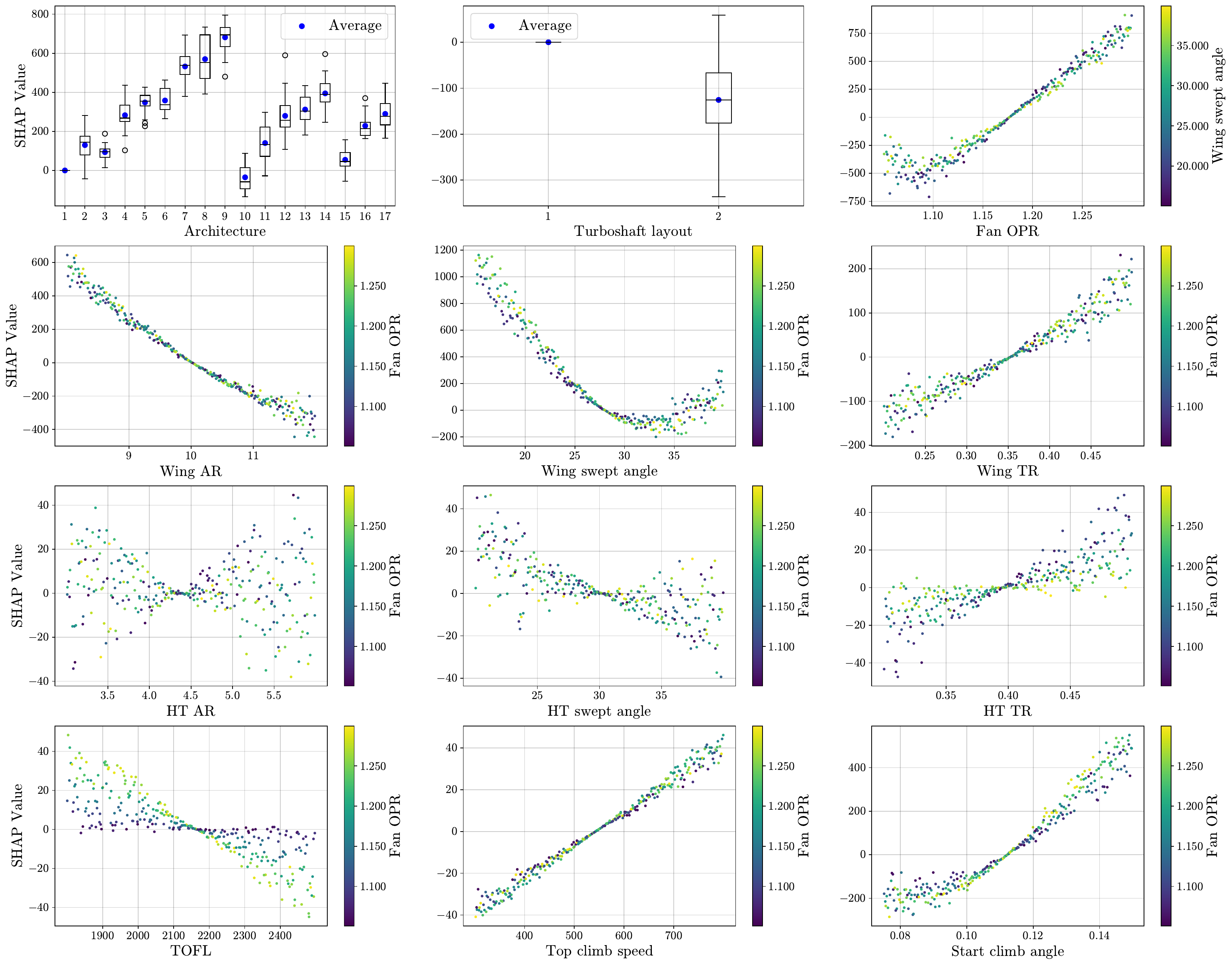}
\caption{SHAP dependence plot for the \texttt{DRAGON} problem.}
\label{fig:DRAGON_SHAP_dependence_plots}
\end{figure}

The PDPs and SHAP dependence plots (Fig.~\ref{fig:DRAGON_PD_ICE_PLOT} and Fig.~\ref{fig:DRAGON_SHAP_dependence_plots}) summarize each variable’s influence and help reveal interactions. 
\color{black}{In our implementation within SMT, PDP are mean-centered (anchored) to facilitate a direct visual comparison with SHAP values. Since SHAP inherently represents the marginal additive deviation from the expected baseline output, its values are naturally centered around zero. Anchoring the PDP aligns its vertical scale with that of SHAP, providing a cohesive visual assessment of feature effects~\cite{robani2025}.}

For visualization, ICE curves are color-coded by Fan OPR (the most influential variable), except the Fan OPR plot itself, which is colored by wing sweep angle; SHAP dependence points are colored by the secondary variable of interest. The wing sweep angle exhibits a clear non-monotonic effect on fuel mass: fuel mass decreases with sweep up to about 30° and then increases beyond that. By contrast, several continuous variables show largely monotonic effects (\textit{e.g.}, higher Fan OPR tends to increase fuel mass, while larger wing aspect ratio tends to decrease it), although some nonlinearities remain (for example, fuel mass rises increasingly steeply with higher start climb angles). PDPs for variables with negligible influence sometimes display both increasing and decreasing trends; these are likely surrogate artefacts and should not be overinterpreted. 

Interactions are visible in both ICE and SHAP plots. For Fan OPR, the ICE curves show a stronger effect when the wing swept angle is small (steeper gradients), and conversely, the influence of the wing swept angle on fuel burn is amplified at high Fan OPR, therefore highlighting a bidirectional interaction. The start climb angle, wing aspect ratio, and other influential variables similarly have larger effects when Fan OPR is high. For variables with minimal global influence (\textit{e.g.}, HT-AR), apparent interactions with Fan OPR are unreliable and should be treated cautiously. SHAP dependence plots show the same interaction patterns, but they can be visually subtler; scattered SHAP values for dominant variables (wing swept angle, Fan OPR) nonetheless corroborate the presence of interactions.

The impact of categorical variables is worth looking at with SHAP, as outlined before, because it is more reliable than ICE in this setting. Using architecture $1$ as the reference in Fig.~\ref{fig:DRAGON_SHAP_dependence_plots}, we observe that architecture 10 generally leads to a smaller fuel mass than the first architecture. On the other hand, some architectures (most notably, architectures 8 and 9) lead to higher fuel mass than the reference architecture. Therefore, ignoring the presence of constraints for now, architectures 10 and 15 seem to be the best choice when the goal is to minimize fuel mass. Similarly, one can also see that the second turboshap layout generally leads to reduced fuel mass compared to the first layout, although its impact on fuel mass is not as significant as the choice of architecture. 

\begin{figure}[H]
\centering
\hspace{-0.25cm}
\includegraphics[width=0.65\linewidth]
{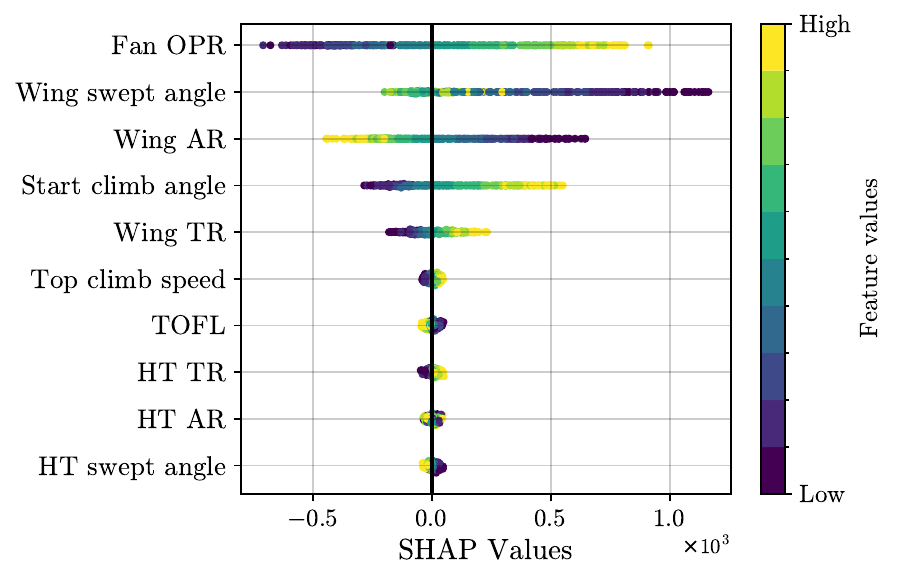}
\caption{SHAP beeswarm plot for the \texttt{DRAGON} problem continuous variables.}
\label{fig:DRAGON_SHAP_PLOT}
\end{figure}

The SHAP summary plot, as shown in Fig.~\ref{fig:DRAGON_SHAP_PLOT}, gives a direct interpretation of all continuous variables at the same time. It is worth noting that it is not useful to include categorical variables in the SHAP dependence plot since there is no natural ordering of the categorical variables. 
Both PDP+ICE and SHAP generally provide similar insights. However, in our opinion, PDP+ICE offers a clearer view of how interactions with other variables influence the way a specific variable affects fuel mass.  In this regard, PDP+ICE provides a faster and clearer intuition by showing direct relationships between input variables and the output. PDP alone is intuitive enough, but it does not supply information on the impact of interactions; this role is fulfilled by ICE.

\begin{figure}[H]
\centering
\hspace{-0.25cm}
\includegraphics[width=1\linewidth]
{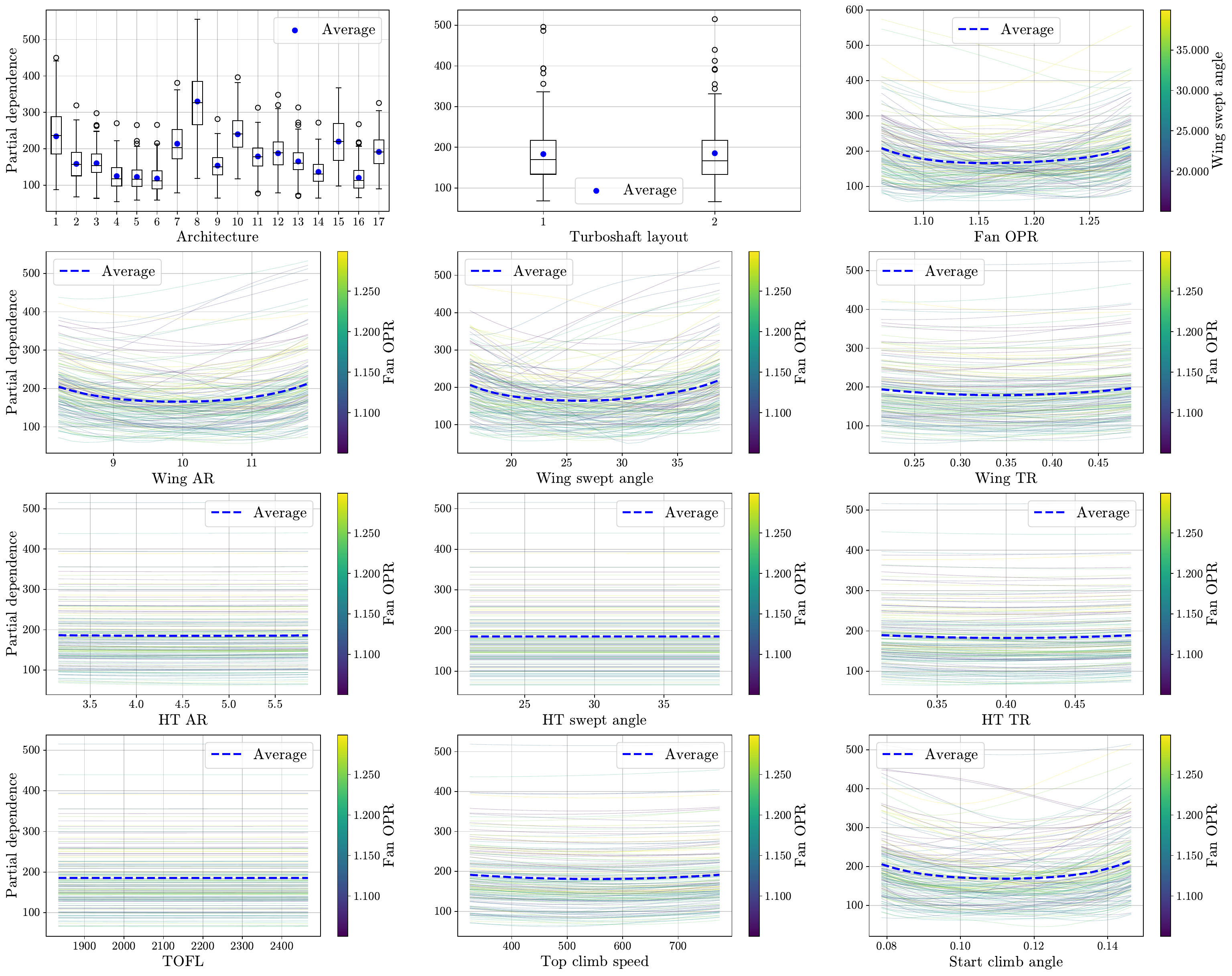}
\caption{PDP and ICE curves for the \texttt{DRAGON} problem output predicted variance.}
\label{fig:DRAGON_PD_ICE_PLOT_var}
\end{figure}

\color{black}{To further investigate epistemic uncertainty, we exploit the Kriging model's capability to predict not only the mean response but also the pointwise predicted variance. First, we compute SHAP dependence plots directly on the predicted standard deviation as shown in Appendix~\ref{app:companal}  on Fig.~\ref{fig:DRAGON_SHAP_PLOT_var}. However, because SHAP values are inherently mean-centered (anchored), this plot is highly noisy and creates a misleading visual artifact: it gives the false impression that uncertainty systematically increases as one moves away from the center of the input space. 
To resolve this anchoring issue and observe the true spatial distribution of uncertainty, we instead rely on unanchored PDP and ICE curves in Fig.~\ref{fig:DRAGON_PD_ICE_PLOT_var}. These unanchored plots reveal a much clearer narrative. For variables previously identified via GSA as having low impact, the epistemic uncertainty is mostly constant and independent. Conversely, the most influential variables exhibit varying, non-independent uncertainty, which primarily peaks at the boundaries of the input space, which is a natural consequence of data scarcity, the curse of dimensionality, and the Kriging surrogate model. Finally, analyzing the categorical effects reveals that Architectures 1, 8, 10, 15, and 17 suffer from notably higher uncertainty. This indicates that these specific architectural configurations are underrepresented in the current 300-point dataset, highlighting them as prime candidates for targeted data acquisition in future active learning iterations.}

\color{black}{
Through our explainability analysis, we explored the impact of key design variables, such as electric generator sizing, battery energy density, distribution of fan thrust, and wing structural stiffness, on fuel mass. By integrating global sensitivity measures and local explainers, we revealed how multi-physics interactions drive design trade-offs, such as the nonlinear coupling between electrical subsystem sizing and aerodynamic drag penalties. Hence, we transform FAST-OAD from a mere predictive tool into a transparent decision-support environment, enabling transparent analysis across aerodynamicists, electrical engineers, structural analysts, and system architects to collaboratively refine the \texttt{DRAGON} architecture.
This first demonstration provides a valuable test case to showcase how metamodels, when coupled with explainability techniques, can reveal the influence of design parameters on the objective of an engineering problem. In this context, it offers insights into how various aircraft parameters impact fuel consumption. When combined with previous optimization studies, this analysis not only validates the key drivers identified through optimization but also enhances understanding by highlighting why certain design choices lead to reduced fuel mass. Such explainability supports more informed decision-making in the early stages of aircraft design.}
Surrogate models combined with XAI are highly useful for early-stage design exploration: surrogates enable rapid exploration of large design spaces, while XAI reveals which inputs, interactions, and architecture choices most strongly influence performance and why. In the DRAGON framework, these insights are actionable: they guide architecture screening, prioritize optimization effort, and indicate where to raise discipline fidelity in FAST-OAD. To capture the stochastic and operational behaviors that static multidisciplinary design analysis and surrogate evaluations cannot represent, the next section introduces socio-technical dynamic simulations based on ABM. Integrating ABM with our surrogate/XAI stack extends interpretability from isolated, pointwise trade-offs to time-dependent, system-level dynamic behaviors, reveals emergent constraints driven by operational processes, and enables richer, context-aware design analysis. Accordingly, we conduct a second, more systematic case study that leverages XAI to quantify how dynamic operational behaviors alter earlier surrogate-based conclusions and to identify which feature attributions remain robust under surrogate model approximations and operational uncertainty.

\section{Agent-based model exploration for decision support}
\label{sec:abs}

Agent-based modeling is a powerful tool for simulating complex systems based on autonomous agents that interact with one another and with their environment. These models are particularly important for studying emergent phenomena, for example, when local interactions give rise to global behaviors that are often difficult to capture with traditional analytical methods~\cite{debosscher2023}. ABM provides a powerful framework for the emulation of complex systems made up of autonomous agents interacting in a dynamic environment, such as epidemiology, economics, or urban planning, offering a bottom-up approach to understand emergent phenomena~\cite{bremer2016,dunne2022complex}. They are fundamentally stochastic to emulate unknown individual behavior and to better capture the epistemic uncertainty over the underlying phenomena.
\begin{figure}[H]
\centering
\hspace{-0.25cm}
\includegraphics[width=0.95\linewidth]
{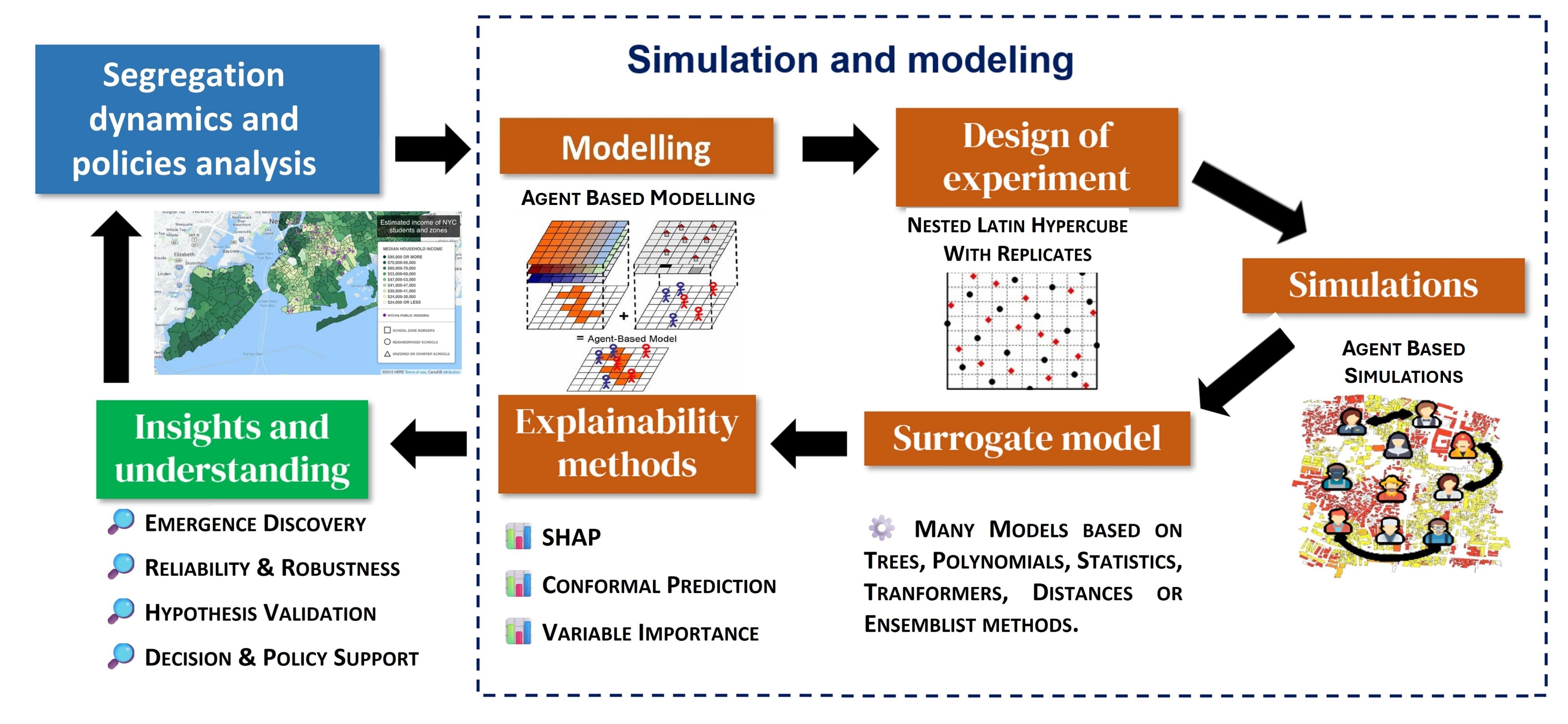}
\caption{Workflow for interpretable surrogate models for an analysis-oriented simulation.}
\label{fig:graph_abs_schelling}
\end{figure}

Figure~\ref{fig:graph_abs_schelling} illustrates our workflow for integrating interpretable surrogate models into an assessment-oriented simulation pipeline: this is an instance of the workflow already introduced in Fig.~\ref{fig:graph_abs} but adapted to this agent-based problem. 
Our workflow starts with the definition of the ABM and the generation of a DoE over the input space of the Schelling segregation model parameters. 
We do not claim that the Schelling model is realistic; it is meant as a proof-of-concept of how our workflow can be instantiated on a socio-technical problem. 
Furthermore, each point in the DoE is evaluated through five independent replications to account for the stochastic variability inherent in socio-technical ABM simulations (\textit{e.g.}, to estimate pointwise mean and variance)~\cite{carmona2024decomposing}.
The choice of five repetitions was driven by computational budget but could also be selected adaptively, driven by variance-based stopping rules using uncertainty-quantification or sequential design methods~\cite{menz2021variance}. 
Each simulation is evaluated using a high-fidelity simulator to obtain two chosen target outputs. 
Various surrogate models are then trained and compared on this dataset to emulate the costly simulator with substantially reduced computational cost. Surrogates principally approximate the simulator’s input-output mapping under uncertainty. Nevertheless, surrogate predictions and derived explanations remain approximations of the true simulator and must be validated accordingly. XAI techniques are applied to the surrogate models to quantify global and local input effects and to generate human-interpretable attributions that support the design process. Overall, the workflow supports four related objectives: (i) support decision-making and assess intervention impacts via systematic what-if exploration; (ii) improve model reliability and robustness through replication and sensitivity analysis; (iii) hypothesis testing and theory validation by linking micro-rules to macro-outcomes; and (iv) discovery of emergent behaviors that may affect policy choices. 
This section demonstrates the workflow on the Schelling model,  extending our prior work~\cite{saves2025mod} by adding an explicit uncertainty quantification study and a systematic XAI comparison across surrogate families, moving towards robust, ensemble-based interpretation for uncertain Simulations of Agent-Based Model (SABM).

First, Section~\ref{subsec:schelling} introduces the social system of interest and describes its main territory characteristics and actionable design variables together with the equilibrium target. After that, the workflow unfolds as described in Fig.~\ref{fig:graph_abs_schelling}. Section~\ref{subsec:schelling_abm} then describes the agent-based modeling approach for multi-agent systems handling; Section~\ref{subsec:doesch} introduces the DoE for the design space and the aggregated output metrics used to train surrogates; Section~\ref{subsec:sch_simulator} details the simulator implementation and analyzes the generated dataset; Section~\ref{subec:sch_surr} presents the surrogate models, and Section~\ref{subsec:rez_abs} reports the global and local XAI analyses, highlights key interactions and trade-offs.

\subsection{Segregation dynamics}  
\label{subsec:schelling}
The residential segregation model introduced by \textit{T. Schelling} investigates how individual local decisions can give rise to global segregation patterns~\cite{schelling1969}. It demonstrates that even agents with only mild preferences for neighbors of their own type, without any explicit intention to form homogeneous neighborhoods, can nonetheless produce highly segregated outcomes.

In our setting of the Schelling model, agents are randomly placed on a square lattice and assigned a \textit{type} (visualized by color). An agent is \textbf{Happy} if the proportion of its \textit{neighbors} that are of a different type is below its \textit{intolerance threshold}; otherwise it is \textbf{Unhappy}. At each time step, all unhappy agents move to a randomly chosen empty cell, and the simulation stops when all agents are happy (an equilibrium is reached). Besides the intolerance threshold, three parameters govern agents' behavior: the agent's \textit{perception distance} (neighborhood radius), the agent's \textit{density} (fraction of occupied cells), and the square map’s \textit{edge length}. Finally, it is useful to distinguish \emph{socially actionable} variables such as density, perception, or intolerance from \emph{contextual} or structural variables such as the city size or its population diversity (number of types), since they have different implications for interpretation and intervention~\cite{saves2025mod}.

This problem involves five input variables, consisting of three continuous and two integer variables, which jointly determine the dynamics of the segregation model.  
The three continuous variables correspond to the population density, intolerance threshold, and perception distance, while the two integer variables specify the number of individual types and the grid size.  
These variables, summarized in Tab.~\ref{tab:Sch_variable}, control both the structural configuration of the environment and the behavioral tendencies of agents, making them central to the analysis.   
\begin{table*}[h!]
\centering
\hspace{-0.1cm}
\begin{tabular}{|l|c|c|l|}
\hline
\textbf{Function/Variable} & \textbf{Nature} &  \textbf{Range} \\
\hline
Number of individual types & int & [2, 5] \\
Density of individuals & cont  & [0.01, 1] \\
Intolerance threshold & cont  & [0, 1] \\
Size of the square grid edges & int  & $[10, 40]$ \\
Perception distance  & cont  & [1, 10] \\
\hline
\end{tabular}
\caption{Design variables for diversity analysis.}
\label{tab:Sch_variable}
\end{table*}

\subsection{\color{black}{Architecture and stochasticity of the simulation model}}
\label{subsec:schelling_abm}
An ABM generally consists of \textit{agents}, which operate according to predefined or adaptive behavioral rules; an \textit{environment}, which structures spatial and network interactions; and \textit{interaction mechanisms}, which govern the behavior of the agents and the evolution of the system~\cite{michel2018}. Depending on their design, ABM can be classified as \textit{rule-based models} (following explicit logical instructions), \textit{stochastic models} (incorporating probabilistic elements), or \textit{hybrid models} (combining adaptive learning mechanisms). In practice, most multi-agent models are stochastic, as randomness is often introduced in agents’ decision-making, initial conditions, or interaction patterns to reproduce realistic variability.

However, as the number of agents and interactions increases, the computational cost associated with ABM grows rapidly, making large-scale simulations computationally intractable without significant computational capacities~\cite{lee2015complexities}. Furthermore, their stochastic nature typically requires running many simulations to obtain statistically meaningful results, further amplifying the computational burden~\cite{dunne2022complex}. In addition, these models may require extensive parameter calibration, involving thousands of runs to achieve a reliable fit. Similarly, uncertainty quantification remains a challenge, as slight variations in input parameters can lead to significant discrepancies in outcomes.  

ABM offers a natural multiscale framework for Schelling’s segregation model or other related multi-agent systems: they explicitly encode heterogeneous agents, local interaction rules, and a spatially structured environment. In Schelling-type models, an agent follows a simple tolerance rule which, under repeated stochastic updates, can amplify mild individual preferences into pronounced macro-scale segregation~\cite{schelling1969}. This bottom-up perspective shows how network topology, update protocols, and randomness interact to produce phase transitions, path dependence, and multiple equilibria, making ABM especially suited to study emergent, multi-scale phenomena under epistemic uncertainty.

\subsection{Design of experiments and aggregated output metrics}
\label{subsec:doesch}

To efficiently explore the parameter space, we generated a DoE using nested LHS, based on a uniform distribution 
~\cite{arenzana2021}. We generate an LHS of 200 points, structured to include a subset of 50 points that preserves the LHS. These nested LHS are generated with the Enhanced Stochastic Evolutionary (ESE) algorithm criterion as implemented in SMT (Surrogate Modeling Toolbox)~\cite{jin2005}. %
The nested structure provides both uniform stratified coverages and separated training and validation subsets for robust performance assessment~\cite{charayron2023towards}. 

To validate our workflow for both regression and classification, we consider two types of outputs: a binary variable indicating convergence and a continuous variable representing sparsity~\cite{singh2009}. 

A simulation is \textbf{convergent} if it stops because no individual moves on the grid anymore. However, knowing that a simulation may never converge, we set a threshold of 1000 iterations; after that, the simulation is stopped and considered \textbf{non-convergent}.

The ratio between the number of unlikely and likely neighbors is called the $u/l$ measure. For an agent at cell coordinates $(i, j)$, its $u/l$ ratio is given by:
\begin{equation}
[u/l]_{i,j} = \frac{q_{i,j} + w_{i,j}}{s_{i,j}}
\end{equation}
where $s_{i,j}$, $q_{i,j}$, and $w_{i,j}$ denote, respectively, the number of similar neighbors, different neighbors, and vacant cells surrounding the agent, with the agent considered a neighbor of itself to avoid division by zero. 

The \textbf{sparsity} $\langle [u/l] \rangle$ of the population is obtained by averaging the $[u/l]$ measure over all agents:
\begin{equation}
\langle [u/l] \rangle = \frac{1}{N} \sum_{(i,j) \in C} [u/l]_{i,j}
\end{equation}
where $N$ is the total number of agents whose associated cells form the set $C$. For a grid with side length $c$ and a density $\rho$, $N = \lfloor \rho c^2 \rfloor$.

\color{black}
    \paragraph{Characterizing aleatoric uncertainty in agent-based simulations}

Because probabilistic rules fundamentally govern decentralized agent-based models, characterizing the intrinsic aleatoric uncertainty of the system is a critical prerequisite to surrogate modeling. A rigorous method to determine the number of simulation replicates required is \textit{a priori} statistical power analysis~\cite{secchi2017controlling}.
Grounded in the Neyman-Pearson hypothesis-testing framework, this approach calculates the necessary sample size to control both Type-I and Type-II error rates strictly~\cite{huberty1993historical}.
For computational models, where data generation lacks the typical constraints of empirical data collection, stringent thresholds are often recommended (\textit{e.g.}, a significance level of $\alpha=0.01$ and a statistical power of $1-\beta=0.95$)~\cite{secchi2017controlling}. Establishing this exact replicate count \textit{a priori} serves as an objective stopping criterion, aligning with broader standards for systematic DoE and model replicability~\cite{lorscheid2012opening}.

However, this classical approach exhibits two limitations in the context of our surrogate-assisted workflow. First, it assumes a homogeneous noise distribution across the entire parameter space, whereas complex systems often exhibit highly localized instabilities. Second, it does not account for the interactions between aleatoric noise and the epistemic uncertainty inherent to surrogate approximation. To address this, we conducted a variance analysis across the replicates of our DoE~\cite{saves2026model}. Let $x_i$ denote the $i$-th configuration in the DoE space of size $200$, and let $k$ represent the number of replicates per configuration. 
Because the true population variance of the simulator is unknown, the uncertainty of the sample mean for a given configuration is estimated by the Standard Error (SE)~\cite{wooldridge2023standard}, which is approximated with the sample variance as $SE_i \approx \sqrt{\widehat{\text{Var}}(x_i) /k}$.

\begin{figure}[H]
\centering
\begin{minipage}{0.48\textwidth}
    \includegraphics[width=\linewidth,height=4.5cm]{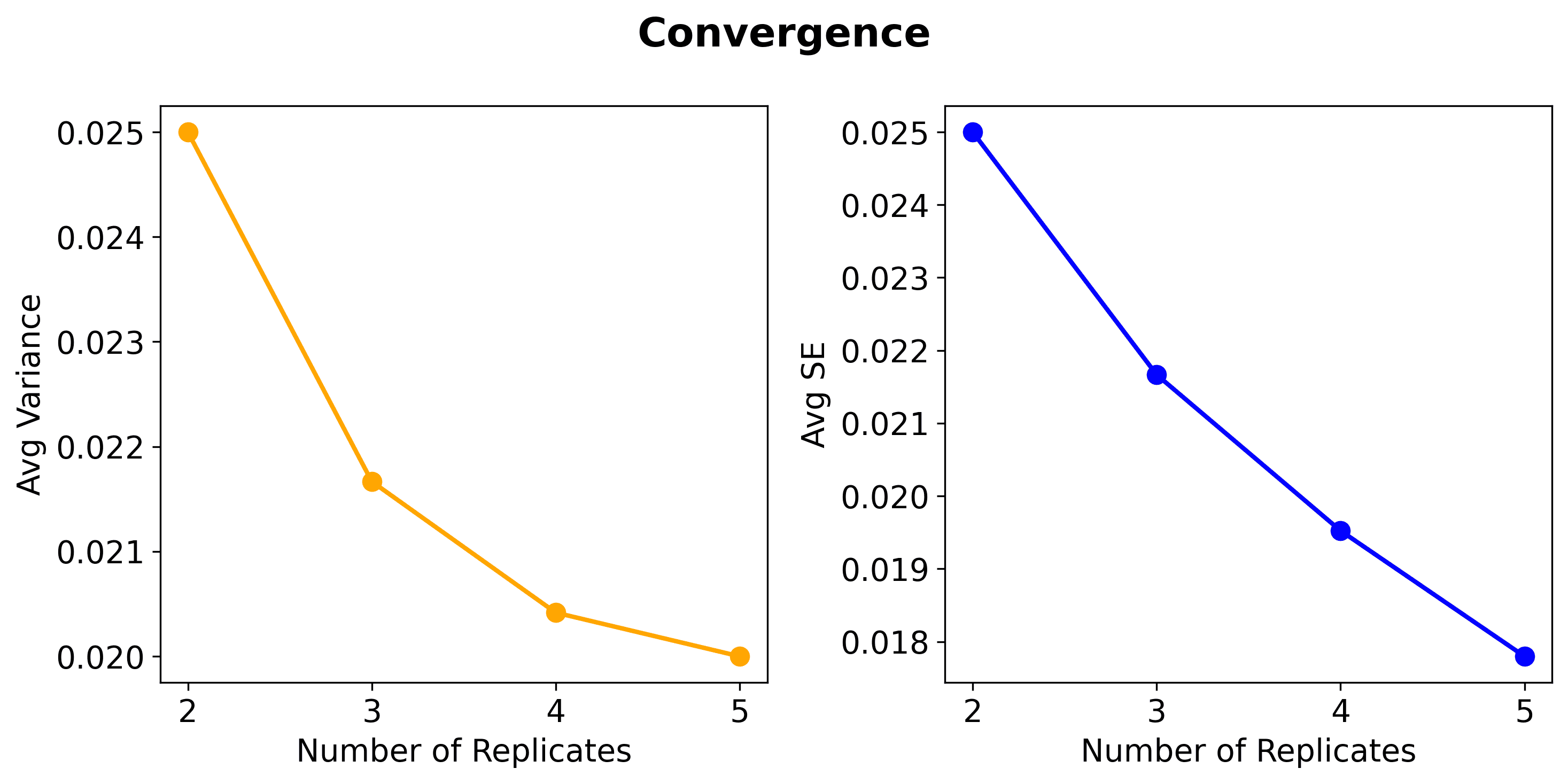}
\end{minipage}\hfill
\begin{minipage}{0.48\textwidth}
    \includegraphics[width=\linewidth,height=4.5    cm]{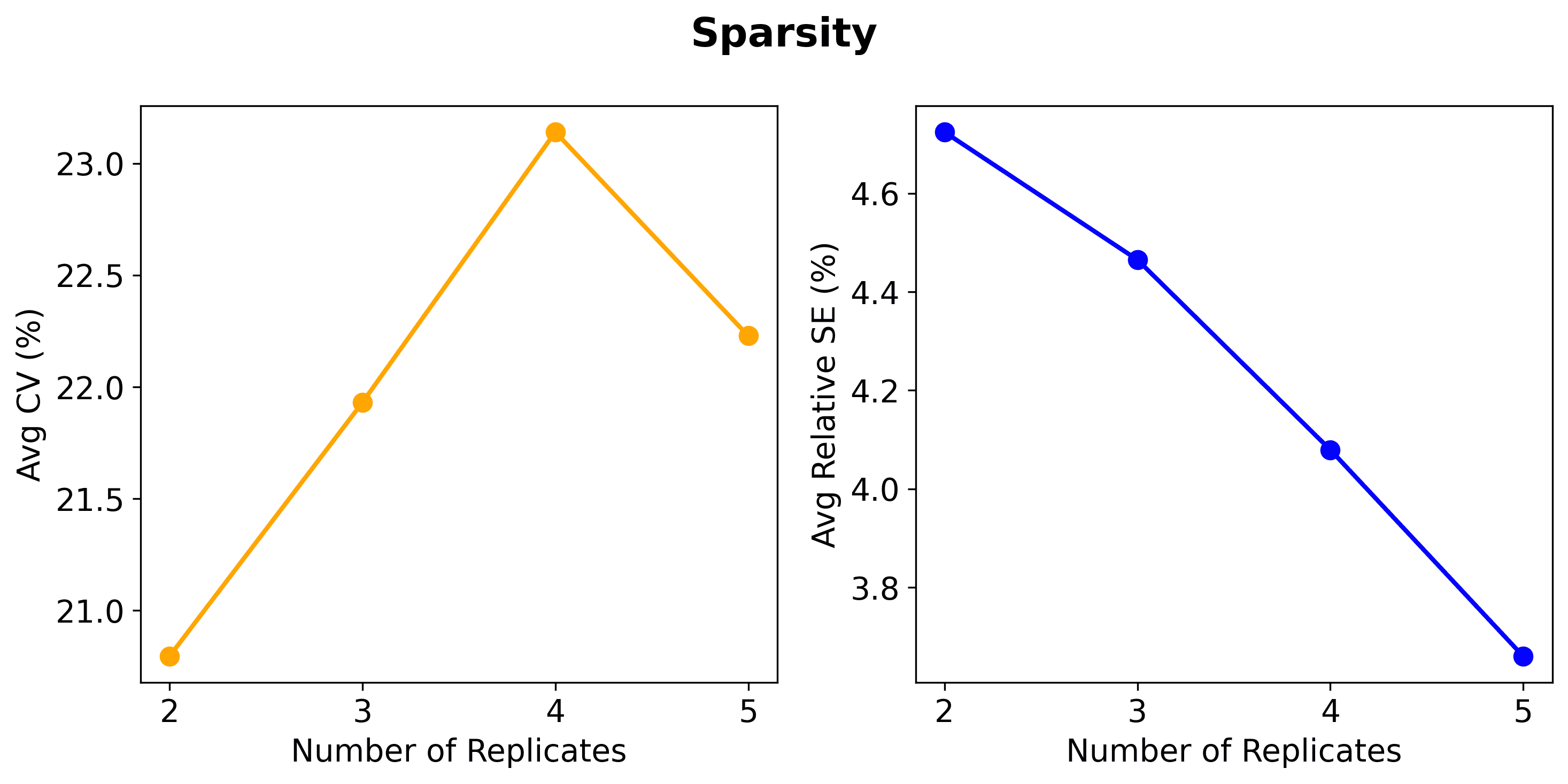}
\end{minipage}
\caption{Evolution of aleatoric uncertainty metrics with respect to the number of replicates. (Left) Raw variance and
standard error for the binary convergence output. (Right) Global
coefficient of variation and relative standard error for the continuous sparsity output. }
\label{fig:aleatoric_analysis}
\end{figure}

Figure~\ref{fig:aleatoric_analysis} traces the evolution of the average relative metrics across all $200$ points as the number of replicates $k$ increases. For the continuous sparsity output, we tracked the Coefficient of Variation (CV) to normalize proportional heteroscedasticity, alongside the Relative SE ($SE / \mu$). Although the CV initially increases up to 4 replicates, a statistical artifact when small sample sizes underestimate the true population variance, it stabilizes around 22.5\% at $k=5$. The Relative SE strictly decreases, reaching an acceptable error margin of approximately 4\%. For the binary convergence output (0 or 1), predicting the state is generally less prone to extreme variations than predicting continuous magnitude. Consequently, the raw variance stabilizes near 2\% at 4 replicates, and the average raw SE on the estimator of the mean drops below 2\% error. This confirms that a uniform computational budget of $k=5$ replicates is sufficient to filter out the simulator's intrinsic noise and provide a highly reliable estimate of the mean macroscopic behavior for subsequent surrogate training.

To support our claim of non-uniformity in the stochastic uncertainty, we investigated which design parameters drive this instability. A classical Pearson correlation analysis reveals a negative correlation ($r = -0.17$) between agents' density and the sparsity CV, indicating that lower density leads to more random behavior. This phenomenon is physically coherent: lower densities imply more empty zones, granting agents excessive freedom of movement and leading to highly variable cluster formations across different random seeds. Similarly, the map size is negatively correlated with both the sparsity CV ($r = -0.18$) and the convergence standard deviation ($r = -0.16$). On small grids, the random initial placement of agents has a disproportionate impact on the simulation outcome, as boundary and corner effects heavily influence the agents' total happiness and probability of convergence.

However, linear metrics alone are often insufficient to capture the full complexity of these interactions. For instance, the Pearson correlation between density and convergence standard deviation is a mere $0.07$. 
This observation suggests that the aleatoric uncertainty of complex systems can be driven by highly nonlinear interactions, which classical linear metrics may struggle to capture fully. Consequently, integrating surrogate-assisted XAI offers a valuable complementary approach to help map a simulator's stability boundaries.

To execute this exploration, we applied the fixed budget of $k=5$ replicates to the 200 DoE parameters, resulting in a total dataset of $m=1000$ simulations. Consistent with the validation methodology employed in Section~\ref{sec:dragon}, we used a 25\%/75\% train-test split on the averaged dataset to evaluate whether training a surrogate on limited data can reliably reproduce the insights obtained from a more exhaustive exploration.
\color{black}
A simulation runtime can vary significantly from 3 seconds to 30 minutes on an Intel® Core Ultra® 9 185H vPro processor, and the computational time was around 1.5 days overall.

\subsection{Implementation of the simulator}
\label{subsec:sch_simulator}

\begin{figure}[H]
\centering

\begin{subfigure}[b]{0.32\linewidth}
    \centering
    \includegraphics[width=\linewidth]{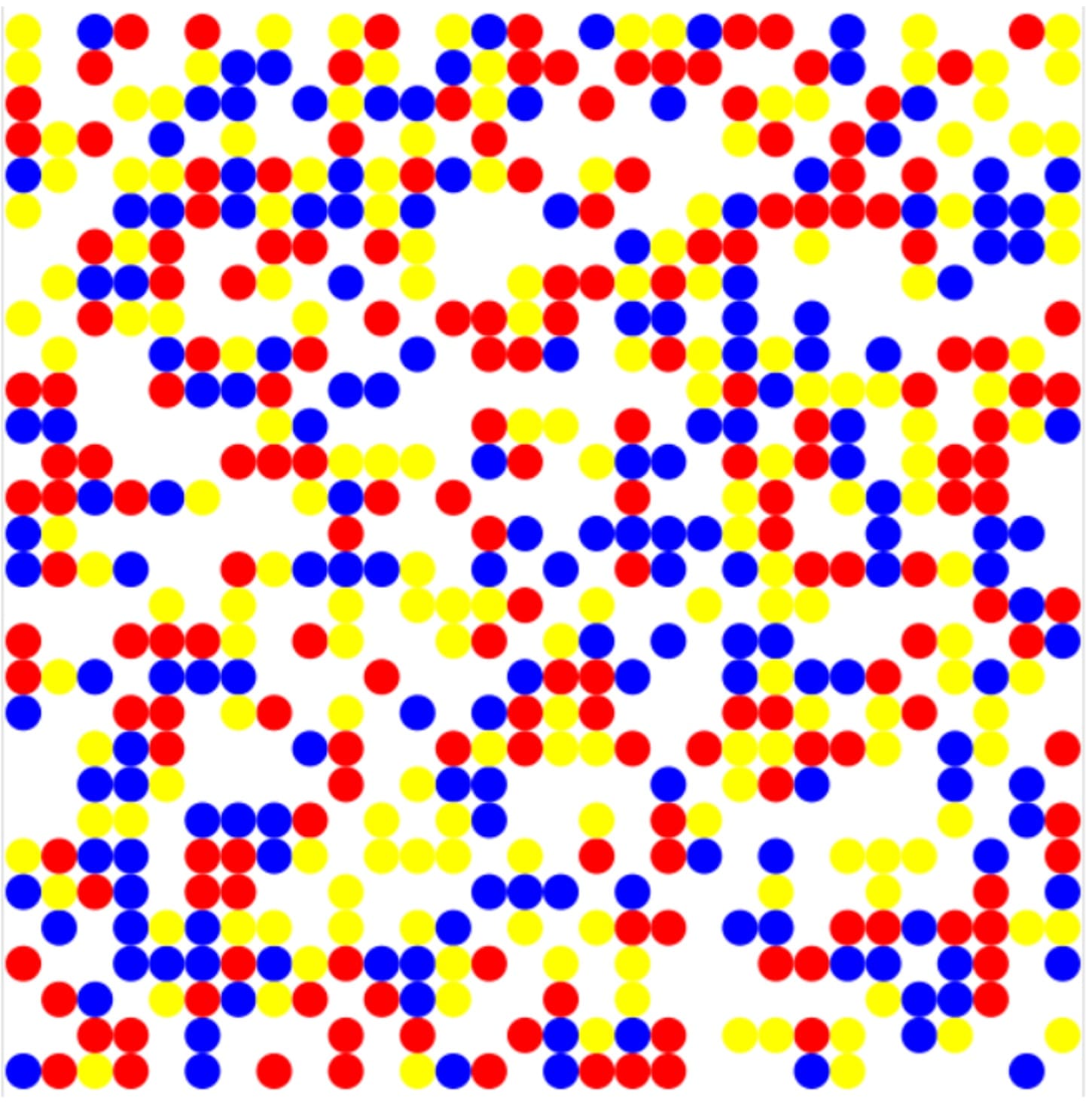}
    \caption{Initial state}
    \label{fig:gamma_simu1}
\end{subfigure}
\hfill
\begin{subfigure}[b]{0.32\linewidth}
    \centering
    \includegraphics[width=\linewidth]{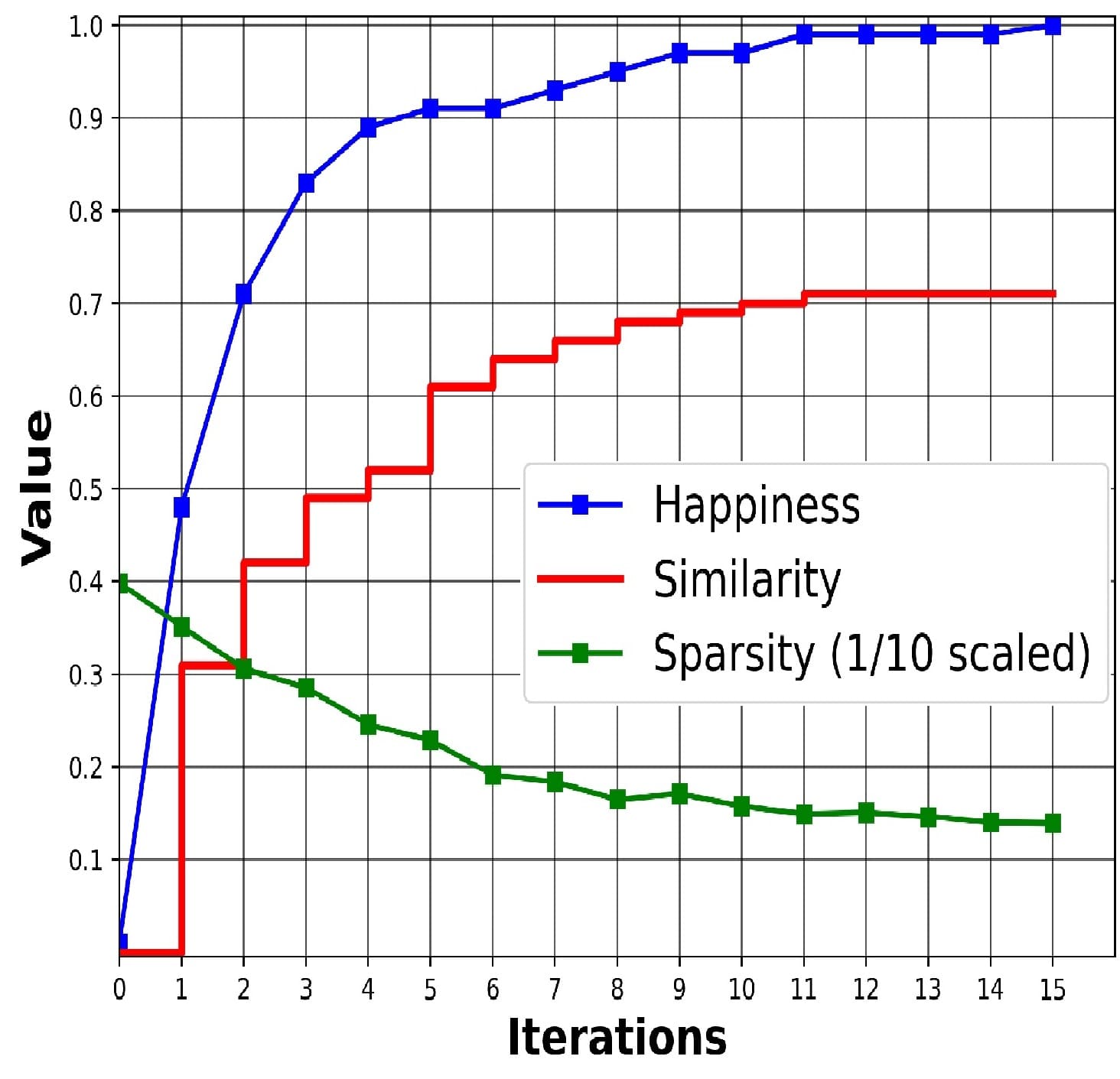}
    \caption{Evolution over time of indicators}
    \label{fig:gamma_simu2}
\end{subfigure}
\hfill
\begin{subfigure}[b]{0.32\linewidth}
    \centering
    \includegraphics[width=\linewidth]{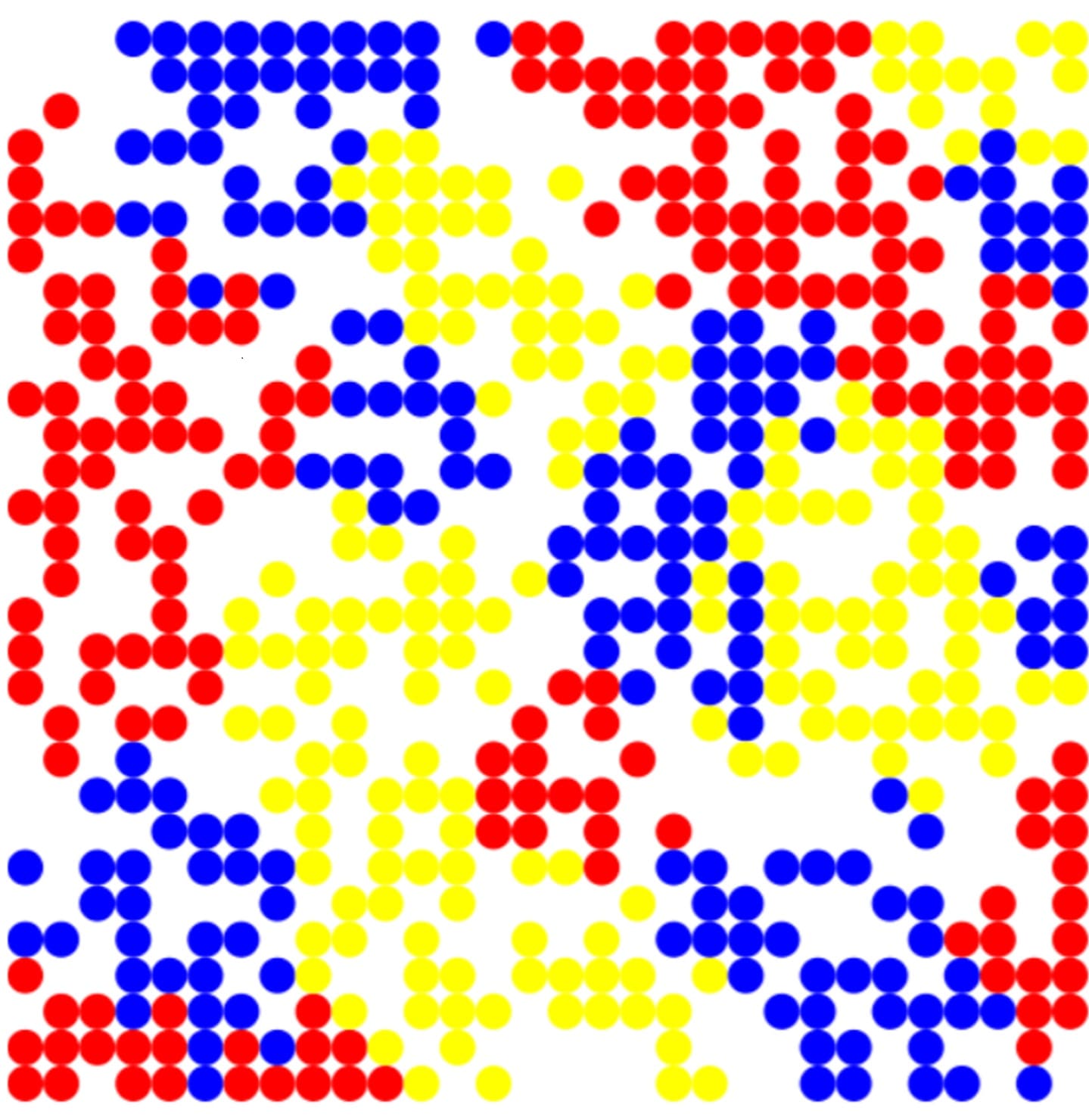}
    \caption{Final state}
    \label{fig:gamma_simu3}
\end{subfigure}

\caption{An instance of the segregation problem simulated on the GAMA platform.}
\label{fig:gamma_simu}
\end{figure}

In terms of implementation, we use the segregation model available in the model library of the open-source GAMA platform~\cite{taillandier2019}.
Figure~\ref{fig:gamma_simu} illustrates a simulation example with three types of agents, a density of 60\%, an intolerance threshold of 33\%, a perception distance of 3 units, and a grid of $30 \times 30$. Cells without agents remain empty, while agents are colored according to their type.  
On the left, the initial state is generated randomly, with an average percentage of individuals present in the agents’ neighborhood, referred to as \textbf{similarity}, almost zero, and many agents are unhappy. Gradually, the rate of happy agents, who then cease moving, increases until a stable state is reached after 15 iterations. This stable state, depicted on the right map, corresponds to an average dissimilarity of 30\%. The sparsity, which characterizes how dispersed the agents are, gradually decreases as the agents cluster together, eventually stabilizing around 1.4. This model is both simple enough to understand the marginal impact of every variable and complex enough to induce multi-scale emergent behaviors through tangled nonlinear dynamics.

 As for the engineering multidisciplinary design analysis that was based on fast approximations, here we use an approximation of the problem that does not represent the geographical data and relies on homogeneous agents. It should also be noted that there are exogenous sources of random uncertainty arising from the initial positions of individuals and their movements from one iteration to the next.
Notably, a simulation may fail to converge for two reasons. First, the parameterization may rule out any theoretical equilibrium. Second, an equilibrium may exist but not be reached within the 1000 iterations: when agents move randomly, independently, and without memory, the system can oscillate or follow a locally non-monotonic trajectory, delaying convergence.

A simulation lasts from 3 seconds to 30 minutes; thus, the agent model can be considered an expensive blackbox, which motivates the use of surrogate models in this context.

\paragraph{Dataset analysis}
\label{subec:sch_data}
The 1000 points suggested by the previously defined DoE have been evaluated through the simulator, yielding 1000 inputs and outputs in the considered dataset. 

\begin{figure}[H]
\centering
\hspace{-0.1cm}
\includegraphics[width=\linewidth, height = 10cm]{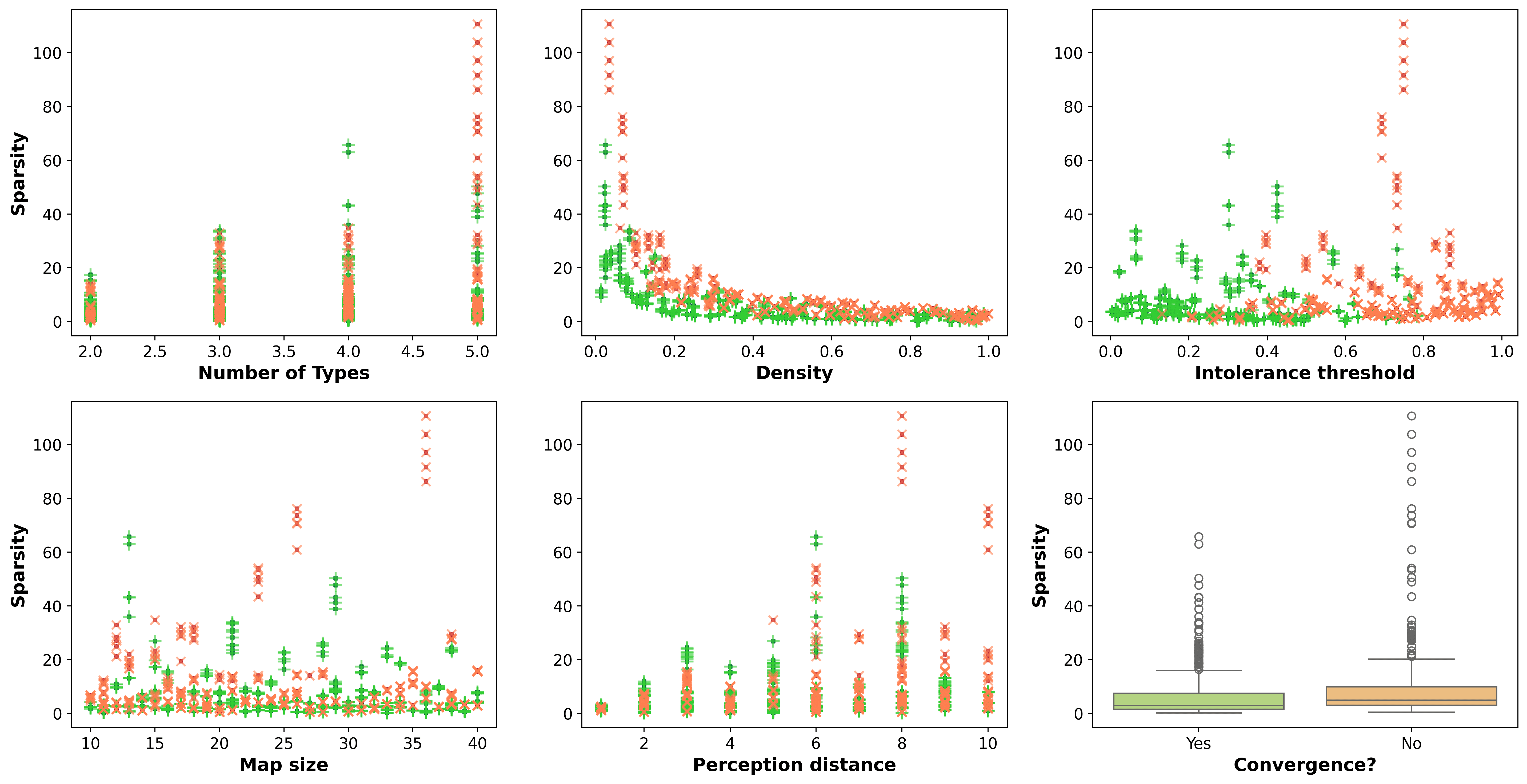}
 \caption{The dataset with converged simulations in green and the others in red.}
\label{fig:data_seg}
\end{figure}

Compared to the aircraft design study, we propose to add a first descriptive data analysis of the complete dataset, as illustrated in Fig.~\ref{fig:data_seg}, where for each of the 5 variables, the sparsity values are plotted against the different variable values.
Converged simulations are shown in green, while simulations that reached 1000 iterations are shown in red. The distribution of sparsity values is presented in the sixth figure as a boxplot.

A high value of sparsity indicates either a strong mixing of agents or many vacant cells, thus translating into low segregation and high spatial entropy. Conversely, low sparsity means that most agents are surrounded by similar neighbors, indicating high segregation and a more orderly spatial organization, which results in reduced entropy~\cite{singh2009}. 

%
%

Out of 1000 simulations, 546 converged and ended with an average sparsity of 6.20 compared to 9.27 for the 454 non-convergent simulations. Note that out of the 200 experiments, 16 partially converged, depending on the randomness in the 5 repetitions. 

\subsection{Surrogate modeling}
\label{subec:sch_surr}
Many studies have already demonstrated the value of surrogate models for learning from SABM for response surface  predictions~\cite{fabiani2024,llacay2025,angione2022}. We will test this approach on the Segregation model by learning a surrogate model using 50 out of the 200 points, each repeated 5 times. Then, we test and compare the predictions of many resulting surrogate models based on this dataset. 
In all experiments, we used each model’s default hyperparameter settings unless explicitly stated otherwise.
We want to predict two outputs, and therefore, we build two surrogate models, one for each output, and we address two tasks: (i) a regression problem to predict sparsity indices, and (ii) a classification problem to predict simulator convergence. In both cases, the surrogate models are trained as interpolative models over fixed, known input bounds.

Regarding the models tested from SMT~\cite{saves2024}, the Gaussian Process (\texttt{GP}) uses a heteroscedastic noise automatically estimated based on the repeated experiments~\cite{arenzana2021}, a linear trend (ordinary kriging), a squared exponential correlation kernel for regression, and an absolute exponential one for classification.
Similarly, Radial Basis Functions (\texttt{RBF}) use a linear trend, and the Inverse Distance Weighting (\texttt{IDW}) interpolation method employs a squared exponential kernel for regression and an absolute exponential kernel for classification. 
For the Linear Regression (\texttt{LR}) and Quadratic Polynomial (\texttt{QP}) models, we use the default options, and for the Regularized Minimal-energy Tensor-product Splines  (\texttt{RMTS}), we use the formulation based on Hermite cubic splines. Finally, for the Gradient-Enhanced Neural Network (\texttt{GENN}) model, we exploit the derivatives predicted by the GP, thereby enriching the information used during the network training.
We also compare these models with methods from Scikit-learn~\cite{pedregosa2011}. Thus, for both regression and classification, we notably use the Decision Tree (\texttt{DT}), Random Forest (\texttt{RF}), k-Nearest Neighbors (\texttt{kNN}), MultiLayer Perceptron (\texttt{MLP}), as well as the Support Vector Machine (\texttt{SVM}). For the kNN method, we chose to use 15 neighbors, corresponding to 3 different locations, due to the repetitions. All these methods are implemented in two versions: a regression version to predict sparsity and a classification version to predict whether the ABM converges. As with Scikit-learn, the Gradient Boosting (\texttt{GB}) from the XGBoost library treats regression and classification separately~\cite{chen2016}. We used the Polynomial Chaos Expansion (\texttt{PCE}) from OpenTurns~\cite{baudin2015open} with default options. We use the foundation model Tabular Prior-Data Fitted Network (\texttt{TabPFN}) from PriorLabs~\cite{hollmann2025accurate}. 
To finish with, we use the Contextual Interactive Ensemble Learning (\texttt{CIEL}) framework developed at IRIT~\cite{blanco-volle2024,levy2025timeciel}, using its default sub-models: logistic regression for regression tasks and support-vector machines for classification. CIEL embeds simple learners in a cooperative multi-agent ensemble that handles nonlinearity via local cooperation and dynamically reweights sub-models by contextual meta-features to improve robustness and interpretability.

To validate our 16 surrogate models, we held out 150 input configurations (750 new observations) that were not used during training.  We evaluate regression accuracy by the RMSE on this complete test set~\cite{armstrong1992error} and classification performance by the MCC, which balances true and false positives and negatives, and is especially informative under class imbalance and appropriate for the 2-class problem~\cite{chicco2021benefits}. 

For the sparsity regression problem, we further assess uncertainty quantification via the PVA metric~\cite{iooss2022,bachoc2013cross} that assesses the adequacy of predicted variances by comparing them to the actual squared errors. A PVA value close to zero indicates that the predicted variances align well with the observed errors.
For GP, we use the posterior variance directly; for RF, we rely on the sample variance across trees; and for TabPFN, we approximate variance from the interquantile range (0.15865, 0.84135) under a normal distribution assumption. All other models' uncertainty quantification is based on a 50/50 split conformal prediction to estimate $\hat\sigma_i^2$~\cite{shafer2008tutorial}. 

\color{black}{Note that our metrics are partial and subject to misinterpretation. For instance, RMSE is an aggregated measure that reflects average error rather than maximal or median error, offering little insight into the overall error distribution. Similarly, MCC is less representative than the full confusion matrix. Finally, PVA relies on the strong assumption of normally distributed errors where predicted variance equals the unknown error.}
We use RMSE to have an idea of the global regression precision of the model, MCC because it is a symmetric metric to quantify the confusion matrices, and PVA because it consists of a simple metric that does not require any quantile computations.
All results (RMSE, MCC, and PVA) for the 16 surrogate models are reported in Tab.~\ref{tab:results}, where we have organized the models into the following categories: tree-based models (RF, GB, DT); kernel-based models (GP, RBF, SVM); linear and quadratic models (LR, QP); reduced basis models (RMTS, PCE); neighborhood-based models (kNN, IDW, CIEL); neural network models (MLP, GENN); and the transformer-based model (TabPFN). These various families are colored accordingly for more visibility, and the overall global aggregated results are in darker green when they have the best performance.

\begin{table}[htb]
    \centering
    \hspace{-0.1cm}
    \resizebox{0.65\columnwidth}{!}{
\begin{tabular}{lcccccc}
\toprule
\textbf{Model} 
  & \textbf{RMSE} 
  & \textbf{MCC} 
  & \textbf{PVA} 
  & \makecell{\textbf{Reg.}\\ \textbf{Rank}}  
  & \makecell{\textbf{Class.}\\ \textbf{Rank}}  
  & \makecell{\textbf{PVA}\\ \textbf{Rank}}  \\
\midrule
\cellcolor[HTML]{A7C8EB}\color{black} RF     
  & \cellcolor[HTML]{00441B}\color[HTML]{F1F1F1}6.86  
  & \cellcolor[HTML]{2C944C}\color[HTML]{F1F1F1}0.83  
  & \cellcolor[HTML]{DDF2D8}\color[HTML]{000000}4.13  
  & 3  & 4   & 11 \\

\cellcolor[HTML]{A7C8EB}\color{black} GB     
  & \cellcolor[HTML]{8ED08B}\color[HTML]{000000}9.21  
  & \cellcolor[HTML]{2C944C}\color[HTML]{F1F1F1}0.82  
  & \cellcolor[HTML]{DDF2D8}\color[HTML]{000000}3.44  
  & 9  & 5   & 10 \\

\cellcolor[HTML]{A7C8EB}\color{black} DT     
  & \cellcolor[HTML]{8ED08B}\color[HTML]{000000}8.34  
  & \cellcolor[HTML]{8ED08B}\color[HTML]{000000}0.75  
  & \cellcolor[HTML]{DDF2D8}\color[HTML]{000000}4.19  
  & 7  & 8   & 12 \\

\cellcolor[HTML]{A1E5C9}\color{black} GP     
  & \cellcolor[HTML]{2C944C}\color[HTML]{F1F1F1}8.22  
  & \cellcolor[HTML]{00441B}\color[HTML]{F1F1F1}0.87  
  & \cellcolor[HTML]{8ED08B}\color[HTML]{000000}3.14  
  & 6  & 1   & 9  \\

\cellcolor[HTML]{A1E5C9}\color{black} RBF    
  & \cellcolor[HTML]{2C944C}\color[HTML]{F1F1F1}8.17  
  & \cellcolor[HTML]{00441B}\color[HTML]{F1F1F1}0.87  
  & \cellcolor[HTML]{FFFFFF}\color[HTML]{000000}5.32  
  & 5  & 1   & 14 \\

\cellcolor[HTML]{A1E5C9}\color{black} SVM    
  & \cellcolor[HTML]{FFFFFF}\color[HTML]{000000}11.53 
  & \cellcolor[HTML]{FFFFFF}\color[HTML]{000000}0.21  
  & \cellcolor[HTML]{8ED08B}\color[HTML]{000000}1.01  
  & 14 & 16  & 7  \\
  
\cellcolor[HTML]{FDDC7A}\color{black} LR     
  & \cellcolor[HTML]{8ED08B}\color[HTML]{000000}8.56  
  & \cellcolor[HTML]{2C944C}\color[HTML]{F1F1F1}0.82  
  & \cellcolor[HTML]{00441B}\color[HTML]{F1F1F1}0.24  
  & 8  & 6   & 1  \\

\cellcolor[HTML]{FDDC7A}\color{black} QP     
  & \cellcolor[HTML]{00441B}\color[HTML]{F1F1F1}6.35  
  & \cellcolor[HTML]{8ED08B}\color[HTML]{000000}0.77  
  & \cellcolor[HTML]{2C944C}\color[HTML]{F1F1F1}0.51  
  & 2  & 7   & 5  \\

\cellcolor[HTML]{D8B8E9}\color{black} PCE    
  & \cellcolor[HTML]{2C944C}\color[HTML]{F1F1F1}7.84  
  & \cellcolor[HTML]{8ED08B}\color[HTML]{000000}0.74  
  & \cellcolor[HTML]{FFFFFF}\color[HTML]{000000}9.12  
  & 4  & 9   & 15 \\

\cellcolor[HTML]{D8B8E9}\color{black} RMTS   
  & \cellcolor[HTML]{FFFFFF}\color[HTML]{000000}11.60 
  & \cellcolor[HTML]{DDF2D8}\color[HTML]{000000}0.52  
  & \cellcolor[HTML]{FFFFFF}\color[HTML]{000000}9.98  
  & 15 & 11  & 16 \\
  
\cellcolor[HTML]{F6D1D2}\color{black} kNN    
  & \cellcolor[HTML]{FFFFFF}\color[HTML]{000000}11.75 
  & \cellcolor[HTML]{FFFFFF}\color[HTML]{000000}0.30  
  & \cellcolor[HTML]{00441B}\color[HTML]{F1F1F1}0.40  
  & 16 & 15  & 3  \\

\cellcolor[HTML]{F6D1D2}\color{black} IDW    
  & \cellcolor[HTML]{DDF2D8}\color[HTML]{000000}10.57 
  & \cellcolor[HTML]{FFFFFF}\color[HTML]{000000}0.41  
  & \cellcolor[HTML]{FFFFFF}\color[HTML]{000000}4.62  
  & 11 & 13  & 13 \\

\cellcolor[HTML]{F6D1D2}\color{black} CIEL   
  & \cellcolor[HTML]{FFFFFF}\color[HTML]{000000}11.03 
  & \cellcolor[HTML]{FFFFFF}\color[HTML]{000000}0.38  
  & \cellcolor[HTML]{8ED08B}\color[HTML]{000000}2.65  
  & 13 & 14  & 8  \\

\cellcolor[HTML]{F4A3A3}\color{black} MLP    
  & \cellcolor[HTML]{DDF2D8}\color[HTML]{000000}9.87  
  & \cellcolor[HTML]{DDF2D8}\color[HTML]{000000}0.73  
  & \cellcolor[HTML]{2C944C}\color[HTML]{F1F1F1}0.60  
  & 10 & 10  & 6  \\

\cellcolor[HTML]{F4A3A3}\color{black} GENN   
  & \cellcolor[HTML]{DDF2D8}\color[HTML]{000000}10.93 
  & \cellcolor[HTML]{DDF2D8}\color[HTML]{000000}0.48  
  & \cellcolor[HTML]{00441B}\color[HTML]{F1F1F1}0.25  
  & 12 & 12  & 2  \\

\cellcolor[HTML]{DBDBE6}\color{black} TabPFN 
  & \cellcolor[HTML]{00441B}\color[HTML]{F1F1F1}4.31  
  & \cellcolor[HTML]{00441B}\color[HTML]{F1F1F1}0.86  
  & \cellcolor[HTML]{2C944C}\color[HTML]{F1F1F1}0.47  
  & 1  & 3   & 4  \\

\bottomrule
\end{tabular}
}
    \caption{Global predictive qualities comparison across 16 surrogate models, with darker green shades representing better performance.}
    \label{tab:results}
\end{table}

Our results confirm that different surrogate models exhibit variable and complementary performances at the global scale, making them versatile and indicating promising future research towards a mixture of experts or ensemble approaches to combine various models. In particular, no method stands out as one of the top three for every metric, and no method falls into the bottom 4 for every metric.  

\begin{figure}[H]
        \centering
        \hspace{-0.1cm}
        \includegraphics[ width=0.65\columnwidth]{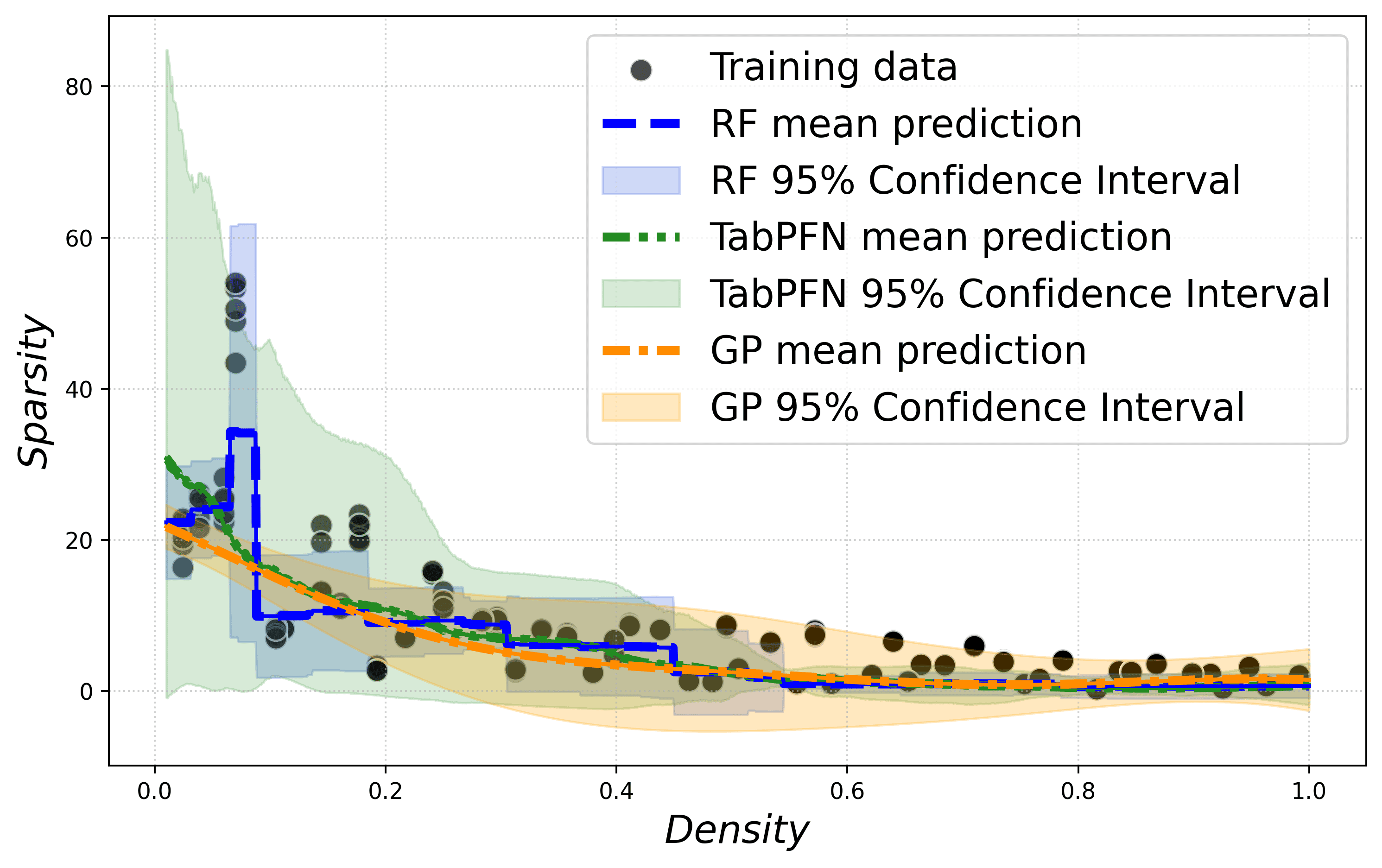}
            \caption{RF, TabPFN, and GP response surfaces of sparsity as a function of density with the other variables fixed.}
        \label{fig:modeles_1D}
\end{figure}

Ensemble methods have the disadvantage of requiring a large number of predictions to estimate uncertainties, thereby generating very discontinuous response surfaces, which can, in turn, limit their analytical capacity, interpretability, and gradient prediction, but they open up many avenues for research.  However, and quite surprisingly, LR and its slightly more complex counterpart QP perform quite well on this problem, even if they are simple polynomials of degree 1 and 2, respectively. \color{black}{Although this specific behavior is difficult to extrapolate to other problems, interpreting first- or second-degree relationships is generally highly informative. Such simple models capture most of the global information and rapidly highlight relevant features.} Similarly, logic explanations such as abductive and contrastive explanations are known to be easy to compute based on a decision tree surrogate without any loss of generality~\cite{audemard2022explanatory}. Therefore, even if RF and GP/RBF can outperform, respectively, QP or DT, we always recommend starting with these elemental, transparent surrogate models as they are the most efficient for a quick and first overview~\cite{henckaerts2022stakes}. 

For the classification of convergent simulations, the RBF and GP models yielded the lowest misclassification rates with identical confusion matrices, followed closely by TabPFN and RF. LR and GB performed comparably: LR achieved more true positives (520 vs. 508) at the expense of fewer true negatives (389 vs. 402), but these compensating shifts led to nearly identical MCC values. Thus, despite different confusion-matrix profiles, LR and GB exhibit equivalent overall correlation between predicted and actual labels.
These results highlight the strength of kernel and probabilistic models in learning decision boundaries under limited data.  In contrast, SVM, IDW, CIEL, and kNN showed a high number of misclassifications, suggesting that distance-based classifiers struggle with the repeated simulations and sparse grid structure, although the weighting helps IDW to perform slightly better. 
LR, GENN, and kNN yielded PVA values close to zero.  We obtained different aggregated metrics for different surrogate models, but to go further, we need to complement these global metrics with local analysis.

To also study the local predictions, let's examine more in depth the qualitative behaviors of the three models that come with in-house uncertainty quantification. To do that, a spatial study comparing TabPFN, GP, and RF is provided in Fig.~\ref{fig:modeles_1D}. In what follows, each variable has been fixed to its average value (or its integer part in the case of discrete variables) except for density, which varies between $0$ and $1$, to illustrate the smooth aspect of the GP and the TabPFN model and the piecewise continuous aspect of the RF along with the poor uncertainty estimation associated with it, obtained by resampling over 100 estimators (default options), whereas the GP can be directly used for variance-based interpretability~\cite{iooss2022}. TabPFN variance is obtained with its quantile prediction and shows a clear non-smooth and non-interpretable behavior. In fact, TabPFN's overall variance prediction has been shown to overestimate the uncertainty when computing the PVA, whereas the GP tends to underestimate it, in particular here with repeated experiences and aleatoric intrinsic noise to model. This one-dimensional example illustrates the complementarity of predictive models and the importance of adapting the model to the studied scale at hand. 
Therefore, the figure shows that two models can agree on the expected mean prediction but still predict very different standard deviations, highlighting the complementarity of global and local analyses.
The TabPFN model is really time-consuming in inference, but still achieved the lowest RMSE and second-best classification error, indicating superior ability to capture the nonlinear mapping from input parameters to sparsity, as is also the case for RF and GP, which are well-fitted for these low-data contexts. In fact, for this type of sparse and uniformly distributed data, it is generally preferable to use a GP or an RBF, as these models learn quickly and effectively capture nonlinear behaviors on small datasets~\cite{dyer2024}. Also, RF, as an ensemble approach, provides comparable results here, which can be explained by its ability to approximately reproduce the behaviors of nonlinear SABM. Gradient boosting and decision tree, coming slightly behind, also confirm the capacity of these methods to handle complex interactions among agents~\cite{singh2009}. 
However, these results should be put in light of the approximation of such variance by conformal prediction, which is generally non-smooth. TabPFN’s variance approximation via interquantile range also performed reasonably, whereas both GP and RF tend to underestimate the error because of repeated experiences.

\subsection{Explainable methods, analyses, and insights extraction}
\label{subsec:rez_abs}

The goal of this section is to leverage surrogate models for extracting insights on the underlying model by leveraging XAI methods.
%
%

\subsubsection{Global sensitivity analysis}
\begin{figure}[H]
\centering
\includegraphics[height=5cm, width=0.54\columnwidth]{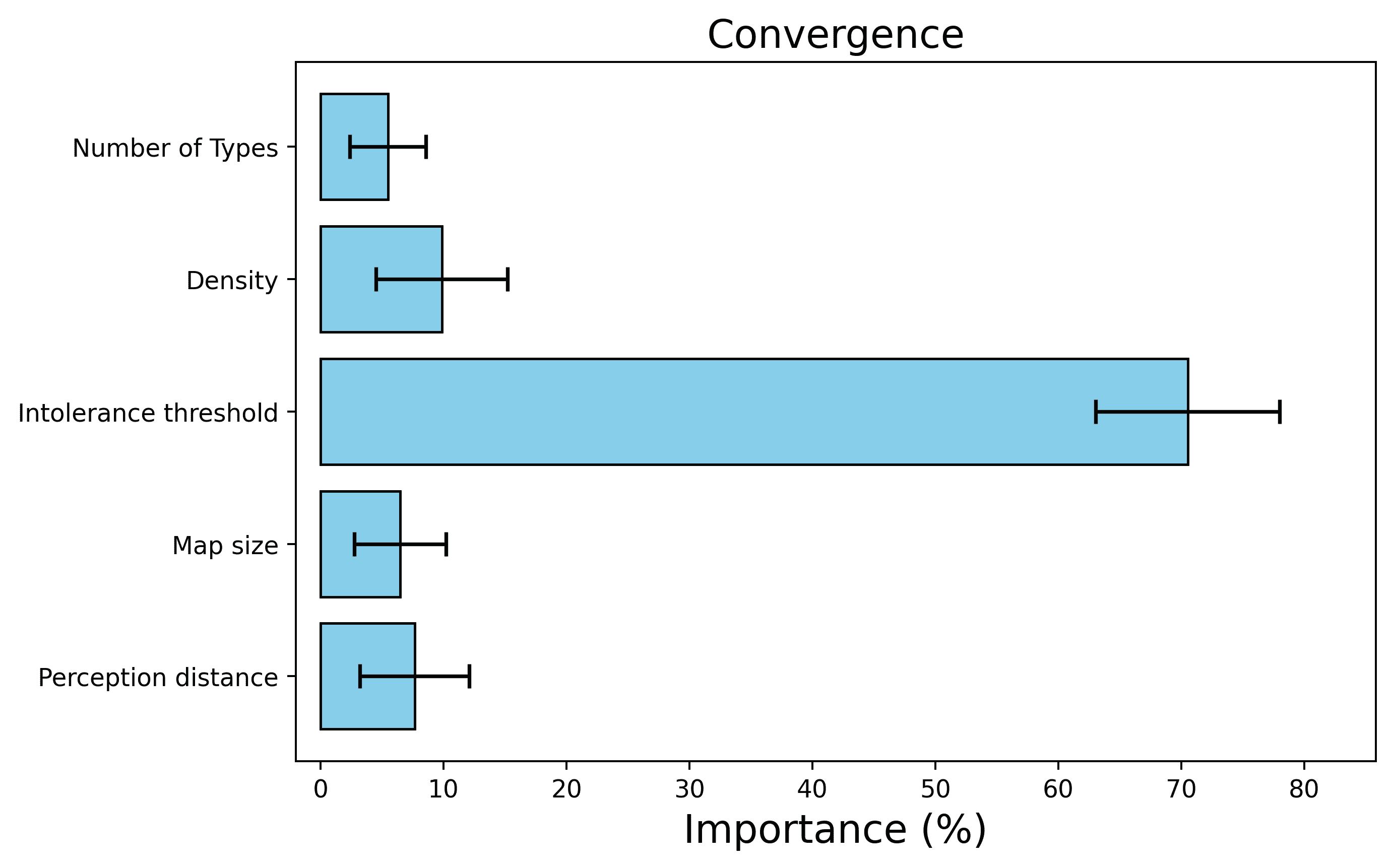}
\hspace{-0.1cm}
\includegraphics[height=5cm, width=0.44\columnwidth]{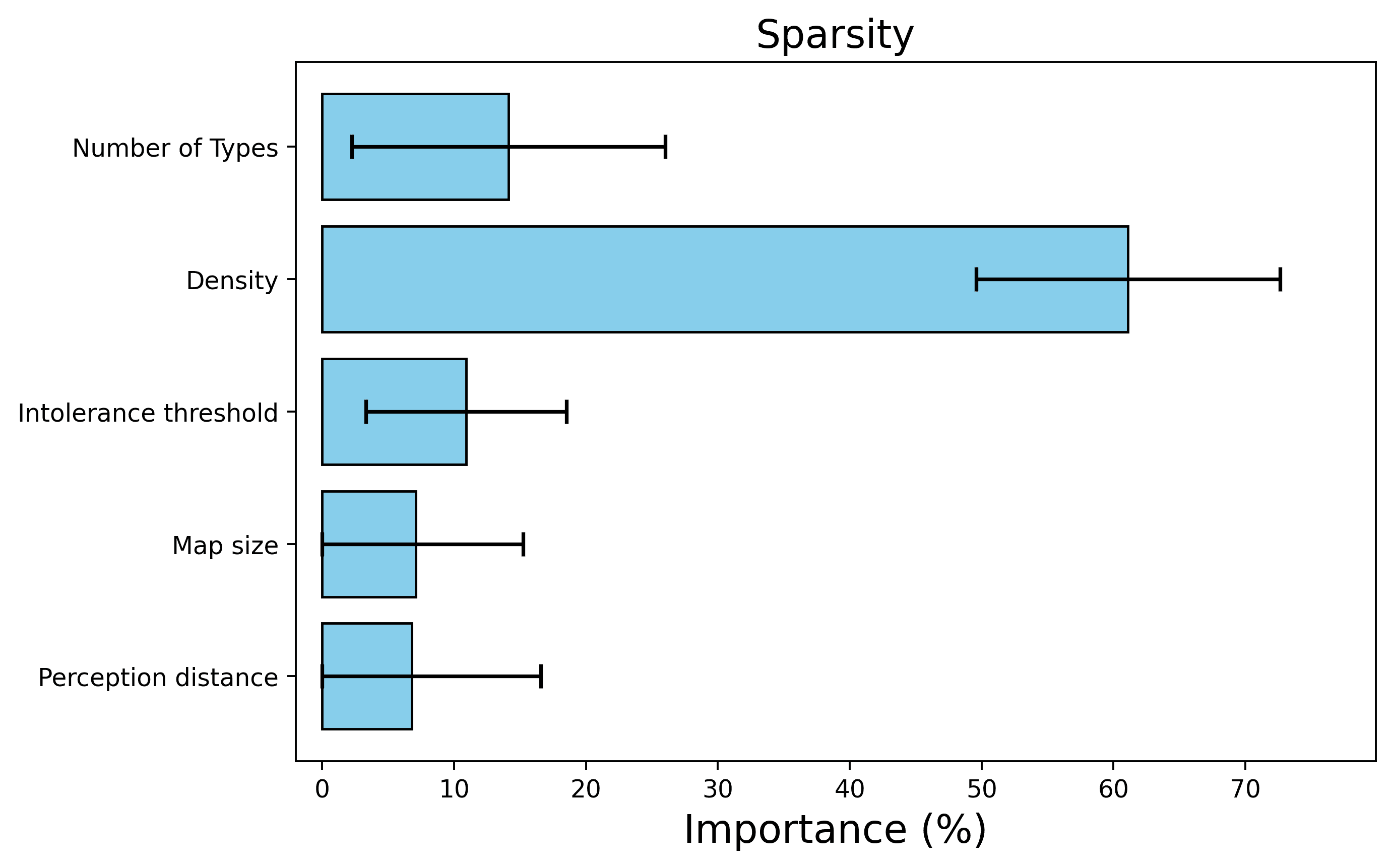}
\caption{Variables importance in the random forest based on MDI.}
\label{fig:imp_rf}
\end{figure}

\begin{figure}[H]
\centering
\includegraphics[height=5cm, width=0.54\columnwidth]{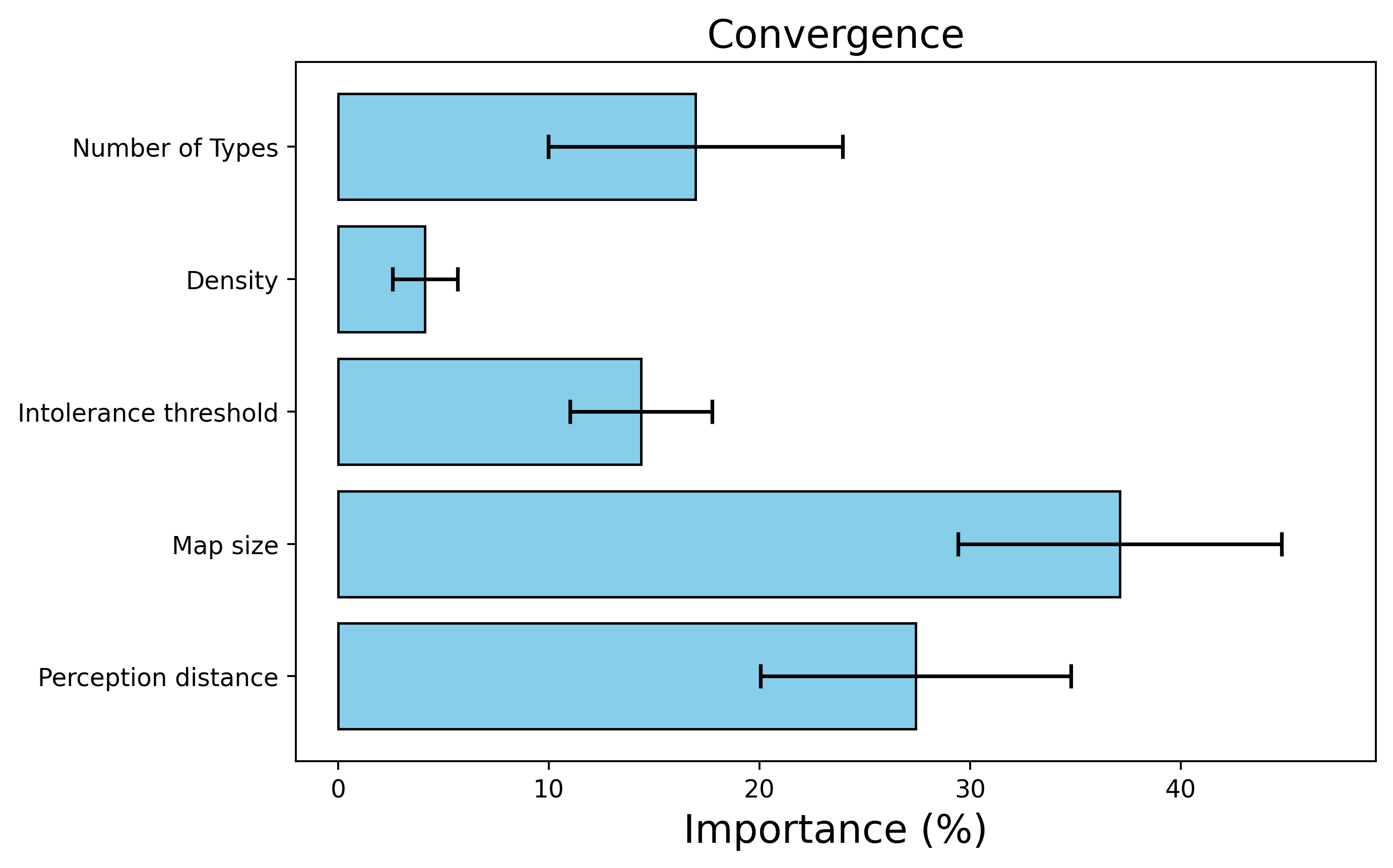}
\hspace{-0.1cm}
\includegraphics[height=5cm, width=0.44\columnwidth]{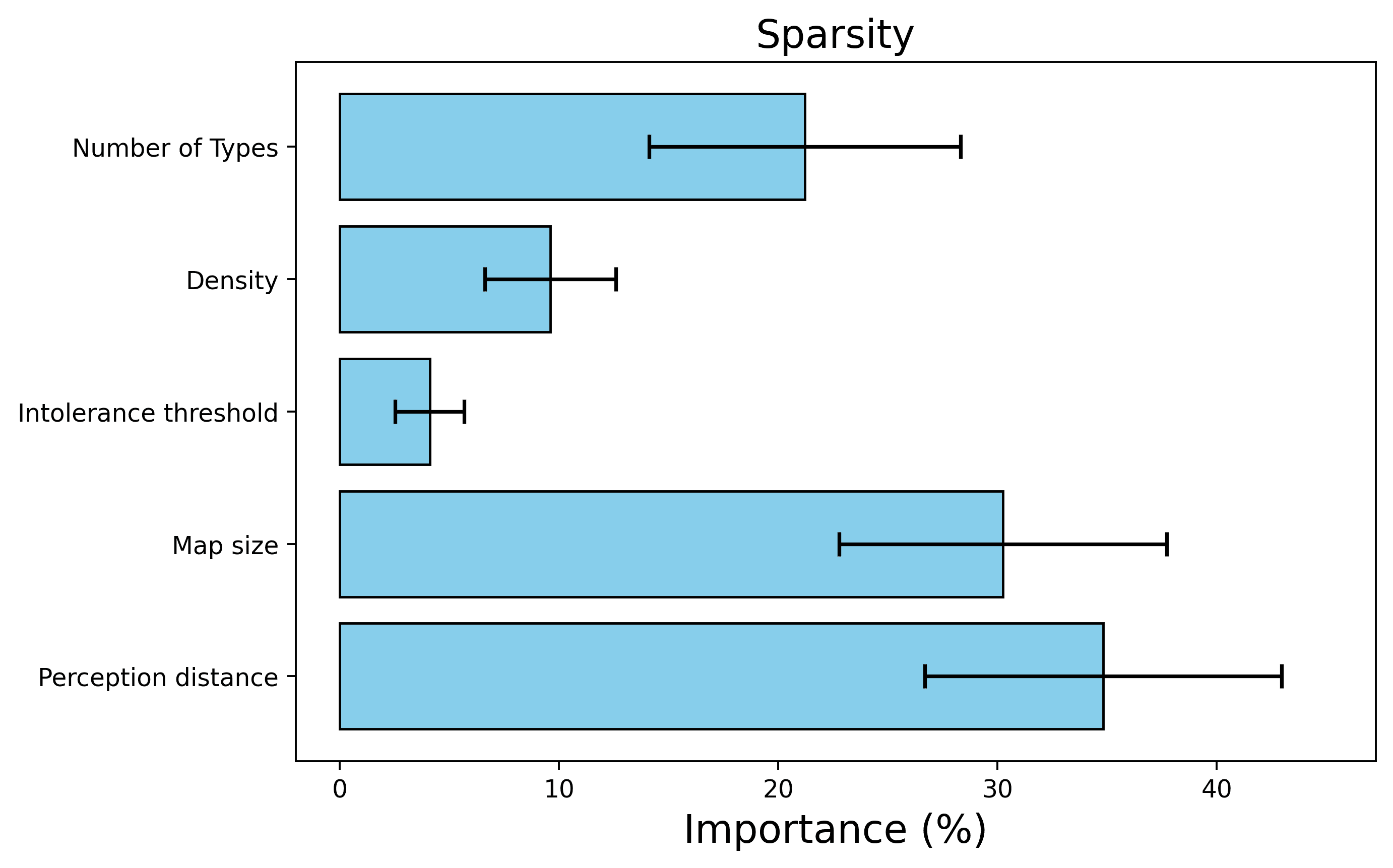}
\caption{Variables importance in the k-Nearest Neighbors based on SHAP.}
\label{fig:imp_knn}
\end{figure}

We aim to quantify how input parameters drive model outputs by applying multiple global sensitivity methods and checking the consistency of their conclusions. Global sensitivity analysis identifies influential inputs, dominant interactions, and how input uncertainty propagates to output variability. 
Although decision trees are limited and discontinuous, they are sequences of simple procedures that are easy to interpret. For example, the first node of the GB or DT is "intolerance threshold below 0.63", which confirms what was graphically observed on the complete dataset.  
In a random forest, the importance of a variable is often measured via the Mean Decrease in Impurity (MDI). At each split of a tree, the reduction in impurity, often measured by the Gini index, quantifies the probability that a sample is misclassified. 
This reduction is attributed to the variable that enabled the split, and by summing these reductions over all trees, a global measure of the importance of each feature is obtained.
However, this approach can bias the estimation in favor of variables with high cardinality, which offer more splitting possibilities. In this way, in our test case, we obtain the importances shown in Fig.~\ref{fig:imp_rf}. For both outputs, the variables that most influence the model's predictions, as identified on the extended dataset, are clearly found to be intolerance for convergence and density for sparsity. It is also observed that grid size has a rather weak influence on the outputs.
Equivalent importance analyses can also be performed with other models. For instance, applying the same procedure to a k-nearest neighbors model and aggregating SHAP absolute values produces the results shown in Fig.~\ref{fig:imp_knn}. The predicted importances differ significantly from those of the random forest analysis, as most variables appear significant. This occurs because kNN SHAP values primarily reflect local, instance-specific effects rather than global trends, making them less suitable for traditional global sensitivity analysis. Appendix~\ref{app:companal} further illustrates that importance outcomes vary depending on whether the surrogate captures broad system-level tendencies or local deviations, and as such, the SHAP importance of a Gaussian process is aligned with the random forest results of Fig.~\ref{fig:imp_knn}. 

\color{black}
To assess the robustness of the derived sensitivity indices and account for the sampling variance, 95\% confidence intervals were computed through a rigorous block subsampling procedure. The original dataset relies on a DoE comprising $N_{lhs}= 200$ unique input configurations, each evaluated 5 times to capture the inherent stochasticity of the Schelling model, resulting in 1000 total evaluations.

Applying a standard bootstrap across the 1000 individual evaluations would disrupt this replicate structure, leading to unbalanced stochastic noise representation in the surrogate training sets. To deal with this problem, we used an alternative block subsampling approach. In standard bootstrapping, drawing $N_{lhs}$ samples with replacement implies that the probability of any specific instance not being selected is $(1 - 1/N_{lhs})^{N_{lhs}}$. As $N_{lhs}$ grows, this probability converges to $1/e$. Consequently, a standard bootstrap sample is expected to contain a proportion of $1 - 1/e \approx 63.2\%$ unique instances from the original dataset.
To strictly preserve the replicate balance while mirroring this mathematical property, our approach randomly selects 126 unique input configurations (approximately 63.2\% of the 200 available configurations) without replacement at each iteration. For each chosen configuration, all 5 of its replicates are retained, producing a robustly balanced training subset of exactly 630 evaluations.

This block subsampling is repeated 100 times. For each iteration, the surrogate models, random forest and k-nearest neighbors, are retrained from scratch, and their respective sensitivity indices are computed. Finally, consistent with established global sensitivity analysis practices~\cite{herman2017salib}, the 95\% confidence bounds are derived using the empirical standard deviation $\sigma$ across the 100 iterations, computed as $\pm 1.96 \times \sigma$. These symmetric confidence intervals are then centered around the primary sensitivity indices, which are estimated by training the surrogate models on the complete dataset.
\color{black}

\subsubsection{Local explainability visualization}

\begin{figure}[H]
\centering
\includegraphics[height=5cm, width=0.45\columnwidth]{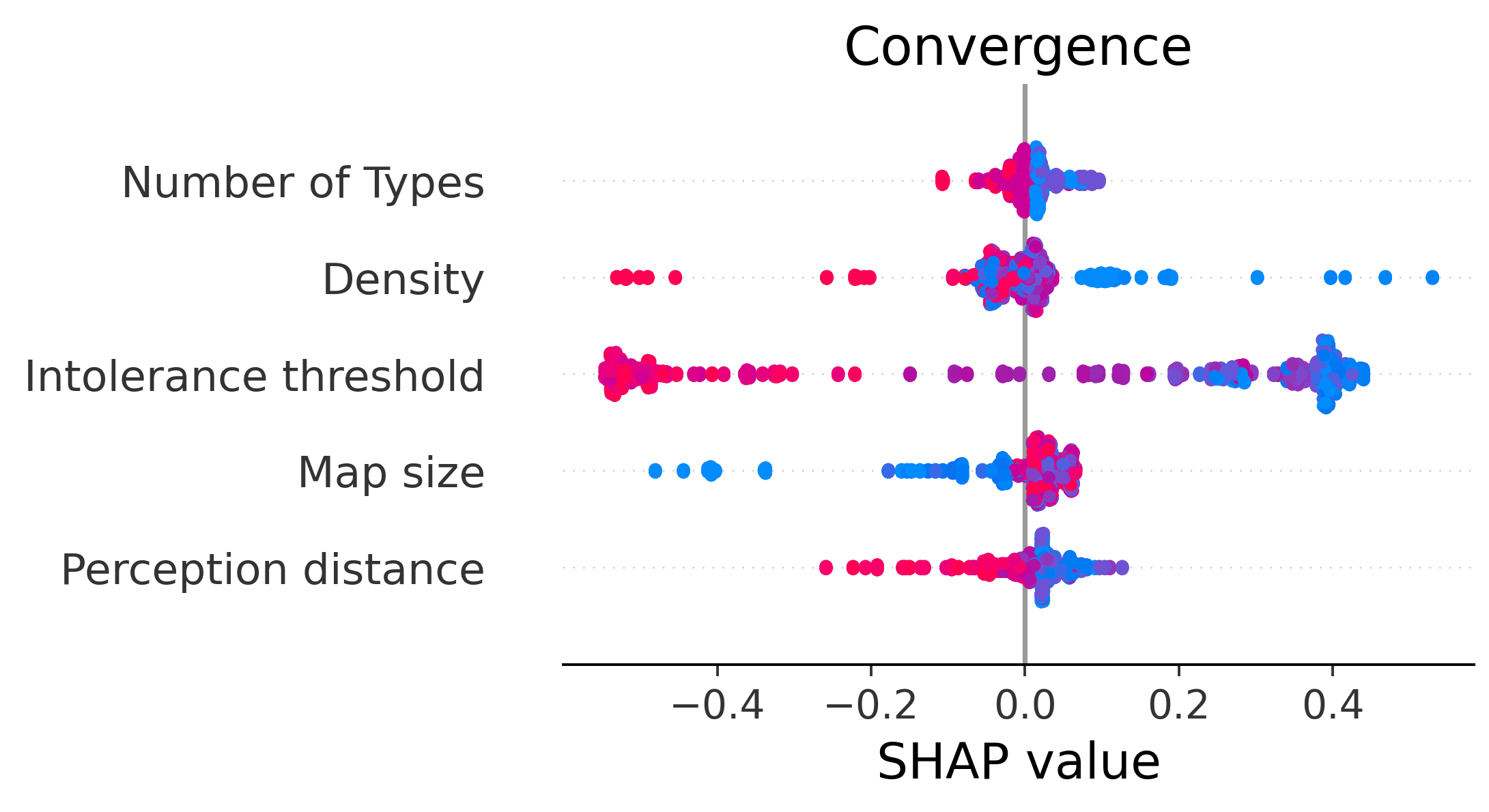}
\hspace{-0.1cm}
\includegraphics[height=5cm, width=0.45\columnwidth]{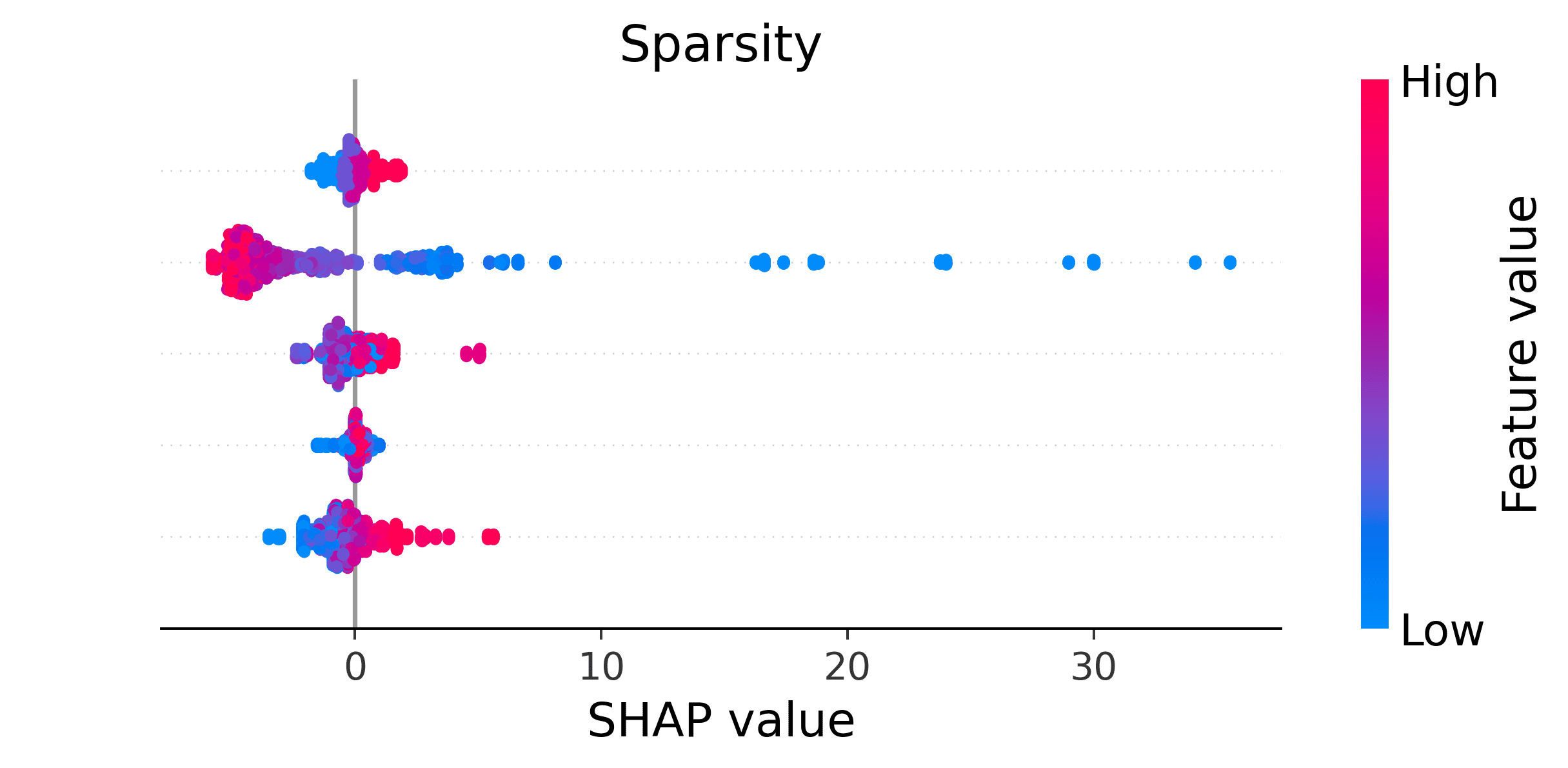}
\caption{SHAP values predicted by the random forest.}
\label{fig:shap_rf}
\end{figure}

In addition, SHAP values 
provide a fine-grained local interpretation by quantifying the contribution of each variable to the prediction~\cite{shapley1953}. Thus, while MDI offers a global importance based on the sum of impurity reductions, SHAP values reveal how each feature locally influences the model's output, enabling a more robust and nuanced analysis of the predictions. The SHAP values are calculated from the model's predictions on the 1000-point dataset and plotted in Fig.~\ref{fig:shap_rf}. Note that, given the very low computational cost of surrogate models, they are fast to evaluate.
SHAP values precisely assess the contribution of each parameter and the distribution of these contributions. For example, we observe that intolerance and density have a major impact, which translates into significant variations in convergence or spatial structure. Thus, SHAP analysis complements the global interpretation provided by the random forest by showing locally how each parameter influences the prediction.
For instance, the analysis reveals that the intolerance threshold has an extremely strong explanatory power, relatively centered and symmetric regarding the simulation's convergence. In contrast, for low density values, sparsity tends to explode and to overwhelm the other variables, revealing a more localized interpretability of density on sparsity.

\color{black}

A critical challenge in surrogate modeling for ABM is disentangling overlapping sources of uncertainty. Classical approaches map inputs deterministically, treating all variance as mere approximation error. However, a trustworthy surrogate must separate \textit{aleatoric uncertainty} (the simulator's intrinsic, irreducible stochastic noise) from \textit{epistemic uncertainty} (the model's lack of data in poorly sampled or highly nonlinear regions). Conflating these sources inevitably leads to overconfident predictions, particularly in unstable parameter regimes where chaotic macro-behaviors emerge~\cite{hullermeier2021aleatoric}.

To rigorously map these stability boundaries and capture the full complexity of the simulation, we implemented a data-driven pipeline inspired by conformal prediction frameworks, adapting a dual-scheme Random Forest methodology originally developed for spatial ecological modeling~\cite{shafer2008tutorial}. This approach quantifies the total uncertainty by independently modeling and then mathematically fusing its two constituent components~\cite{saves2026model}.

First, we establish the epistemic bounds of our approximation. We train a primary random forest ensemble to capture the central tendency of the system, specifically, a classifier for the binary convergence probability, and a regressor for the continuous sparsity metric. Because random forests rely on bagging, the epistemic variance \(\sigma_{\mathrm{epistemic}}^2\) can be extracted directly from the structural disagreement among the individual
decision trees. A high epistemic variance indicates a region where the surrogate struggles to infer a clear mapping due to complex, unresolved nonlinearities.

Second, we isolate the aleatoric component. To maximize data efficiency without requiring a separate calibration dataset, we leverage the unbiased out-of-bag absolute residuals from the primary ensemble. We deliberately model the absolute error rather than squared residuals, as the latter are notoriously sensitive to outliers and can severely destabilize tree-based algorithms. Statistically, the expected absolute deviation is directly proportional to the conditional standard deviation; for instance, under standard Gaussian assumptions, $\mathbb{E}[|y-\hat{y}|] =\sigma\sqrt{2/\pi}$. Therefore, the absolute residual serves as a mathematically robust, linear proxy for the local noise. A secondary random forest regressor is trained exclusively on these residuals, acting as a localized estimator of the system's intrinsic stochasticity $\sigma_{\mathrm{aleatoric}}$ and successfully identifying regions where the simulator generates highly divergent outcomes for identical initial
parameters.

Statistically, the internal randomness of the simulator and the surrogate's approximation limits constitute two fundamentally independent sources of error~\cite{hullermeier2021aleatoric}. Consequently, their variances are strictly additive. This allows us to formally synthesize them into a unified metric of total uncertainty through a Euclidean sum:
\(\sigma_{\mathrm{total}} = \sqrt{\sigma_{\mathrm{aleatoric}}^2 + \sigma_{\mathrm{epistemic}}^2}\).

\begin{figure}[H]
\centering
\includegraphics[height=7cm, width=0.95\columnwidth]{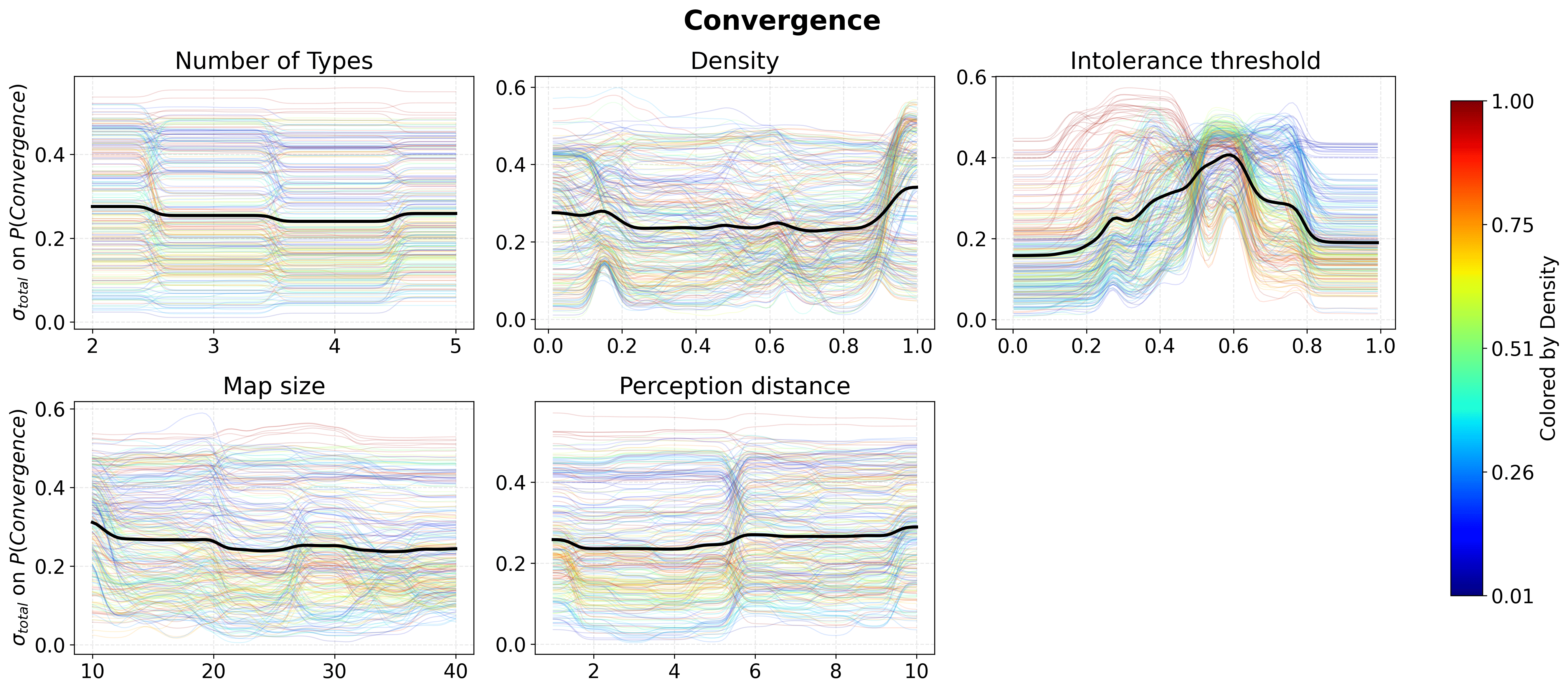}
\hspace{-0.1cm}
\includegraphics[height=7cm, width=0.95\columnwidth]{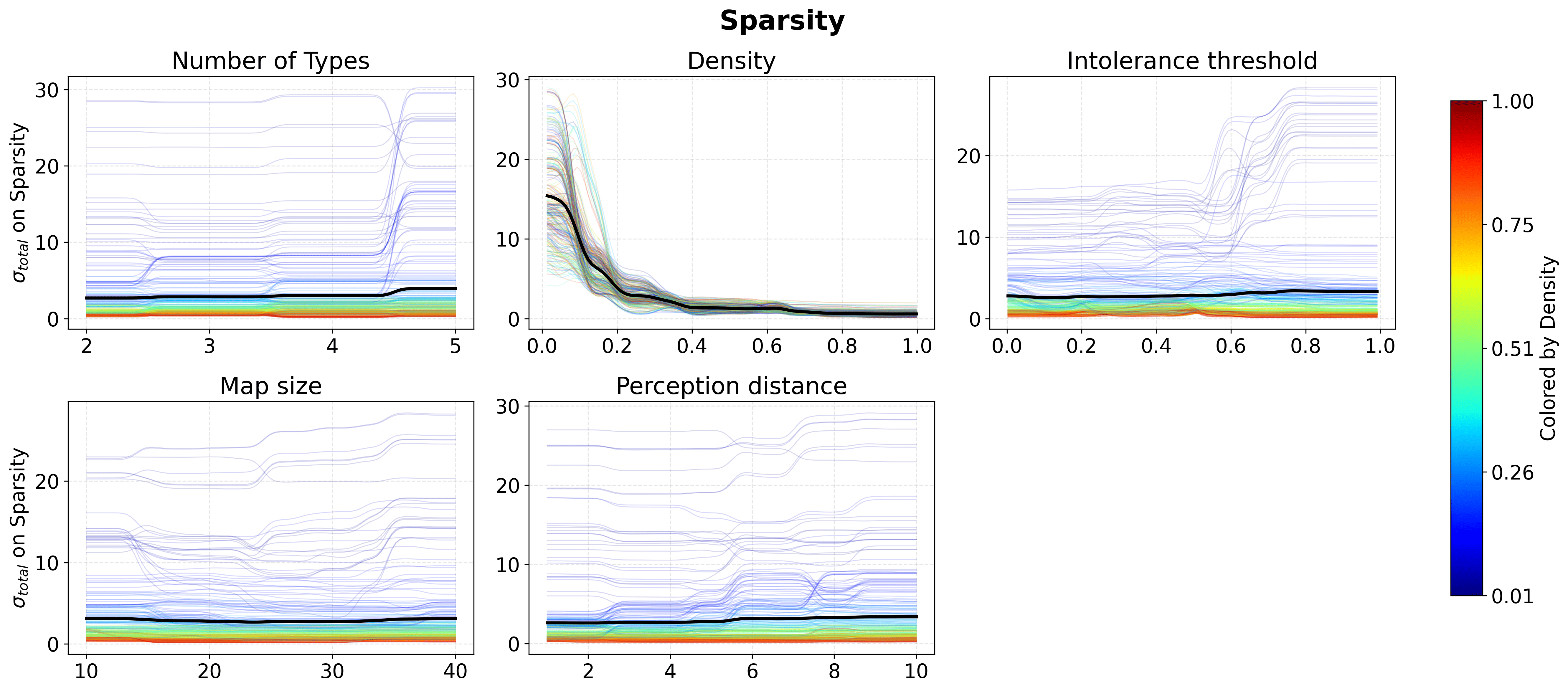}
\caption{PDP (thick black lines) and ICE (colored lines) curves showing the variations of total uncertainty (\(\sigma_{\mathrm{total}}\)), predicted
through the dual-scheme random forest, regarding convergence probability (top) and sparsity value (bottom).}
\label{fig:shap_dual_uncertainty}
\end{figure}

To translate this multi-dimensional uncertainty into actionable insights, we rely on PDP+ICE curves~\cite{goldstein2015peeking}. While PDPs isolate the marginal mean response of the uncertainty across the parameter space, the ICE curves expose individual \textit{ceteris paribus} deviations, revealing hidden localized interactions. 
Figure~\ref{fig:shap_dual_uncertainty}
visualizes this analysis, mapping the total uncertainty across the design space for the simulation's input features.

By mapping $\sigma_{\mathrm{total}}$, we provide a rigorous topographical map of the urban system's tipping points. The pronounced peaks in the uncertainty curves explicitly highlight critical phase transitions where the system hesitates between divergent macro-states. In these singular regimes, the simulator becomes physically unstable, driving an explosion in aleatoric noise, while the abrupt regime shifts simultaneously become exceptionally difficult for the surrogate to interpolate, thereby maximizing the epistemic variance~\cite{scheffer2009early}.

For the convergence probability, the total uncertainty peaks reveal a critical tipping point driven by the agents' intolerance levels. Crucially, the ICE curves, colored by agent density, demonstrate that this phase transition does not occur uniformly. Instead, the precise intolerance threshold that triggers a systemic shift is heavily modulated by the population density, highlighting a profound, nonlinear interaction between these two parameters.
Regarding sparsity, the uncertainty map confirms that stochastic behavior dominates at low density levels. Physically, a sparse environment offers many empty locations, granting agents excessive freedom of movement. Also, it renders the final spatial macro-state extremely sensitive to the initial random placement of agents, most of which will stay at the original position, massively increasing the role of initialization noise. 

Ultimately, identifying these structural vulnerabilities allows modelers to move beyond simple prediction, pinpointing precisely where minor policy interventions or initial random events could trigger massive systemic shifts~\cite{cavagna2010empirical}.
  \color{black}

\subsubsection{Comparing Instance-Level Explanations Across Multiple Surrogate Models}

\begin{figure}[H]
\centering
\hspace{-0.1cm}
\includegraphics[height=7cm, width=0.49\columnwidth]{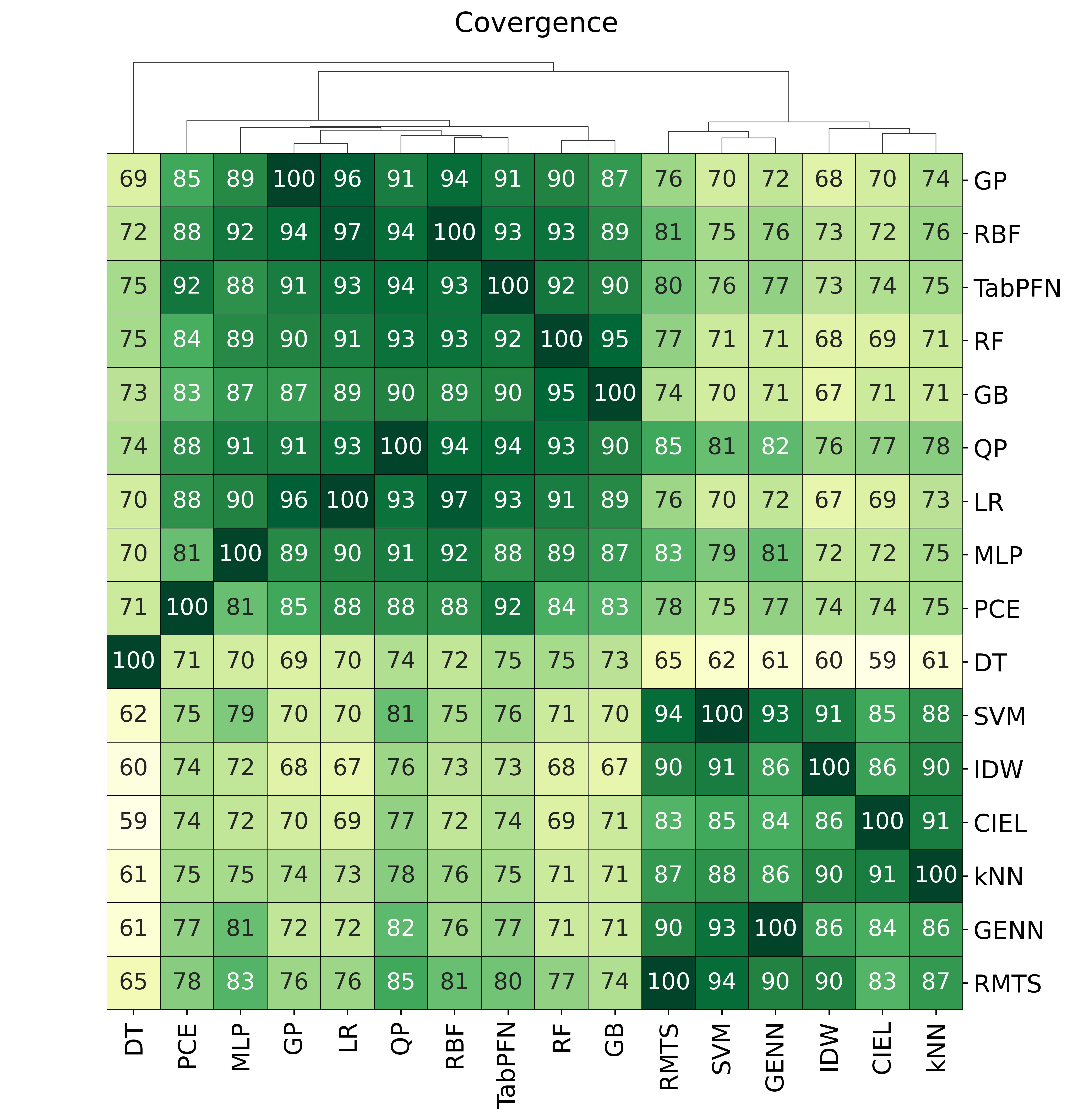}
\hspace{-0.1cm}
\includegraphics[height=7cm, width=0.49\columnwidth]{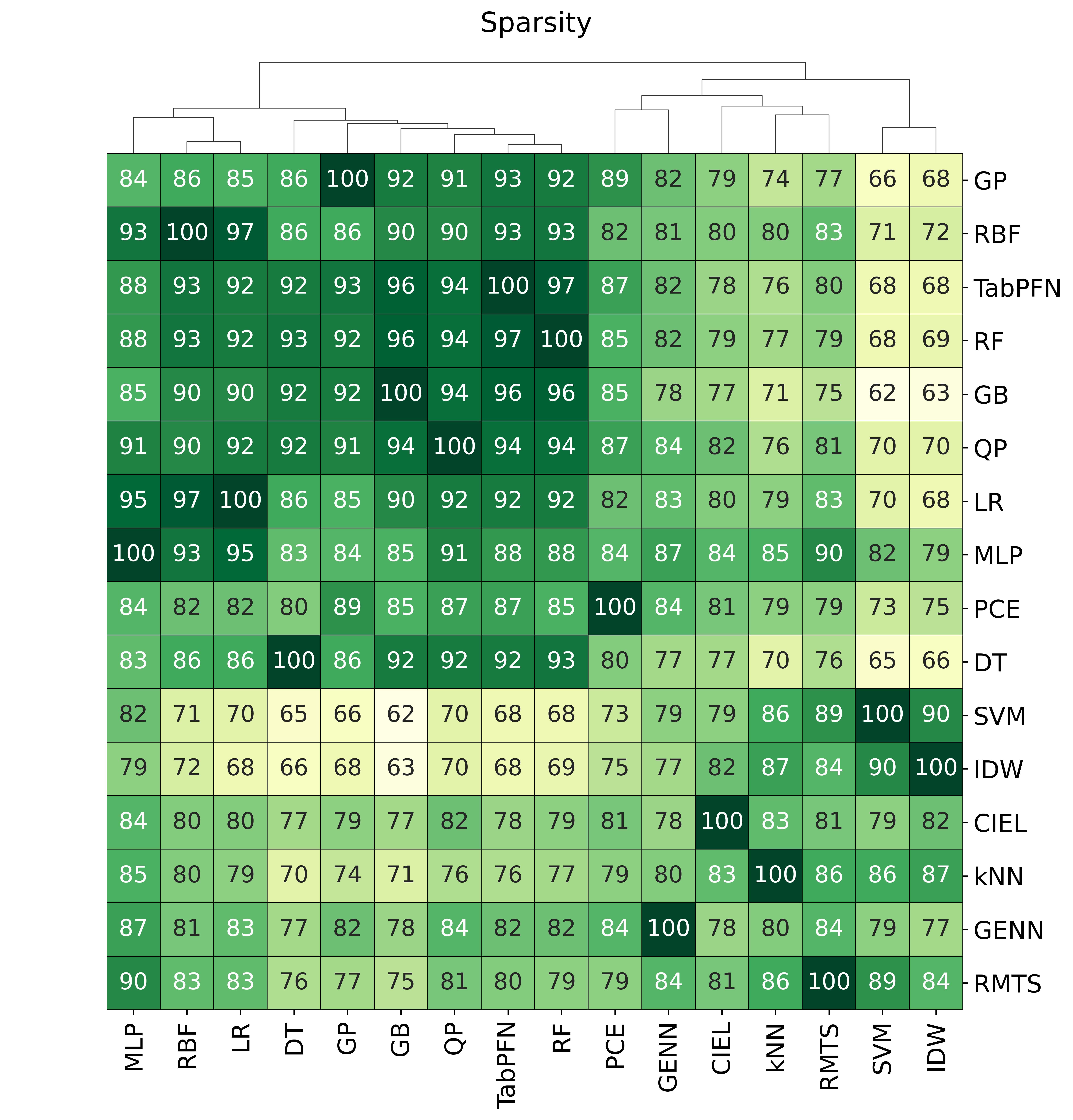}
\caption{SHAP values NDCG (\%) for every pairs of clustered models.}
\label{fig:ndcg_pairs}
\end{figure}

To quantify the quality of the explanations, we compute \color{black}{ the symmetrized NDCG correlations ($\rho_{NDCG}$) over each pair of models’ SHAP-value importance rankings in both the regression and classification tasks~\cite{burges2005learning}.}
Consequently, Fig.~\ref{fig:ndcg_pairs} displays, for both convergence classification and sparsity predictions, the NDCG adequacy of every pair of models in terms of SHAP value importance prediction on the validation test dataset. These SHAP prediction analyses reveal two subgroups of models whose diagonal elements belong to the subgroups with high internal NDCG scores and lower intergroup relationships, highlighting two groups of surrogate models sharing similar SHAP-based feature importance patterns, seemingly the local and the global models. 

\begin{figure}[H]
\centering
\hspace{-0.1cm}
\includegraphics[height=4.5cm, width=0.95\columnwidth]{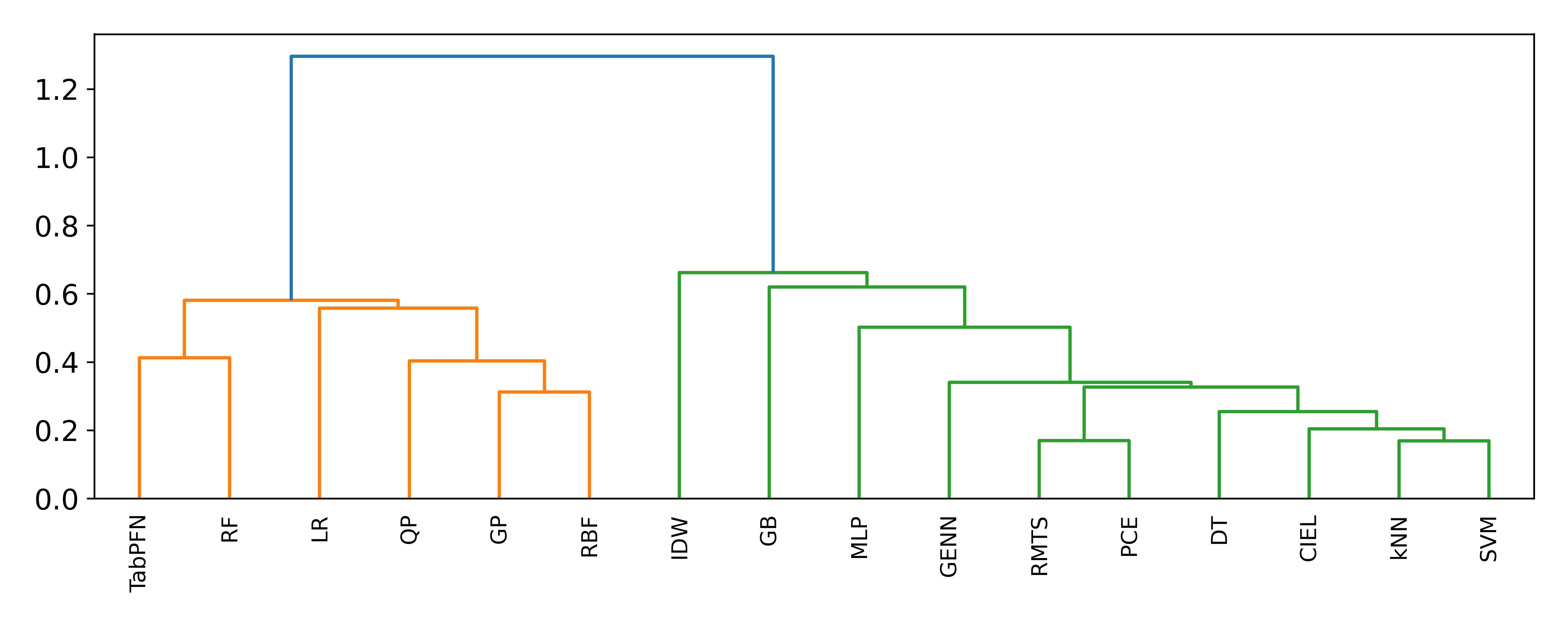}
\caption{SHAP values for NDCG-based hierarchical clustering.}
\label{fig:dendo}
\end{figure}

To have more information on this subject, we show a deeper analysis based on a clustering algorithm that gives the dendrogram in Fig.~\ref{fig:dendo}.  The dendrogram is obtained by agglomerative hierarchical clustering (Lance-Williams) using the average-linkage criterion on the dissimilarity matrix based on the pairwise NDCG similarities. The latter produces a linkage matrix whose entries record the clusters merged, the distances, and the new cluster size, which can then be transformed into the dendrogram shown~\cite{cormack1971review}.
The dendrogram confirms the presence of two distinct groups, consistent with the separation observed in the NDCG matrices, as more than half of the total dissimilarity is accounted for by the first major branching (from approximately 1.3 to 0.6 on the dissimilarity scale).  A first group containing models like GP and RBF that indeed appear in the same leaf as the first generalizes the second, followed by the similar models QP and LR, and to finish with, TabPFN seems to behave similarly to RF when it comes to generating SHAP explanations.  In the second group, we have all other methods; indeed, GB and DT are in the same group, and in fact, DT gives the worst results on convergence prediction, being divergent in behavior, with its ability to classify being really poor except with the GB methods, as the latter use DT as its core mechanism. We also have a subgroup with methods based on neighborhoods, such as CIEL or kNN, which are closely related and, knowing their locality, would be more fit for LIME explanation. This is particularly true for local or linear-based models, including SVM or LR. Note that LIME is free to evaluate with a local linear predictor as in CIEL. In this second group, we also find IDW, the last local model, but also the two closely related neural networks (MLP and GENN) and the two last other nonlinear models based on spline and on polynomial, which are both smooth combinations of a few basis elements.

The first group, in orange, contains the algorithms most prone to over-specialization, which tend to excel on familiar scenarios but struggle to generalize to broader, unseen patterns.    
Conversely, the second group, in green, contains models most prone to overfitting when they become too complex, all of which capture noise rather than general patterns. 
These findings suggest that, in most cases, our explanations characterize the surrogate rather than the underlying Schelling model we intended to understand.

An additional local analysis using SHAP and PDP is given in Appendix~\ref{app:companal}.  It inspects model-wise SHAP values and model agreement/disagreement to assess variable importance and explanation reliability, confirming that some models capture global trends while local effects appear as noise, whereas others recover local tendencies but overfit. While SHAP and PDP were primary, alternatives (Sobol' indices~\cite{robani2025}, HSIC~\cite{iooss2022}, conformal prediction~\cite{robani2025}, LIME~\cite{blanco-volle2024}) are applicable. Overall, surrogate models provide meaningful explanations for SABM even with sparse data.

Overall, TabPFN, RF, and GP strike the best balance between accuracy, uncertainty quantification, and interpretability.  Kernel methods (RBF) excel in classification, whereas ensemble and Bayesian approaches provide reliable uncertainty estimates.  Models with poor regression fit (\textit{e.g.} kNN or IDW) also yield low explanation fidelity, underscoring the link between predictive performance and interpretability. However, the most striking result is that RF, GN, DT, or QP, even if really simple models, could yield highly reliable results, and this seems to confirm their use in practice~\cite{dovsilovic2018explainable}. 

These findings offer practical guidance for selecting surrogate models in SABM, depending on whether regression precision, classification reliability, uncertainty calibration, or interpretable explanations are the primary objectives.  Future work should explore benchmark problems with analytically known mechanisms (\textit{e.g.} Lotka-Volterra/SIR dynamics) to further validate and compare SHAP-based explanation methods in this context~\cite{fabiani2024,shafer2008tutorial} and discuss many other types of explanations beyond SHAP~\cite{hu2020surrogate}. Still, these results provide a solid foundation for model selection depending on the targeted applications. These findings support recent advances in the use of surrogate models for SABM, which not only accelerate computations but also provide crucial uncertainty estimates for interpreting emergent phenomena in complex systems~\cite{fabiani2024}.

  \color{black}
Ideally, SHAP values should be computed directly from the high-fidelity simulator to ensure that the explanations faithfully reflect the simulator's behavior. In practice, however, SHAP-based explanations may require a large number of model evaluations per instance, rendering their direct computation prohibitively expensive for computationally costly simulators. Surrogate models therefore provide a practical means of approximating these explanations while substantially reducing the required computational cost~\cite{palar2026interpretable}.  Note that the aforementioned NDCG importance values reflect only absolute influence, and that other metrics that account for the sign of the influence, such as the composition of rank, influence, and accuracy described in~\cite{wang2023explanations}, may yield different results. 
  
Therefore, to theoretically ground our methodology, we systematically benchmark explanation consistency against standard predictive similarity. Beyond the symmetrized NDCG ($\rho_{NDCG}$), we compute a comprehensive set of cross-model similarity matrices across the entire surrogate ensemble. For consistency in explanation, we measure the Pearson correlation of raw SHAP values ($\rho_{SHAP}$) and the Kendall Tau rank correlation of absolute importance of SHAP characteristics ($\tau_{SHAP}$). For predictive similarity, we evaluated the Pearson correlation of the continuous predictions of the surrogates ($\rho_{pred}$) and the structural similarity of their errors ($S_{RMSE}$ for continuous outputs, and Cohen's Kappa $\kappa_{pred}$ for discrete classification outputs)~\cite{levy2026analyzing}.

\begin{figure}[H]
    \centering
    \hspace{-0.1cm}
    \includegraphics[height=7cm, width=0.49\columnwidth]{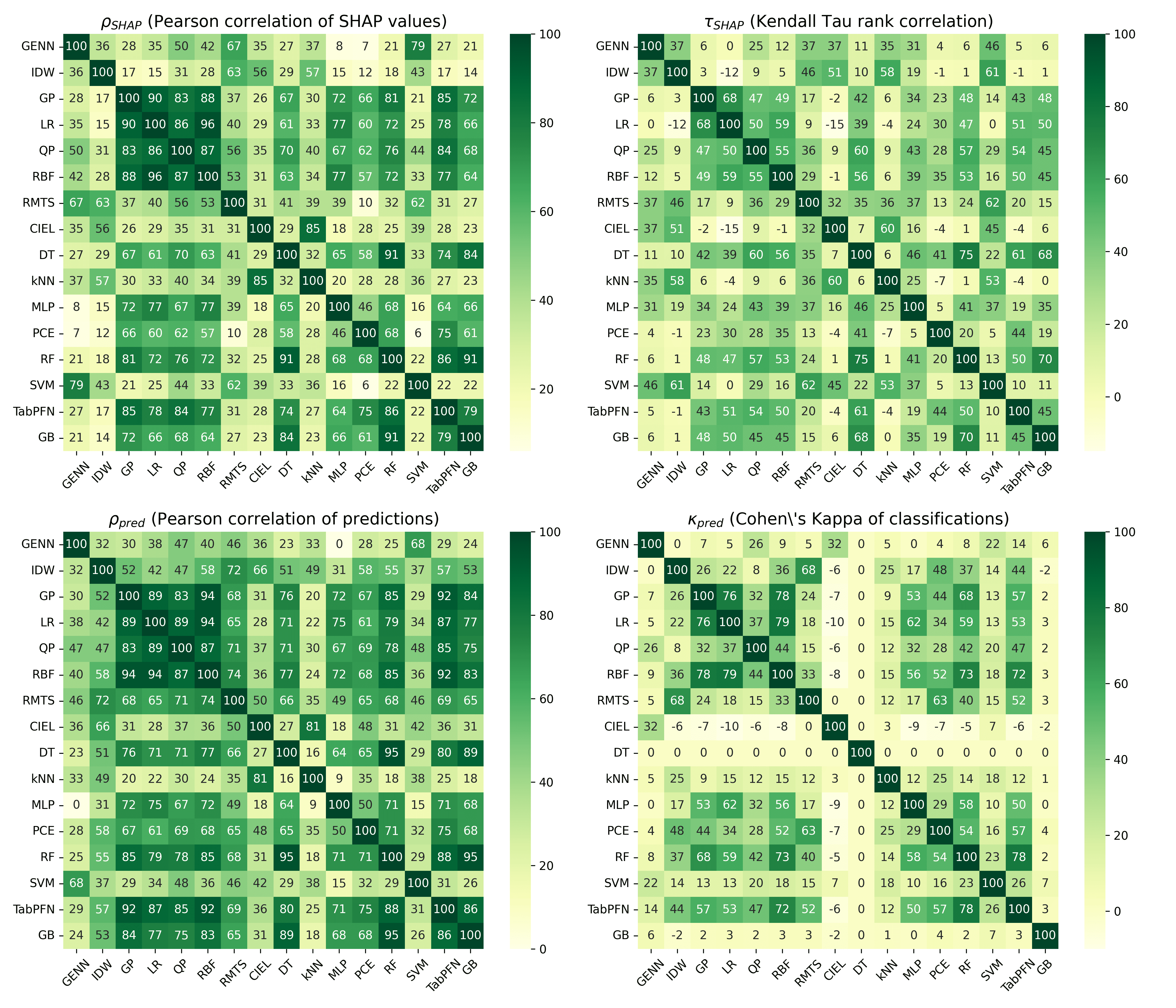}
    \hspace{-0.1cm}
    \includegraphics[height=7cm, width=0.49\columnwidth]{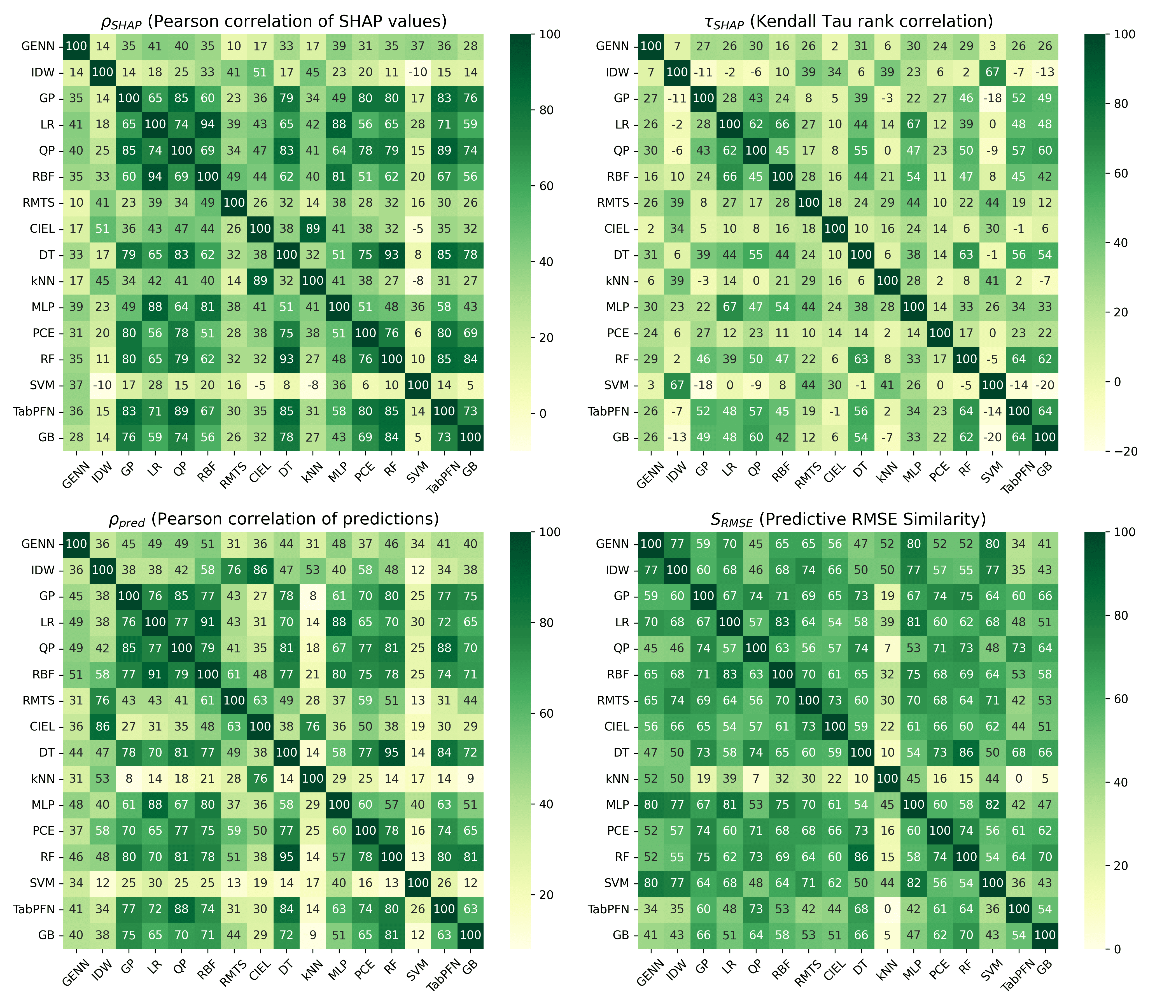}
    \caption{Cross-model similarity matrices for predictive performance ($\rho_{pred}$, $\kappa_{pred}$ or $S_{RMSE}$)
  and explanation consistency ($\rho_{SHAP}$, $\tau_{SHAP}$) on convergence (left) and sparsity (right) outputs. }
    \label{fig:4matrices}
    \end{figure}

\begin{figure}[H]
    \centering
    \hspace{-0.1cm}
    \includegraphics[height=7cm, width=0.49\columnwidth]{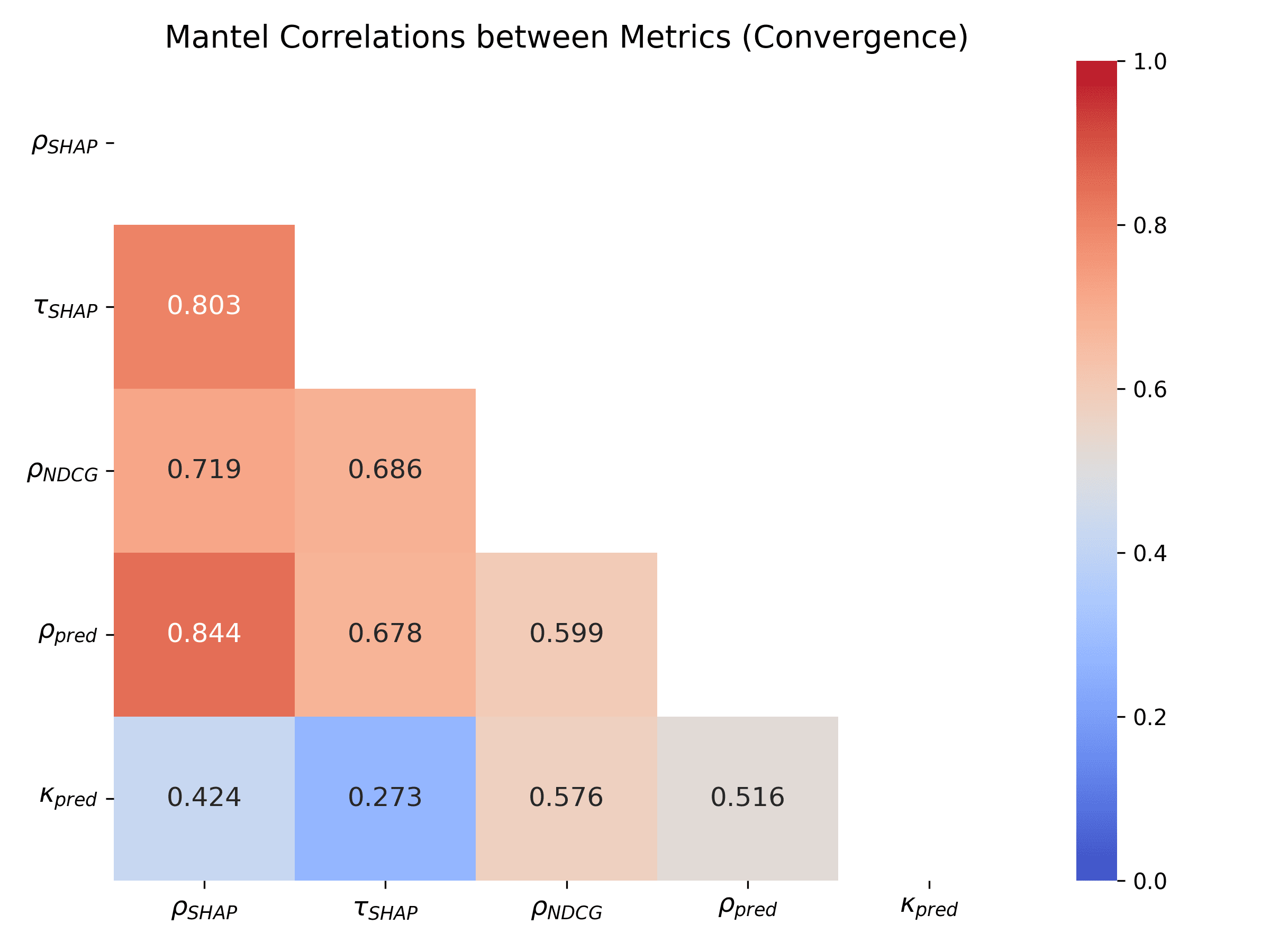}
    \hspace{-0.1cm}
    \includegraphics[height=7cm, width=0.49\columnwidth]{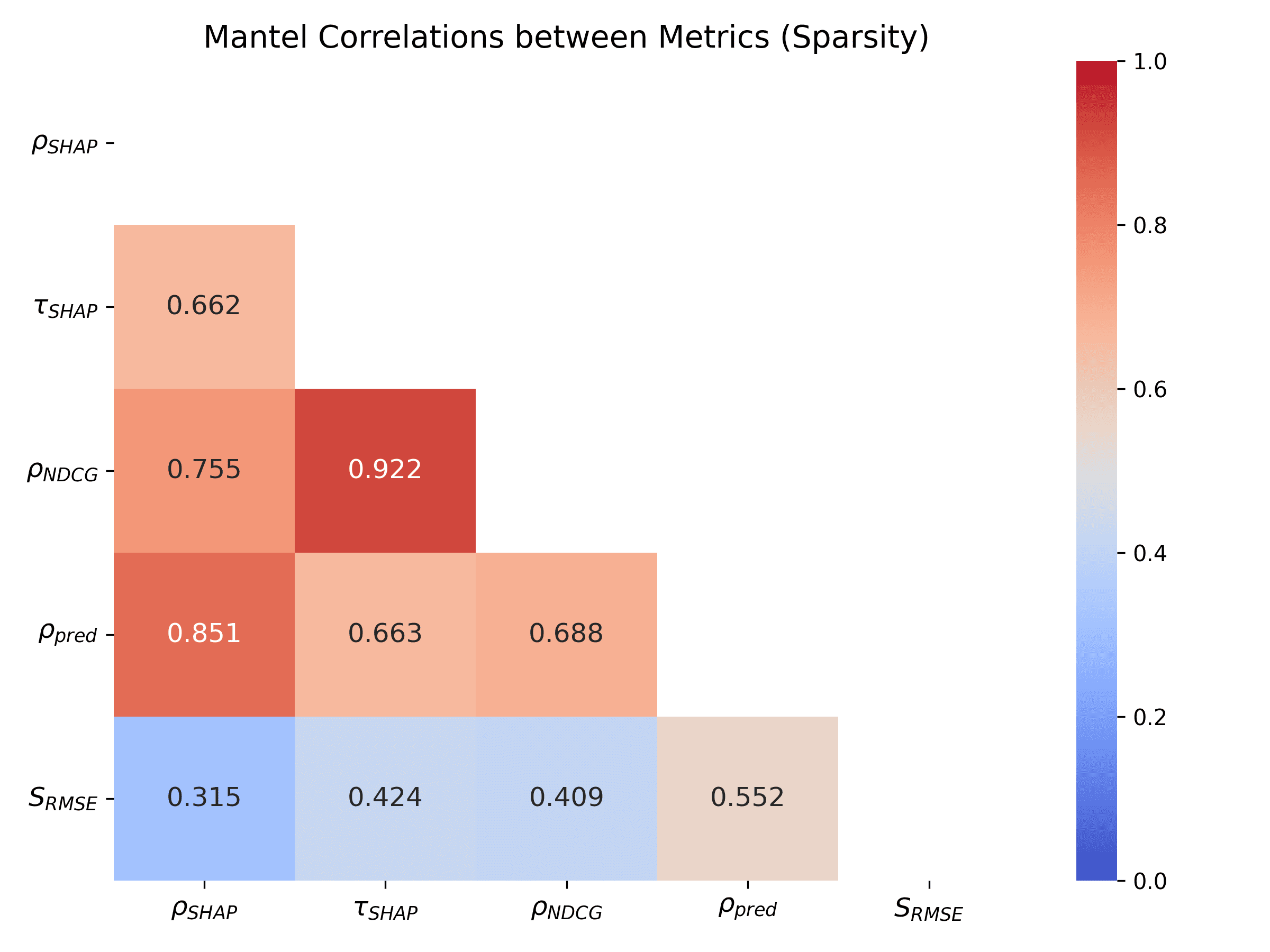}
    \caption{Mantel test meta-correlation matrices for convergence (left) and sparsity (right).}
    \label{fig:meta_matrices}
    \end{figure}

Figure~\ref{fig:4matrices} presents these four pairwise similarity matrices for both the convergence (classification) and sparsity (regression) outputs.
The visual clustering across these matrices reveals a general structural alignment with the previous study on NDCG SHAP.  This qualitative trend observation suggests that models producing similar predictive errors are relying on the same underlying feature attributions. To further quantify this structural alignment, we compute Mantel permutation tests to extract the meta-correlations between these similarity matrices, as displayed in Fig.~\ref{fig:meta_matrices}.
The results demonstrate highly significant correlations between predictive and d explanation similarities. For example,  Pearson's similarities of predictions ($\rho_{pred}$) and explanations ($\rho_{SHAP}$) are positive and around $85\%$ based on Mantel. Remarkably, the Mantel meta-correlations are consistently positive across all metric pairings (ranging from 0.27 to 0.92). This definitively demonstrates that throughout the surrogate ensemble, an increase in predictive agreement is always associated with an increase in explanation consistency, ruling out any severe inverse Rashōmon effect.

If several diverse surrogate models independently converge to both high predictive agreement and high explanation consistency, it is highly unlikely that they have converged on the exact same algorithmic artifacts. In our workflow, XAI metrics like NDCG do not replace standard predictive metrics; rather, they act as a critical secondary filter. As demonstrated through the  Mantel meta-correlations in Fig.~\ref{fig:meta_matrices}, models sharing similar predictive errors empirically rely on highly correlated SHAP attributions. Ultimately, RMSE and MCC ensure the surrogate's outputs match the simulator, while NDCG ensures the surrogate's internal causal mapping is robust and not an isolated algorithmic artifact.  Filtering the ensemble via explanation consistency thus explicitly isolates the most physically sound models within the Rashōmon set.

Building upon this comparison, \cite{levy2026analyzing} recently reframed this paradigm through the theoretical lens of the bias-variance-diversity trade-off in information and surrogate approximations. By evaluating ensemble strategies across 16 diverse problem domains, the study advocates hybridizing surrogate modeling and XAI to isolate complementary explanatory signals. Moreover, it investigates how ensemble performance depends on the underlying dataset and problem, showing how predictions and diversity correlate while being shaped by different problem settings.
  
    \color{black}

\color{black}

\subsubsection{Communicating model outputs for thematic insights}

The proposed surrogate-based XAI workflow serves as a stepping stone toward facilitating discussions with domain experts. In both case studies, the combination of surrogate acceleration and XAI transparency helps address the operational constraints of real-life stakeholders by fostering a more informed dialogue~\cite{drogoul2002multi}.

In the MDAO engineering context, the primary stakeholders are aerospace engineers and system architects~\cite{chan2025goal}. Traditionally, exploring the design space of an electric-hybrid aircraft requires navigating complex trade-offs between aerodynamics, structural integrity, and battery sizing, often bottlenecked by computationally expensive high-fidelity solvers~\cite{lambe2012extensions}. By deploying our workflow, design teams can use the trained surrogate to support iterative "what-if" explorations. When an engineer adjusts a design parameter like a wing aspect ratio or a battery mass, the surrogate provides rapid feedback on aircraft performance~\cite{donelli2023value}. The integrated XAI layer (via SHAP and PDPs) visualizes the underlying factors driving these performance changes, allowing engineers to discuss whether the surrogate's internal logic aligns with physical intuition. This transparent feedback loop supports iterative, multidisciplinary trade-off analysis between engineering departments~\cite{palar2026global}.

Conversely, in the Agent-Based Modeling context, the stakeholders are sociologists, urban planners, and policy-makers. Socio-technical simulations like the Schelling segregation model often explore potential policy interventions such as zoning laws or diversity initiatives~\cite{drogoul1994multi}. However, running thousands of stochastic simulations to observe emergent macro-behaviors is impractical during interactive policy workshops. Here, the surrogate model acts as a "computational laboratory" that runs in milliseconds, while the XAI maps how local micro-level policies (e.g., individual tolerance thresholds) influence macro-level systemic outcomes (e.g., urban segregation sparsity). Because our workflow robustly isolates aleatoric noise and filters for explanation consistency, stakeholders can use the provided SHAP values to explore potential policy levers, fostering a more informed dialogue between modelers and thematicians~\cite{saves2026model}.

More broadly, in the rapidly expanding literature on participatory modeling and group model building, simulations are increasingly conceptualized as "boundary objects" that facilitate communication and consensus among stakeholders with conflicting interests~\cite{taillandier2019participatory}. By embedding XAI directly into this collaborative loop, our framework helps transition the simulation from a static oracle to a basis for interactive dialogue~\cite{jamshidi2022building}. This ensures that the generated models do not merely reflect mathematical optimization, but also serve as a foundation for thematicians to question and refine the model's assumptions, ultimately supporting the iterative discussions required in complex environments~\cite{lorscheid2012opening}.

Beyond the urban segregation and hybrid-electric aircraft paradigms, this surrogate-based XAI pipeline has been generalized to support exploration across a wide array of complex application domains. For instance, in socio-ecological modeling, the methodology has been deployed to map complex interaction effects and highlight potential tipping points of spatial predator-prey dynamics~\cite{saves2026model}. Furthermore, the integration of spatial explainability within ensemble multi-agent systems has proven useful for analyzing abrupt regime changes and evaluating both epistemic and aleatoric uncertainties in dynamical systems~\cite{blanco-volle2024}. This framework also naturally extends to industrial ecology, where it serves as an exploratory tool for understanding decentralized circular economy markets through multi-agent reinforcement learning simulations~\cite{Mastio2026Adaptive}. Finally, in purely deterministic engineering contexts, novel Global Sensitivity Analysis (GSA) metrics derived from Individual Conditional Expectations (ICE) have been used to explore masked interaction effects in complex aircraft aerodynamic design~\cite{palar2026global} for communication to stakeholders.
\color{black}

\section{Conclusions and perspectives}
\label{sec:conclu}

\color{black}

This work set out to demonstrate how surrogate modeling combined with Explainable AI (XAI) can turn computationally expensive, blackbox simulators into interpretable, trustworthy tools for decision support. By training lightweight emulators on a relatively small number of simulations, we obtained order-of-magnitude speedups while successfully reproducing and explaining key complex behaviors across two complementary domains: a multidisciplinary design analysis and Optimization (MDAO) engineering test case, and an Agent-Based Model (ABM) for socio-technical policy assessment. 

The MDAO case highlights the strengths of surrogate models in structured, high-dimensional design spaces where the main bottleneck is the cost of high-fidelity solvers. In the DRAGON electric-hybrid aircraft design study, a surrogate trained on approximately 300 configurations recovered critical trade-offs between sizing parameters, identified dominant variables, and revealed interaction patterns that would have otherwise required hundreds of costly computational fluid dynamics and structural runs, \color{black}{effectively reducing exploration time from weeks to seconds}. In this context, surrogates act as accelerators, while XAI techniques illuminate which design variables drive critical outputs, supporting engineering intuition and multidisciplinary communication.

By contrast, the ABM case exposes domains dominated by uncertainty, stochasticity, and emergent behavior. In the urban segregation model, a surrogate trained on roughly 50 stochastic realizations accurately reproduced segregation indices and spatial patterns. SHAP-based explanations successfully mapped model assumptions and input distributions to emergent collective outcomes, highlighting policy levers (such as population density and perceived neighborhood diversity) that influence sparsity and segregation. Here, surrogates do more than speed up computation: they enable systematic replication, variance decomposition, and exploration of policy scenarios that would otherwise be prohibitive, bridging the gap between technical modeling and stakeholder understanding.

Seen together, the two cases illustrate complementary facets of a surrogate-driven XAI workflow. The MDAO setting emphasizes precision, smooth response surface methodologies, and exploration of high-dimensional multimodal design spaces; the ABM setting emphasizes stochastic exploration, robustness, and interpretation of emergent phenomena. Both converge on the methodological insight that surrogates and XAI are enablers, rather than end goals, of transparent, efficient simulation science.

Methodologically, our experiments indicate several practical recommendations. Careful diagnostic-driven workflows that combine space-filling and adaptive sampling, probabilistic and ensemble surrogates, and complementary explainability tools provide a practical path to both accelerate and interpret complex simulations. Ensemble approaches are often particularly valuable because model diversity helps capture different aspects of the underlying simulator at both global and local scales. For stochastic simulators, repeated sampling and ensembles help capture intrinsic variability and avoid misleading point estimates.

However, trustworthy use of surrogate-based XAI pipelines requires explicit attention to evaluation. Original simulators can be hard to validate when ground truth is unknown or when outputs are aggregated and may mask emergent behaviors. The surrogate introduces a further layer of approximation: assessing its predictive fidelity and uncertainty requires independent validation, replication to capture stochastic variability, and attention to both error metrics and uncertainty quality. More notably, the surrogate-based explanations characterize the surrogate rather than the complex system itself. Mixed variable types (discrete architectural choices, categorical socio-types, continuous sizing parameters) demand surrogate and XAI methods that can handle heterogeneous inputs without losing interpretability.

Looking ahead, several directions can increase robustness and practical value. Fidelity and hybrid approaches that combine coarse and fine models can reduce costs while preserving accuracy. Sequential and adaptive active learning schemes, such as multi-objective Bayesian optimization, can progressively enrich datasets and focus computing on informative regions for exploration, rare-event estimation, or reliability assessment. \color{black}{Building on the insights from the aleatoric uncertainty analysis, future work could explore adapting this methodology to define adaptive, heterogeneous replicate budgets, dedicating more simulation runs specifically to the highly stochastic regions identified by the XAI and improving the management of the relationship between epistemic and stochastic uncertainty sources.} Embedding causal reasoning and counterfactual explanations could make recommendations more actionable under uncertainty. Finally, real-time integration of surrogate-based XAI loops in participatory platforms can enable continuous stakeholder engagement, rapid iteration, and transparent decision-making by integrating visual interfaces to communicate predictions, uncertainty, and explanatory insights.

\color{black}

\section*{Conflict of interest}
The authors have no conflict of interest to declare.

\section*{Data availability and replication}
The codes and data used to produce the results are open-source and available online: {\url{https://github.com/ANR-MIMICO/SMPT_XAI}}.

\section*{Acknowledgements}
Pramudita Satria Palar would like to acknowledge financial support from Institut Teknologi Bandung through the Riset Kolaborasi International 2025 scheme.  This work is part of the activities of the ONERA - ISAE - ENAC joint research group. The research presented in this paper has been performed in the framework of the 
MIMICO research project funded by the Agence Nationale de la Recherche (ANR) n$^\text{o}$ ANR-24-CE23-0380. 
The research presented in this paper has received funding from the National Association for Research and Technology under grant number CIFRE 2023/1398 and from Soben company.
\color{black}{The authors gratefully acknowledge Gojek.}
\appendix

\section{Complementary Local Analysis of Variable Importance}
\label{app:companal}
\begin{figure}[H]
\centering
\hspace{-0.25cm}
\includegraphics[width=1\linewidth]
{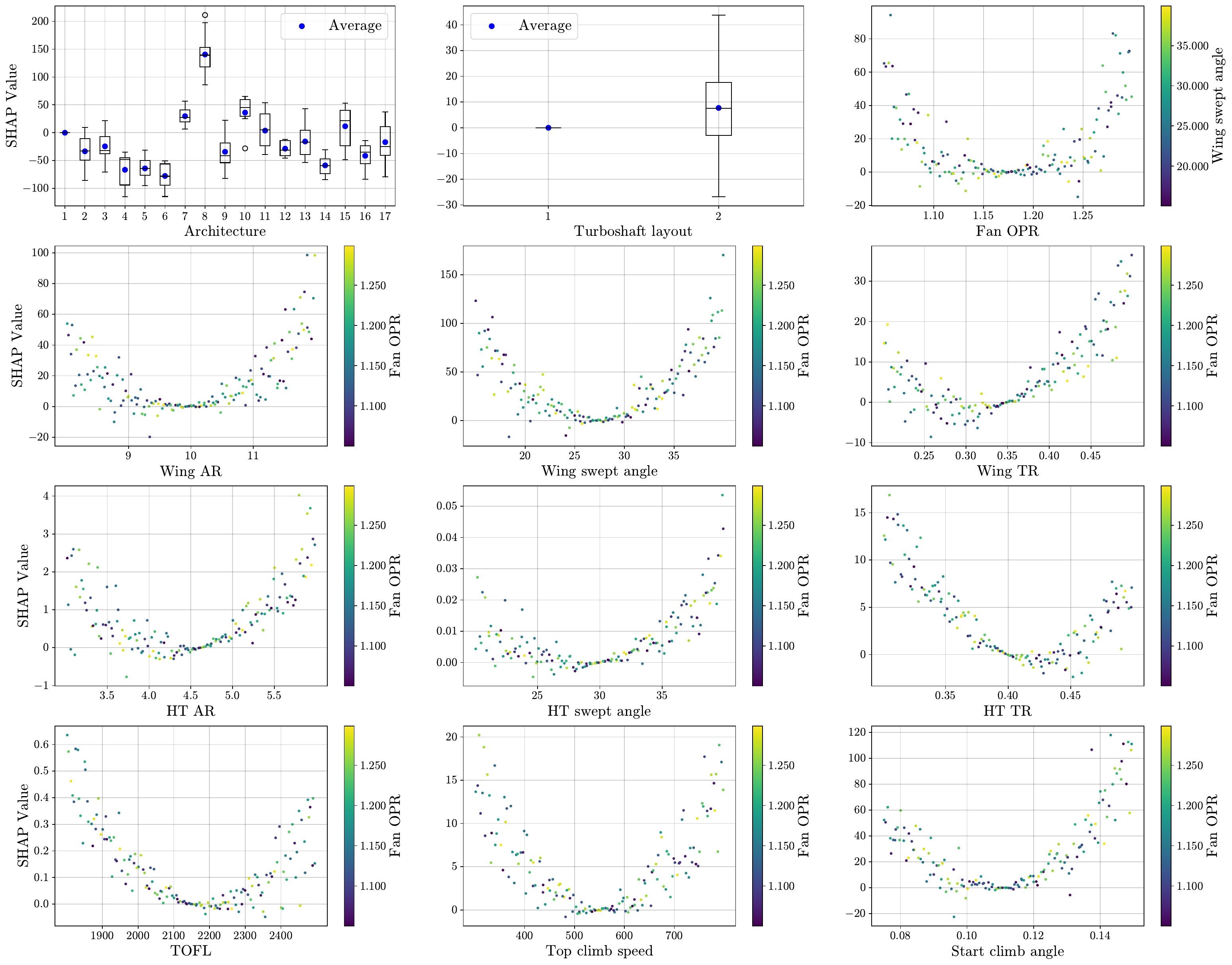}
\caption{SHAP dependence plots for the \texttt{DRAGON} problem output predicted variance.}
\label{fig:DRAGON_SHAP_PLOT_var}
\end{figure}

\color{black}
Figure~\ref{fig:DRAGON_SHAP_PLOT_var} complement the result of Section~\ref{subsec:resdragon}
on the necessity of having an anchored PDP+ICE figure to locally analyze the uncertainty and that the anchor is misleading for the SHAP plots.

\color{black}

\begin{figure}[H]
\centering
\hspace{-0.1cm}
\includegraphics[width=0.9\columnwidth]{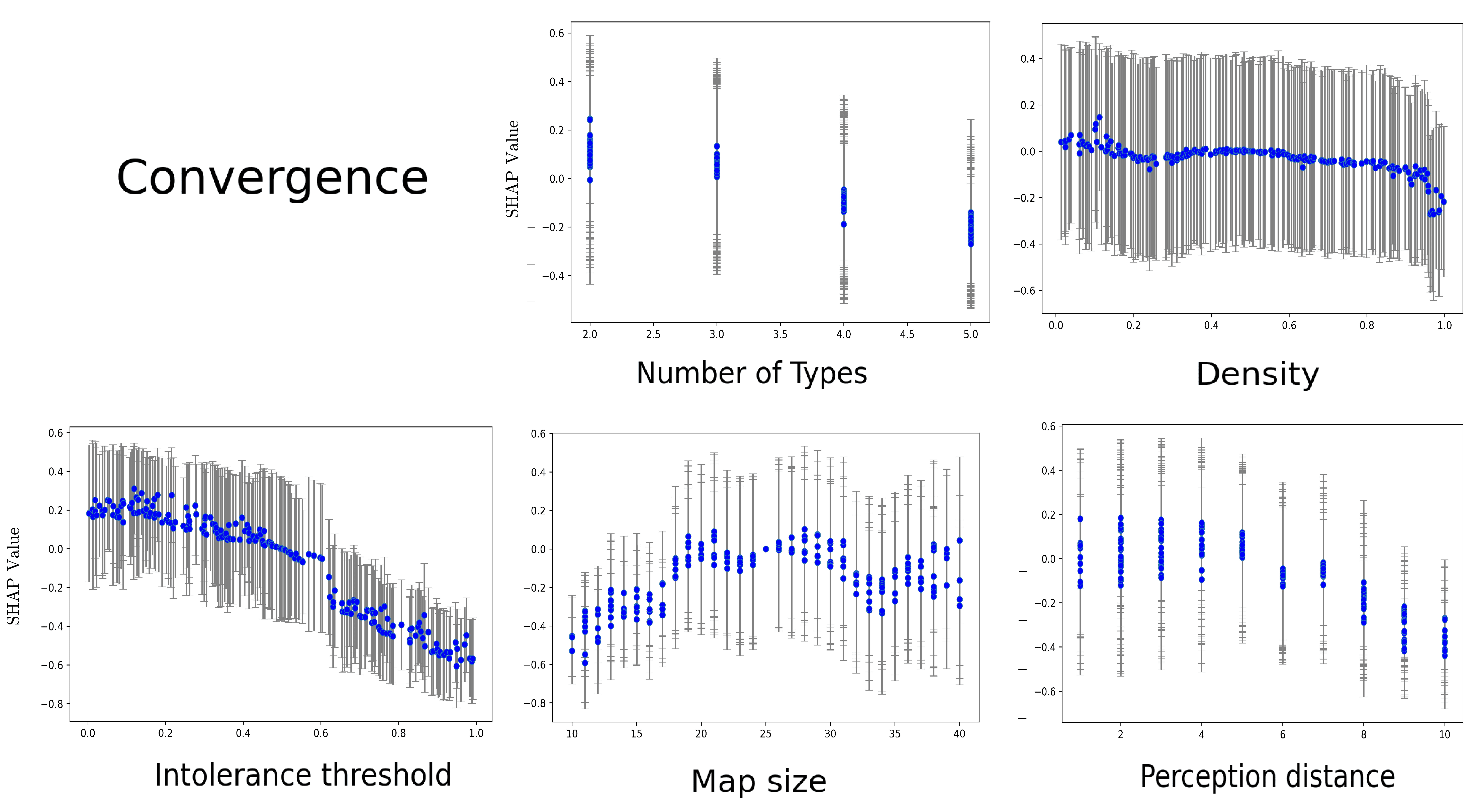}
\hspace{-0.1cm}
\includegraphics[ width=0.9\columnwidth]{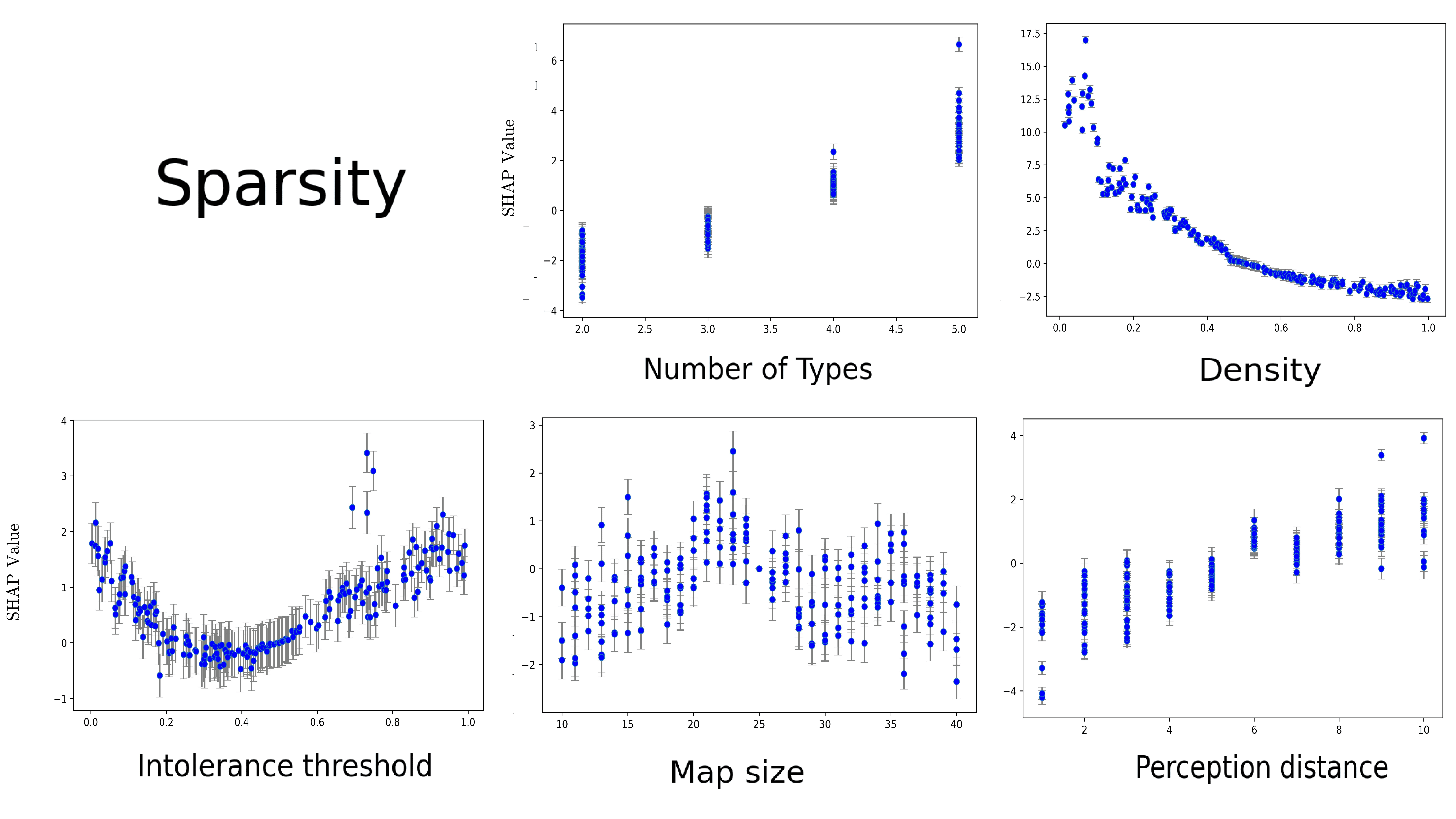}
\caption{SHAP values as a function of the variables predicted by the average of all surrogate models.} 
\label{fig:shap_vars}
\end{figure}

To complement the pairwise agreement analysis of Section~\ref{subsec:rez_abs}, Fig.~\ref{fig:shap_vars} shows the mean SHAP values as a function of input variables, with uncertainty represented as agreement $\pm$ disagreement across the 120 model pairings.
Together, these analyses provide both a global perspective via pairwise and clustered agreement, and a local view through variable-level SHAP behavior. The latter enables a more complete interpretability assessment across surrogate model families. 

 Figure~\ref{fig:shap_vars} shows an aggregated visualization of all the models, but the explanations could also be drawn model-wise, generating the two groups identified in the dendrogram of Fig.~\ref{fig:dendo}.
 These plots emphasize which regions in the input space yield robust explanations and where model disagreement suggests higher epistemic uncertainty. For instance, in low-density data regions, SHAP values tend to fluctuate more across models, reflecting instability in the learned attributions. Conversely, high-agreement regions highlight stable model behavior and consistent variable influence, supporting model trust.
 The models in the orange group predict a strong influence for intolerance on convergence and for density on sparsity, which are the most prominent average effects, whereas the models in the green group give more importance to the local effect even if not significant on average, such as density on convergence and map size on sparsity, which explains the two groups disagreeing on variable importance as shown in the NDCG matrices of Fig.~\ref{fig:ndcg_pairs}.
 
All features being treated equally, SHAP mainly helps identify which features influence the similarity measure most. Comparing SHAP values across models offers valuable insights into feature importance and model behavior (highlighting consistency, model-specific priorities, and robustness checks) yet requires caution because SHAP ranges and meanings differ by algorithm. While such comparisons can inform feature selection and deepen understanding, they don’t directly indicate which model is superior.

In conclusion, SHAP values help explain model predictions, but their interpretation differs depending on the algorithm. Orange models, such as GP, reflect how each feature affects the predicted mean and uncertainty of the output, and SHAP highlights feature importance in shaping the underlying function rather than direct instance similarity. In contrast, green models such as kNN make predictions based on the similarity between data points, using distances in feature space. SHAP values here indicate how much each feature contributes to determining the nearest neighbors and shaping the final prediction. There is no clear response on what is a "really important" variable, except that it depends on whether the interest of the practitioner is more in the global trends or in the potential local effects.
Moreover, studying a complex system through a surrogate model is a highly impactful method to gather information, as most of the resulting information comes from the structure of the surrogate model itself instead of coming from the complex system.

\begin{figure}[H]
\centering
\hspace{-0.1cm}
\includegraphics[ width=0.92\columnwidth]{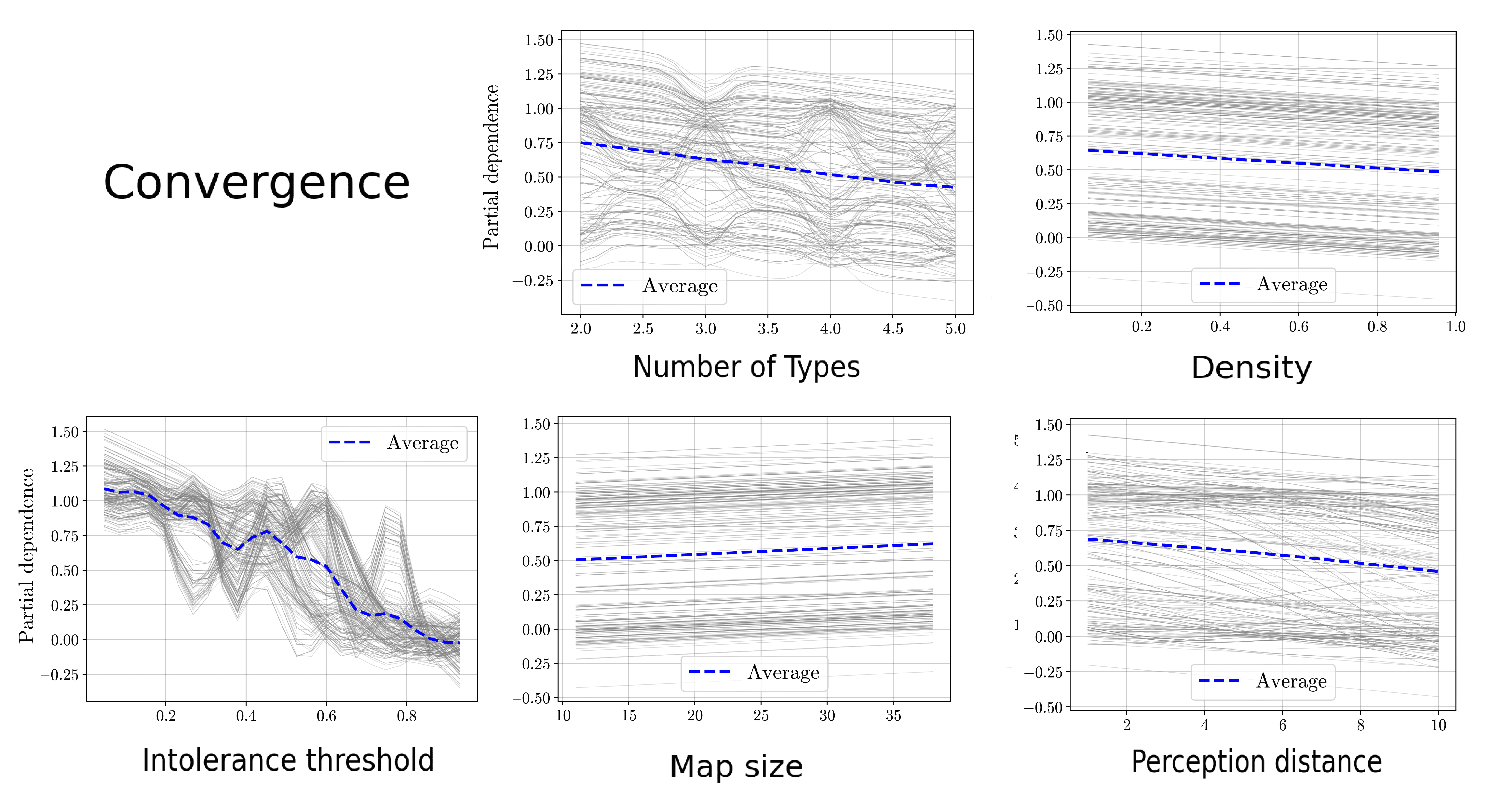}
\hspace{-0.1cm}
\includegraphics[ width=0.92\columnwidth]{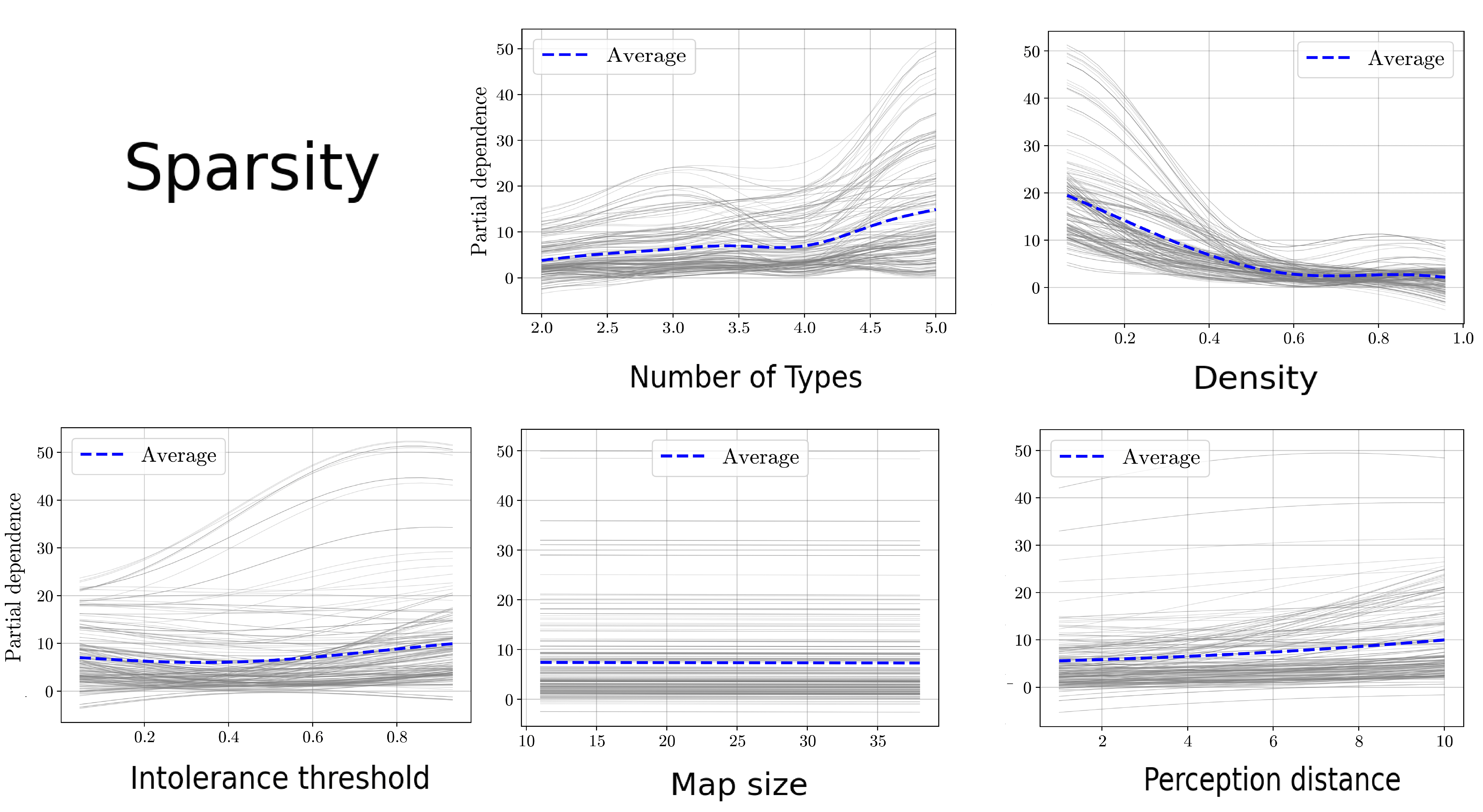}
\caption{PDP and ICE plots as a function of the variables predicted by the GP model.}
\label{fig:pdp_ice}
\end{figure}

In Fig.~\ref{fig:pdp_ice}, the Partial Dependence Plots (PDP) illustrate the average effect of each variable on the prediction by neutralizing the influence of the others, while the Individual Conditional Expectation (ICE) curves reveal the variability of individual responses, highlighting the interactions between the parameters~\cite{molnar2019} predicted on the validation set by the GP surrogate model.

In the presented figure, the global trends observed in the PDP+ICE, such as a monotonic or nonlinear effect on the output, are confirmed by the dispersion of the SHAP values, which also reveals the presence of interactions, although these are relatively weak for some parameters. These complementary analyses thus provide a deep and nuanced understanding of the model's behavior and the influence of each variable on the simulation.
The PDPs show the average effect of each variable on the predicted outputs by neutralizing the influence of the other parameters. This allows us to determine whether the effect of a variable is monotonic, such as increased density making convergence more difficult, or if there is a nonlinear relationship, such as an optimum or a tipping point. The ICE curves reveal the prediction trajectories for each individual simulation. If these deviate significantly from the PDP curve, it highlights interactions between the studied variable and other variables or random factors: for instance, the intolerance threshold may increase or decrease sparsity depending on the perception distance and whether the simulation converges or not, leading to a non-representative PDP average.

\bibliographystyle{plain}
\bibliography{main.bib}
\pdfbookmark[1]{References}{sec-refs}

\end{document}